\documentclass[runningheads]{llncs}

\usepackage{eccv}

\usepackage{eccvabbrv}
\usepackage{graphicx}
\usepackage{booktabs}
\usepackage{multirow}
\usepackage{tabularx}
\usepackage{pifont} % 提供打钩符号 \ding{51}
\usepackage{array}
\usepackage{float}
\usepackage{algorithm}
\usepackage{algorithmic}
\newcolumntype{Y}{>{\centering\arraybackslash}X}

\usepackage[table]{xcolor}
\usepackage[accsupp]{axessibility}  % Improves PDF readability for those with disabilities.
\usepackage{subcaption}
\usepackage{adjustbox}
\usepackage{hyperref}

\usepackage{orcidlink}

\begin{document}

\title{Magnitude-Direction Decoupling for Fast Video Generation with Flow Matching Models} 

\titlerunning{Abbreviated paper title}

\author{Haonan Xu\inst{1,2}\textsuperscript{*,\dag} \and
Feiyang Chen\inst{2} \and
Songkui Chen\inst{2} \and
Hongpeng Pan\inst{2} \and
Zhefeng Wang\inst{2} \and
Xinyu Duan\inst{2} \and
Baoxing Huai\inst{2} \and
Yang Yang\inst{1}\textsuperscript{*,\ddag}}

\insert\footins{\footnotesize \textsuperscript{\dag}Work done while Haonan Xu was with Huawei.}
\insert\footins{\footnotesize \textsuperscript{\ddag}Corresponding author. Email: yyang@njust.edu.cn}

\authorrunning{H. Xu et al.}
% First names are abbreviated in the running

\institute{Nanjing University of Science and Technology, Nanjing, China
\and Huawei, Shanghai \& Hangzhou, China}

\maketitle

\begin{abstract}
  Flow matching models for video generation achieve impressive performance but suffer from high computational overhead due to iterative denoising. In fact, the original model is not necessary for all denoising steps, allowing some steps to use lightweight alternatives for faster sampling. However, directly using caching or lightweight models can deviate from the original denoising trajectory, resulting in suboptimal performance. Through empirical analysis, we find that lightweight models can robustly capture the magnitude components of the original model's output, while caching provides reliable directional guidance. Building on this insight, we propose the Magnitude-Direction Decoupling (MDD) method, which adaptively employs a direction-calibrated lightweight model as a substitute for the original model to accelerate inference and effectively correct deviations in the denoising trajectory. Moreover, MDD further reduces inference costs by reusing magnitude information under classifier-free guidance (CFG). As a result, MDD offers a more reliable and lightweight solution to accelerate sampling. Experiments show that MDD outperforms existing acceleration methods, delivering promising speedups (e.g., up to 2.95× on Wan2.1) while preserving high visual fidelity and content richness.
  \keywords{Fast Video Generation \and Flow Matching Models \and Magnitude-Direction Decoupling}
\end{abstract}

\section{Introduction}
Driven by diffusion models \cite{DBLP:conf/nips/DhariwalN21,DBLP:conf/nips/HoJA20}, visual generation has achieved remarkable success in recent years. A growing body of research \cite{DBLP:journals/corr/abs-2503-20314,DBLP:journals/corr/abs-2412-20404,DBLP:conf/iclr/YangTZ00XYHZFYZ25,DBLP:journals/corr/abs-2405-18991,DBLP:journals/tmlr/Ma0CJ0LC025} continues to explore the frontiers of diffusion models, achieving impressive levels of fidelity and temporal coherence in video generation. Notably, flow matching \cite{DBLP:conf/iclr/LipmanCBNL23} has emerged as a compelling framework for achieving faster convergence and improved controllability in video generation \cite{DBLP:conf/icml/EsserKBEMSLLSBP24, DBLP:conf/iclr/LiuG023}, and it is widely being adopted by state-of-the-art (SOTA) methods.

\begin{figure}[!htbp]
    \centering
    \begin{subfigure}{\textwidth}
        \centering
        % \caption*{\small{Prompt: ``A boat sailing leisurely along the Seine River with the Eiffel Tower in background by Vincent van Gogh.''}}
        \begin{subfigure}{0.240\textwidth}
            \centering
            \includegraphics[width=\textwidth]{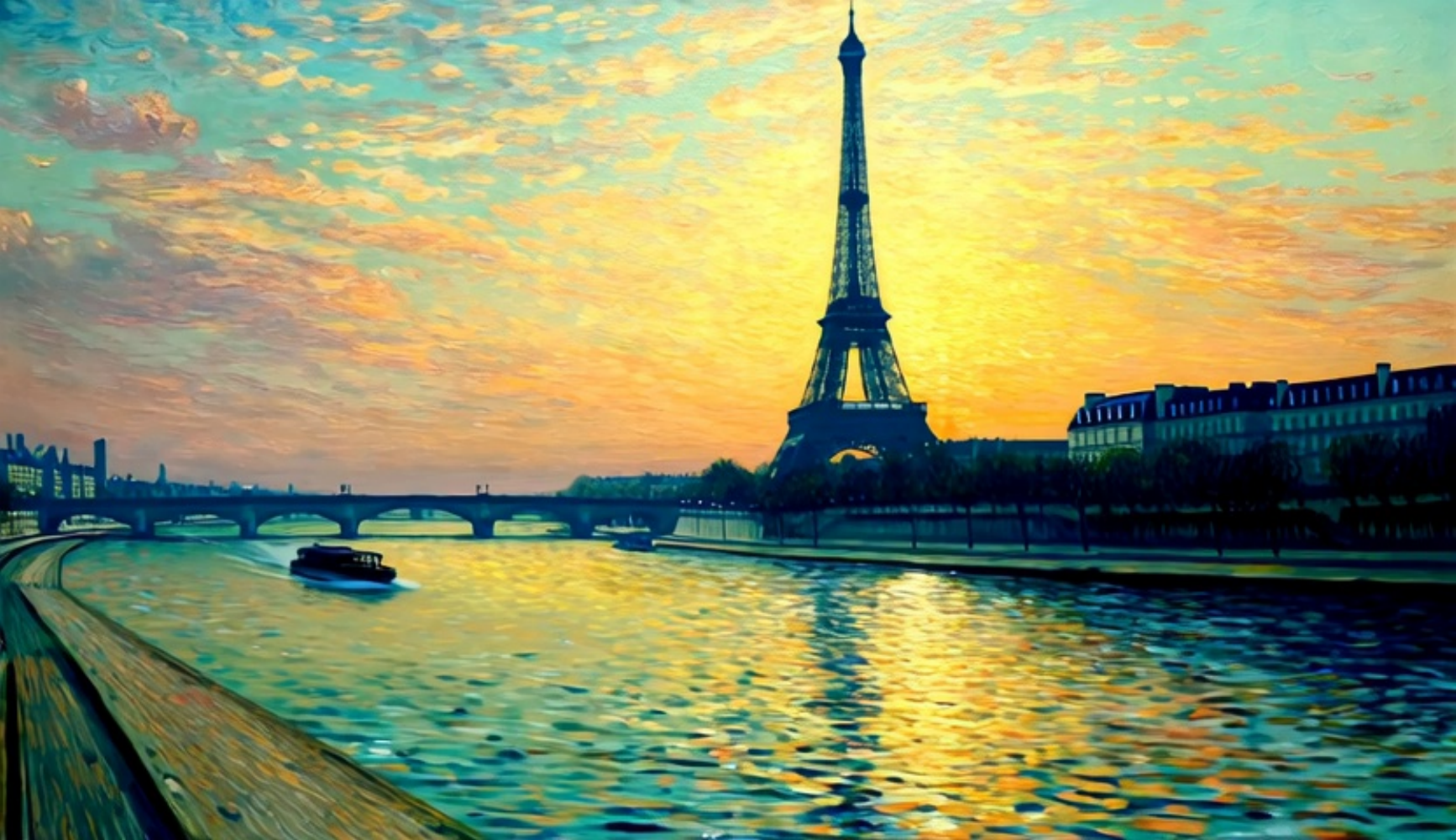}
            \captionsetup{skip=2.5pt}
        \end{subfigure}%
        \hfill
        \captionsetup{skip=2.5pt}
        \begin{subfigure}{0.240\textwidth}
            \centering
            \includegraphics[width=\textwidth]{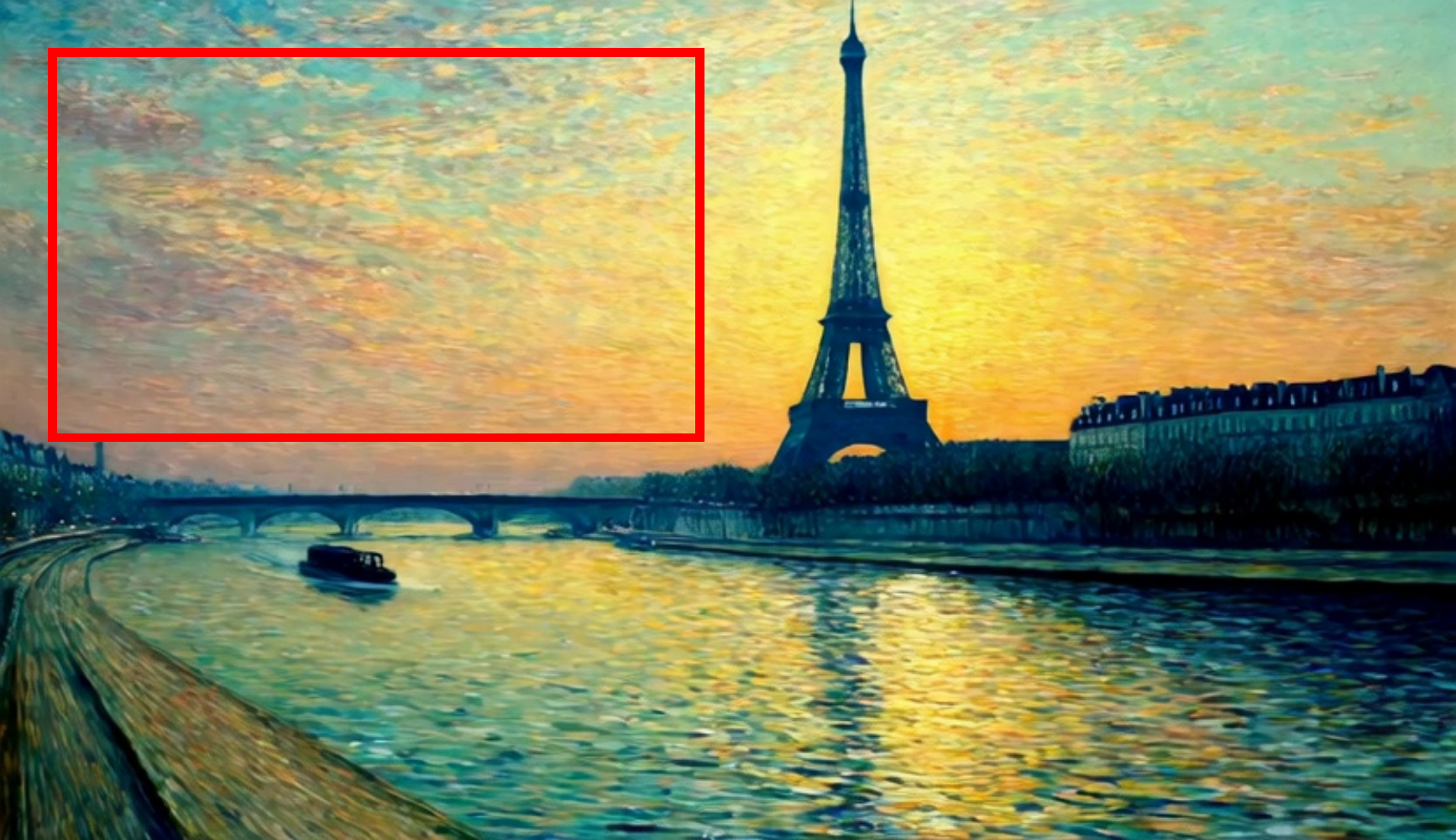}
            \captionsetup{skip=2.5pt}
        \end{subfigure}
        \hfill
        \captionsetup{skip=2.5pt}
        \begin{subfigure}{0.240\textwidth}
            \centering
            \includegraphics[width=\textwidth]{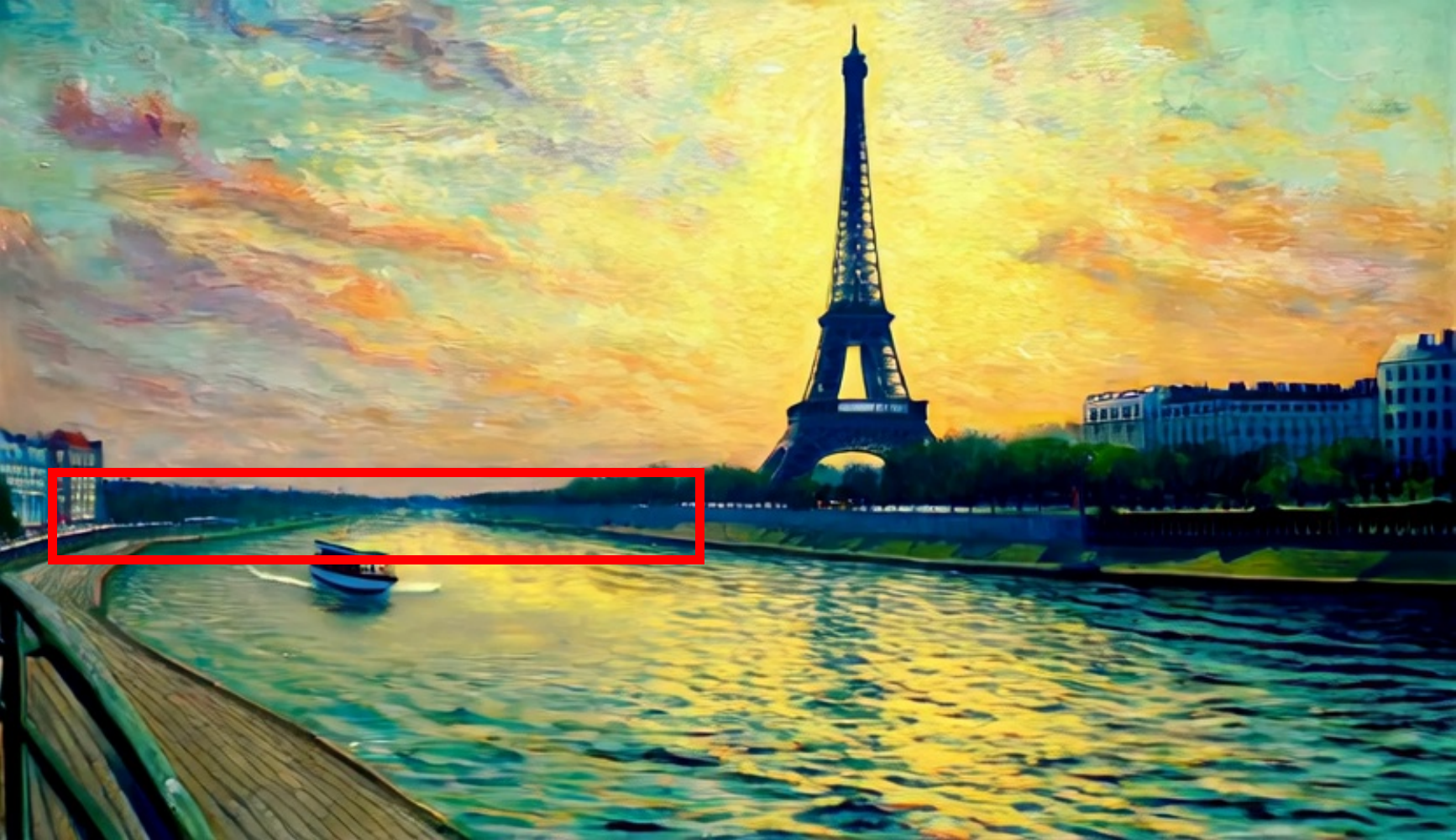}
            \captionsetup{skip=2.5pt}
        \end{subfigure}
        \hfill
        \captionsetup{skip=2.5pt}
        \begin{subfigure}{0.240\textwidth}
            \centering
            \includegraphics[width=\textwidth]{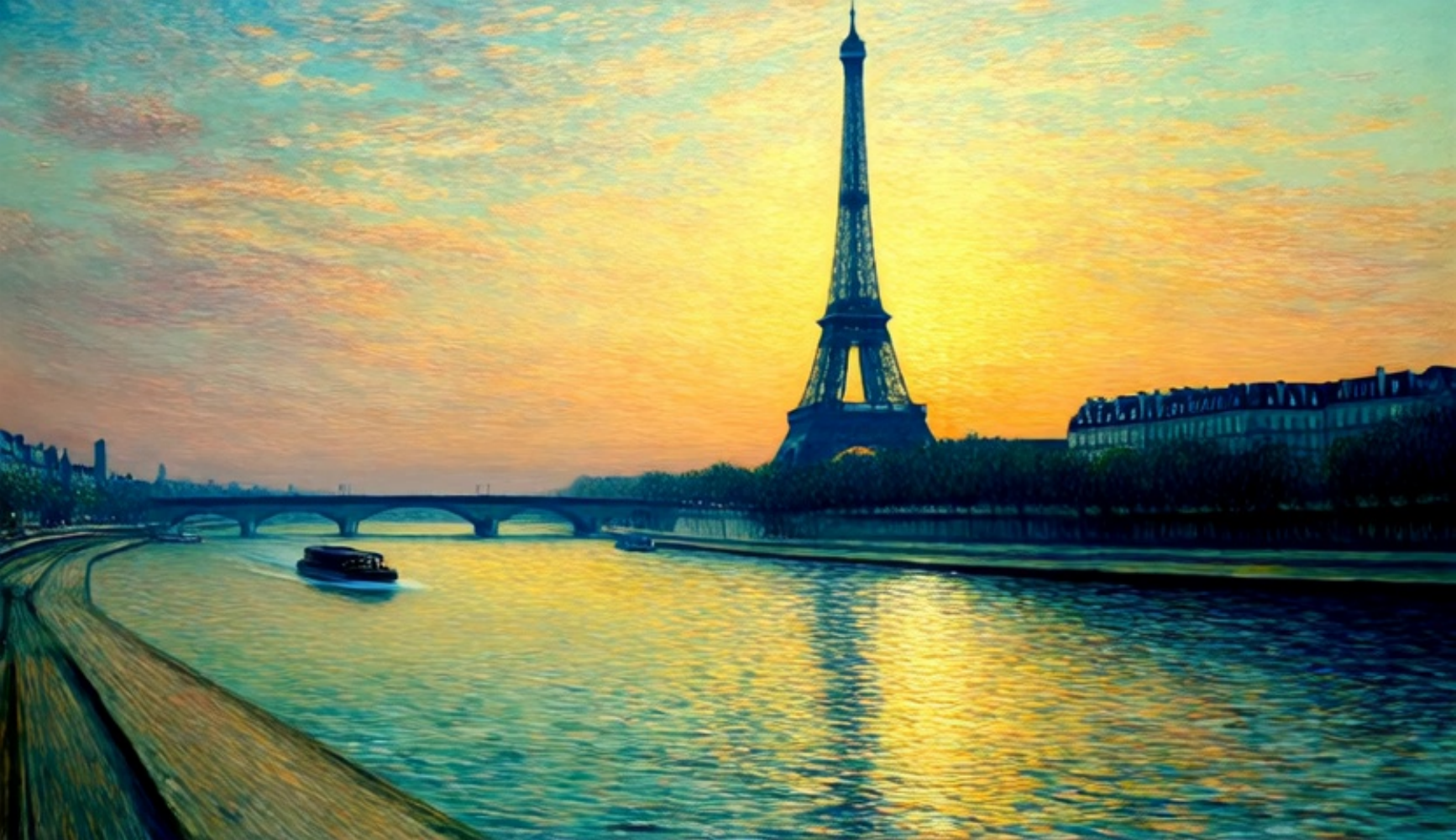}
            \captionsetup{skip=2.5pt}
        \end{subfigure}
    \end{subfigure}%
    \\
    \begin{subfigure}{\textwidth}
        \centering
        \begin{subfigure}{0.240\textwidth}
            \centering
            \includegraphics[width=\textwidth]{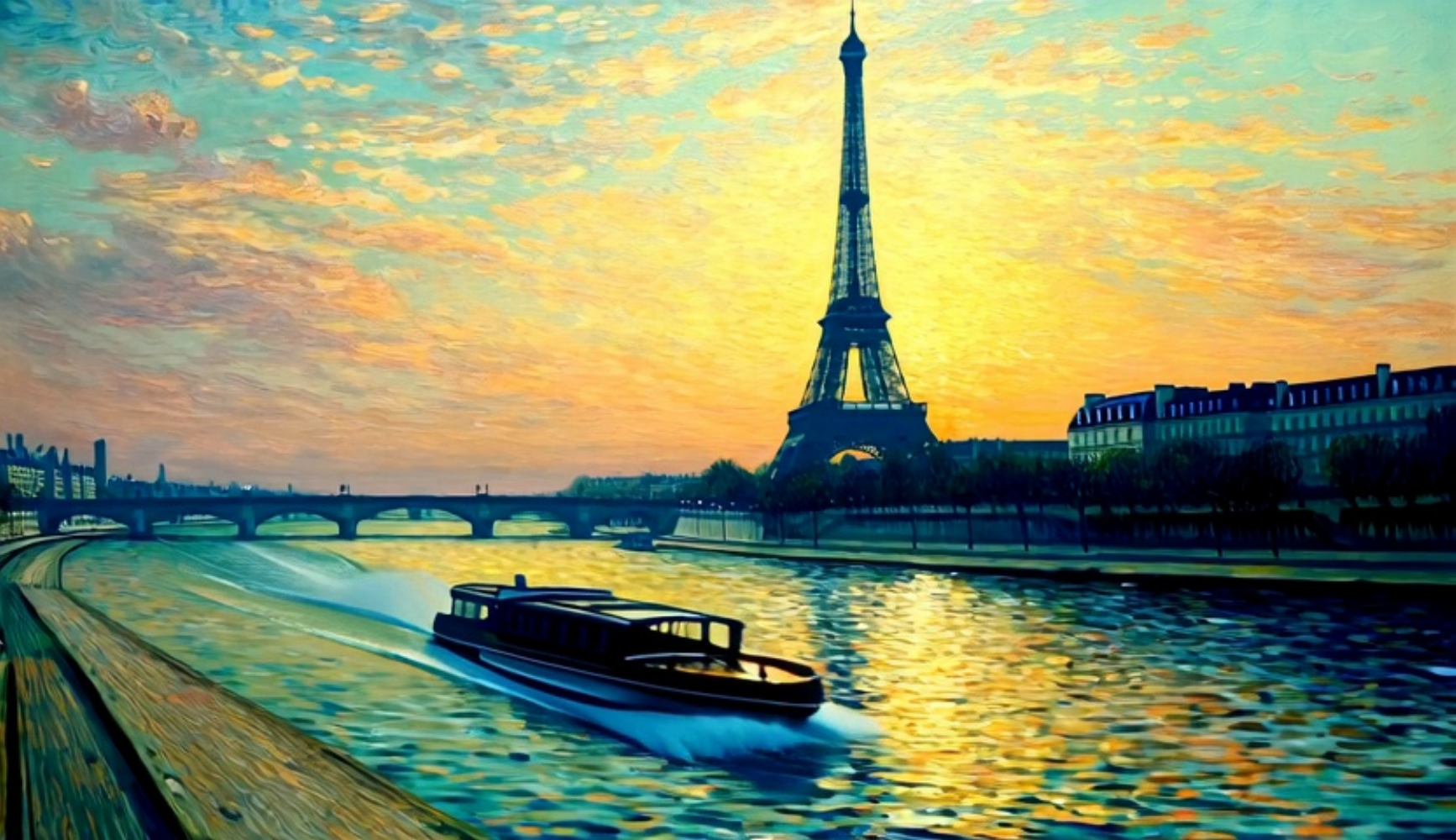}
            \captionsetup{skip=2.5pt}
        \end{subfigure}%
        \hfill
        \captionsetup{skip=2.5pt}
        \begin{subfigure}{0.240\textwidth}
            \centering
            \includegraphics[width=\textwidth]{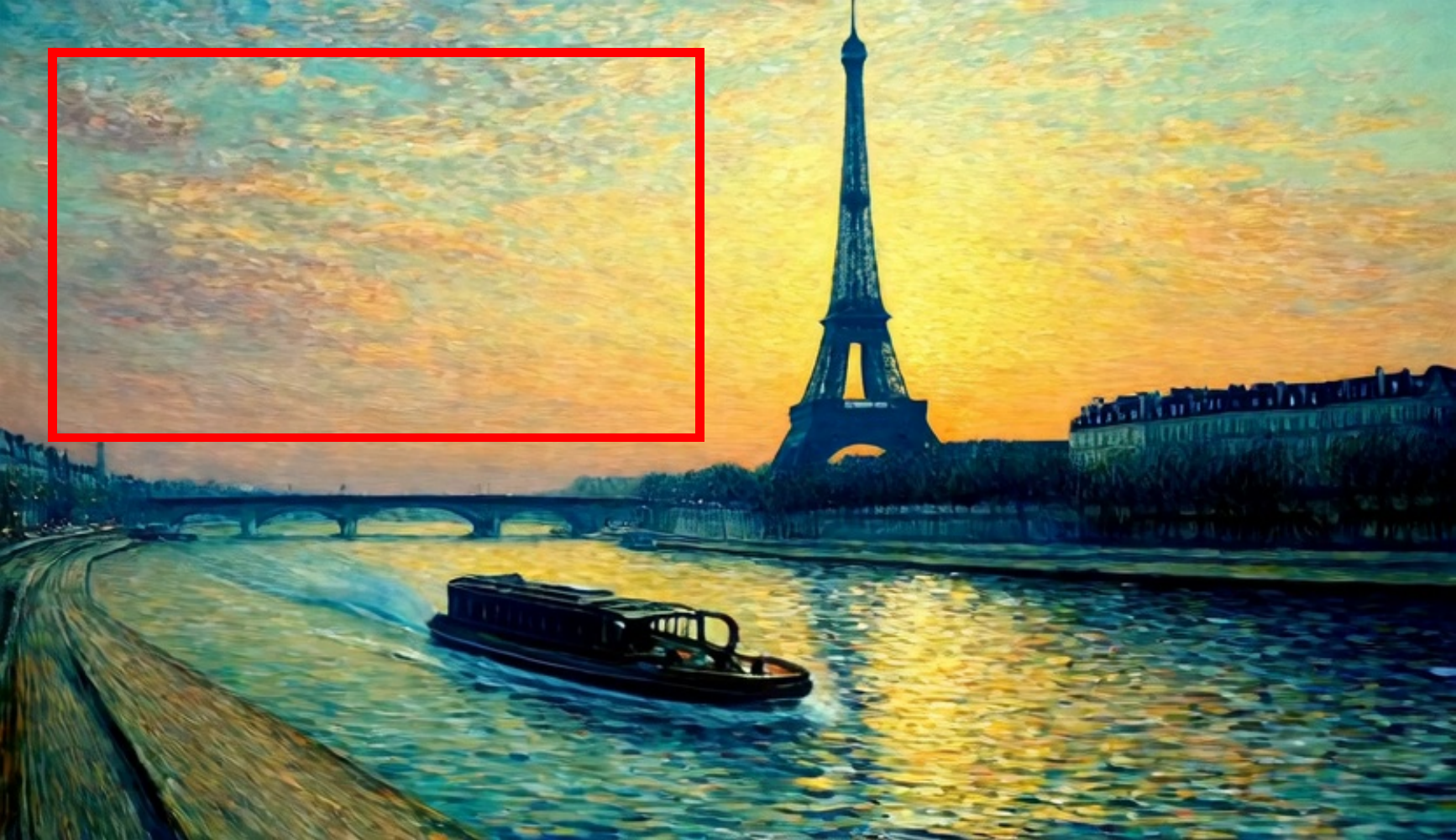}
            \captionsetup{skip=2.5pt}
        \end{subfigure}
        \hfill
        \captionsetup{skip=2.5pt}
        \begin{subfigure}{0.240\textwidth}
            \centering
            \includegraphics[width=\textwidth]{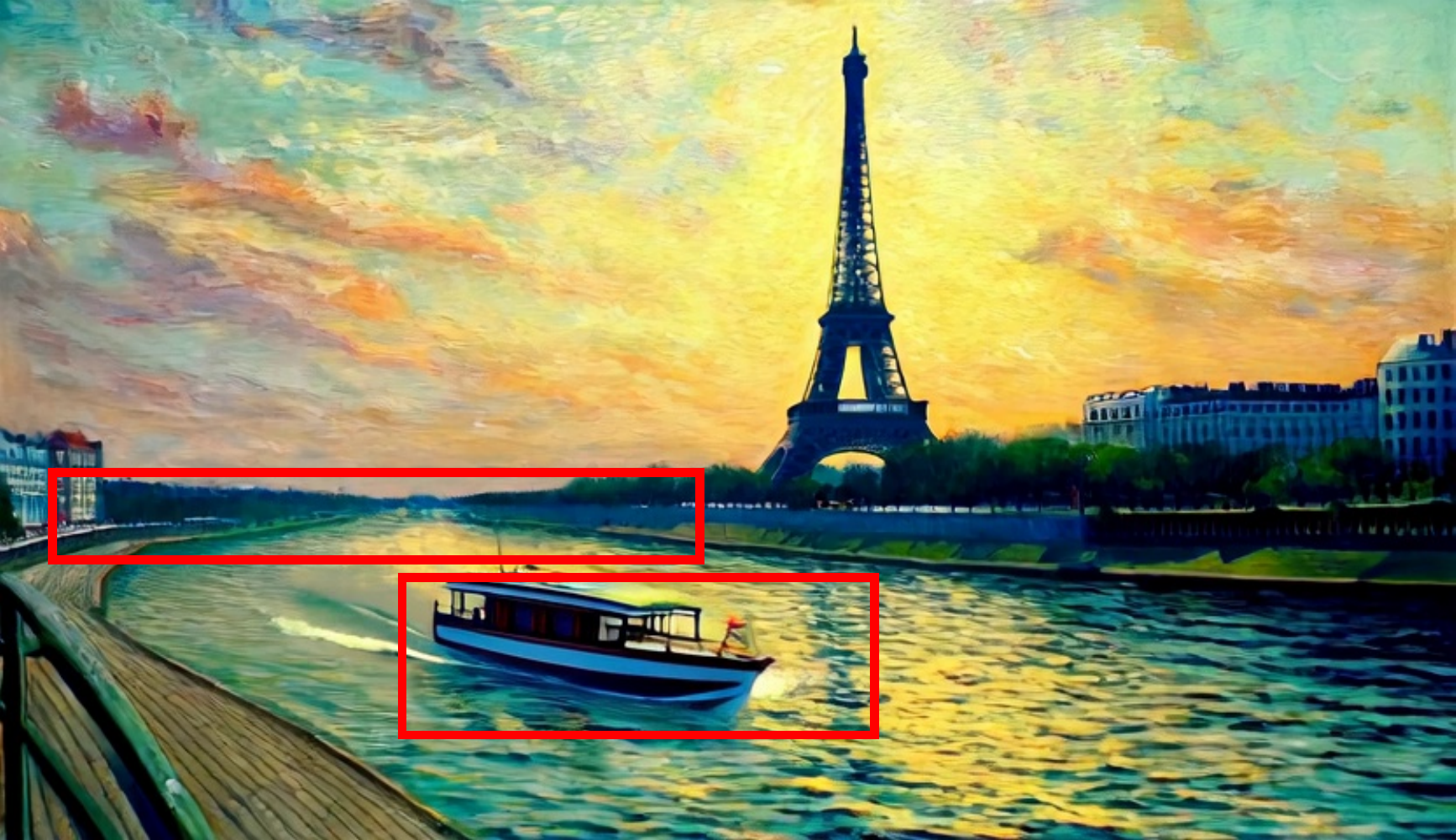}
            \captionsetup{skip=2.5pt}
        \end{subfigure}
        \hfill
        \captionsetup{skip=2.5pt}
        \begin{subfigure}{0.240\textwidth}
            \centering
            \includegraphics[width=\textwidth]{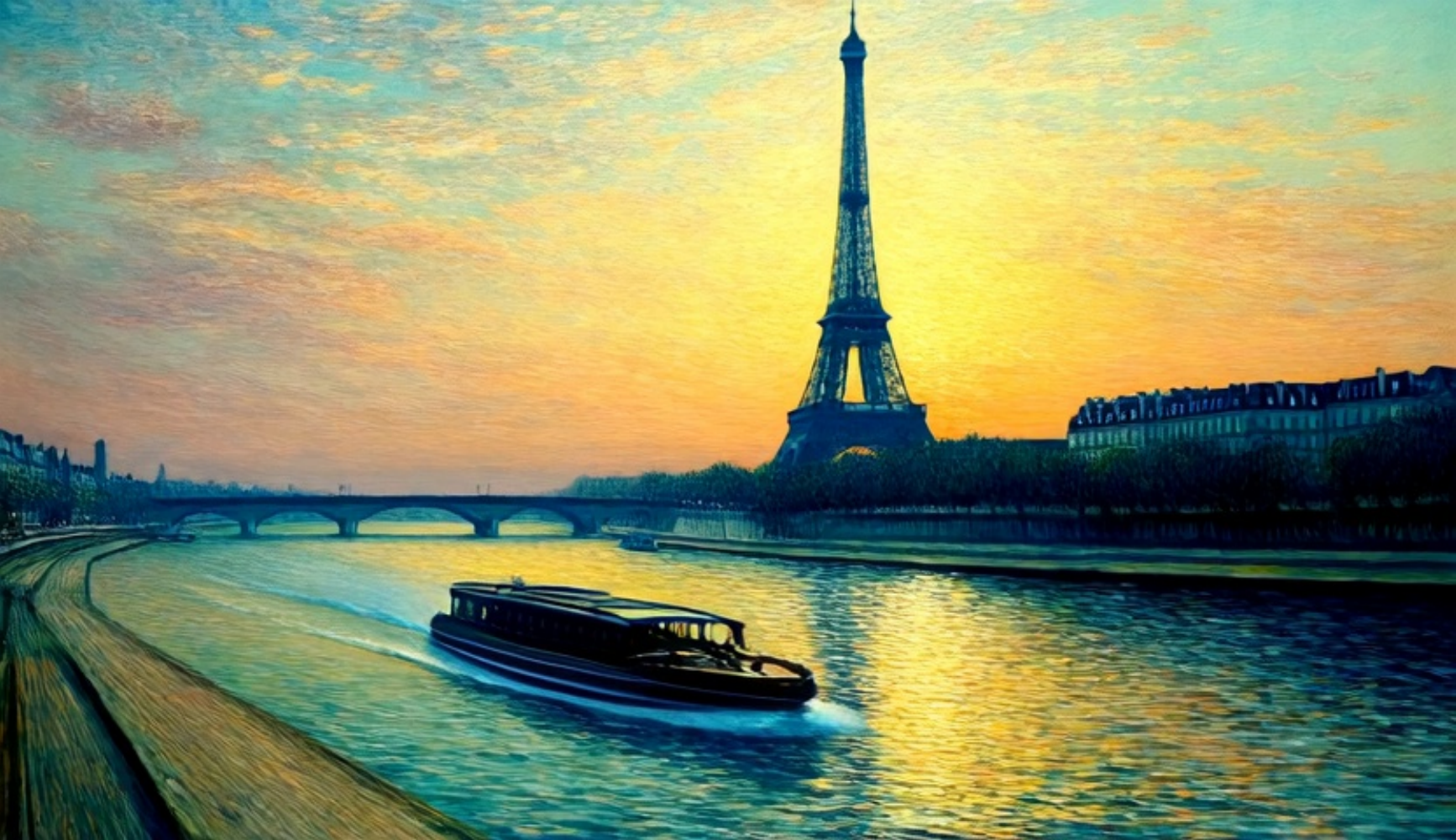}
            \captionsetup{skip=2.5pt}
        \end{subfigure}
    \end{subfigure}%
    \\
    \begin{subfigure}{\textwidth}
        \centering
        % \caption*{\small{Prompt: ``The bund Shanghai by Hokusai, in the style of Ukiyo.''}}
        \begin{subfigure}{0.240\textwidth}
            \centering
            \includegraphics[width=\textwidth]{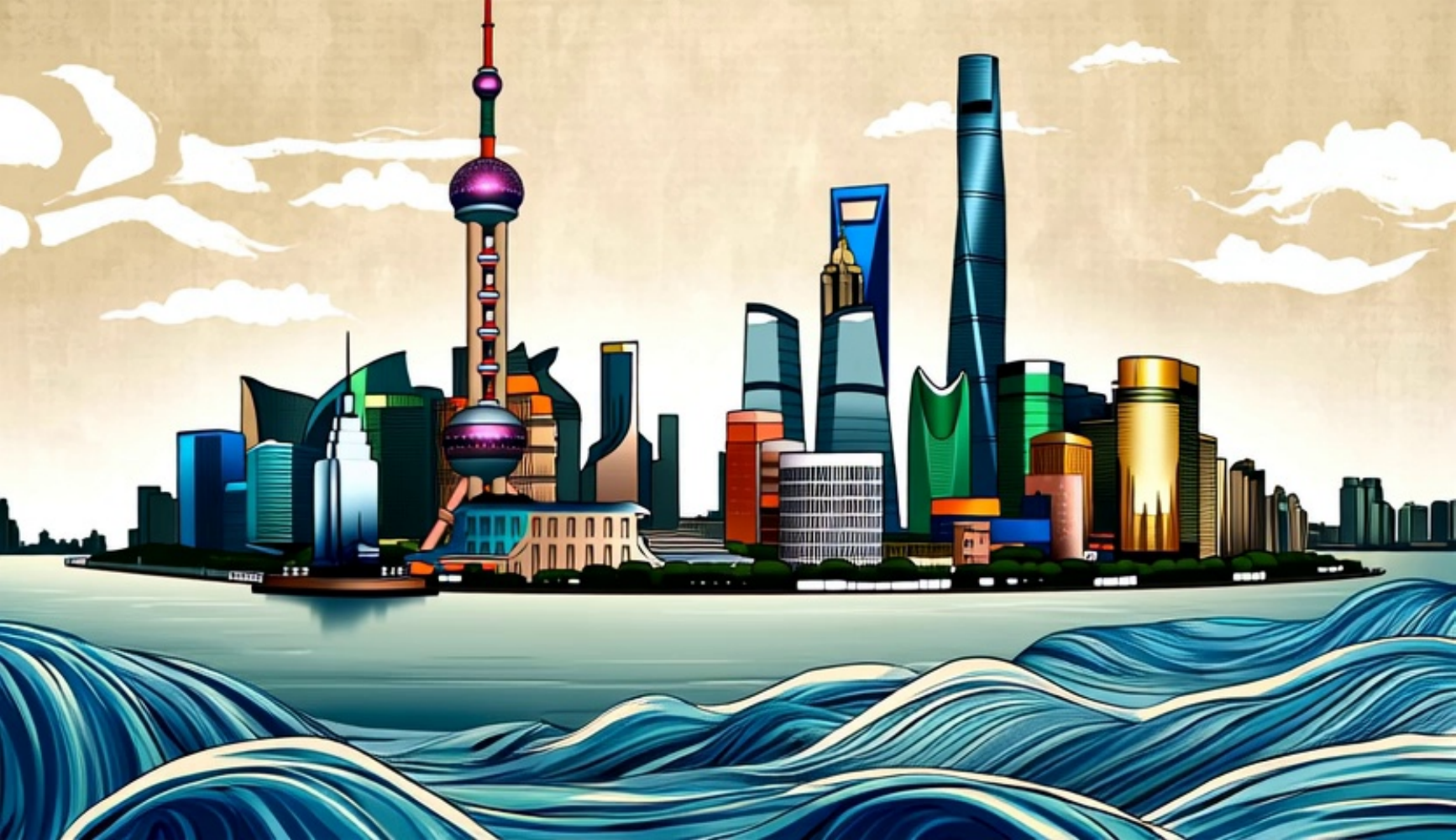}
            % \captionsetup{skip=2.5pt}
        \end{subfigure}%
        \hfill
        \captionsetup{skip=2.5pt}
        \begin{subfigure}{0.240\textwidth}
            \centering
            \includegraphics[width=\textwidth]{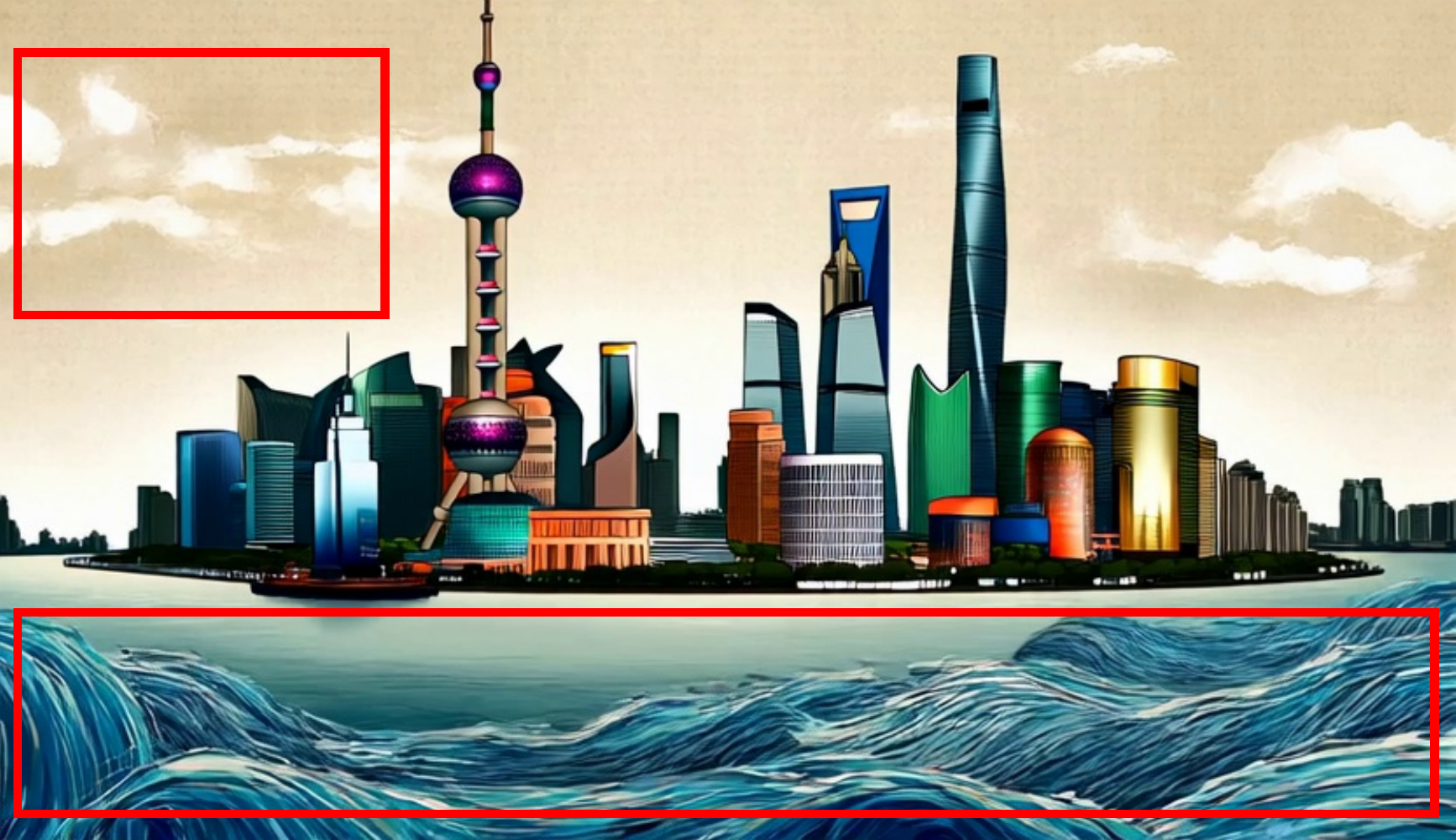}
            % \captionsetup{skip=2.5pt}
        \end{subfigure}
        \hfill
        \captionsetup{skip=2.5pt}
        \begin{subfigure}{0.240\textwidth}
            \centering
            \includegraphics[width=\textwidth]{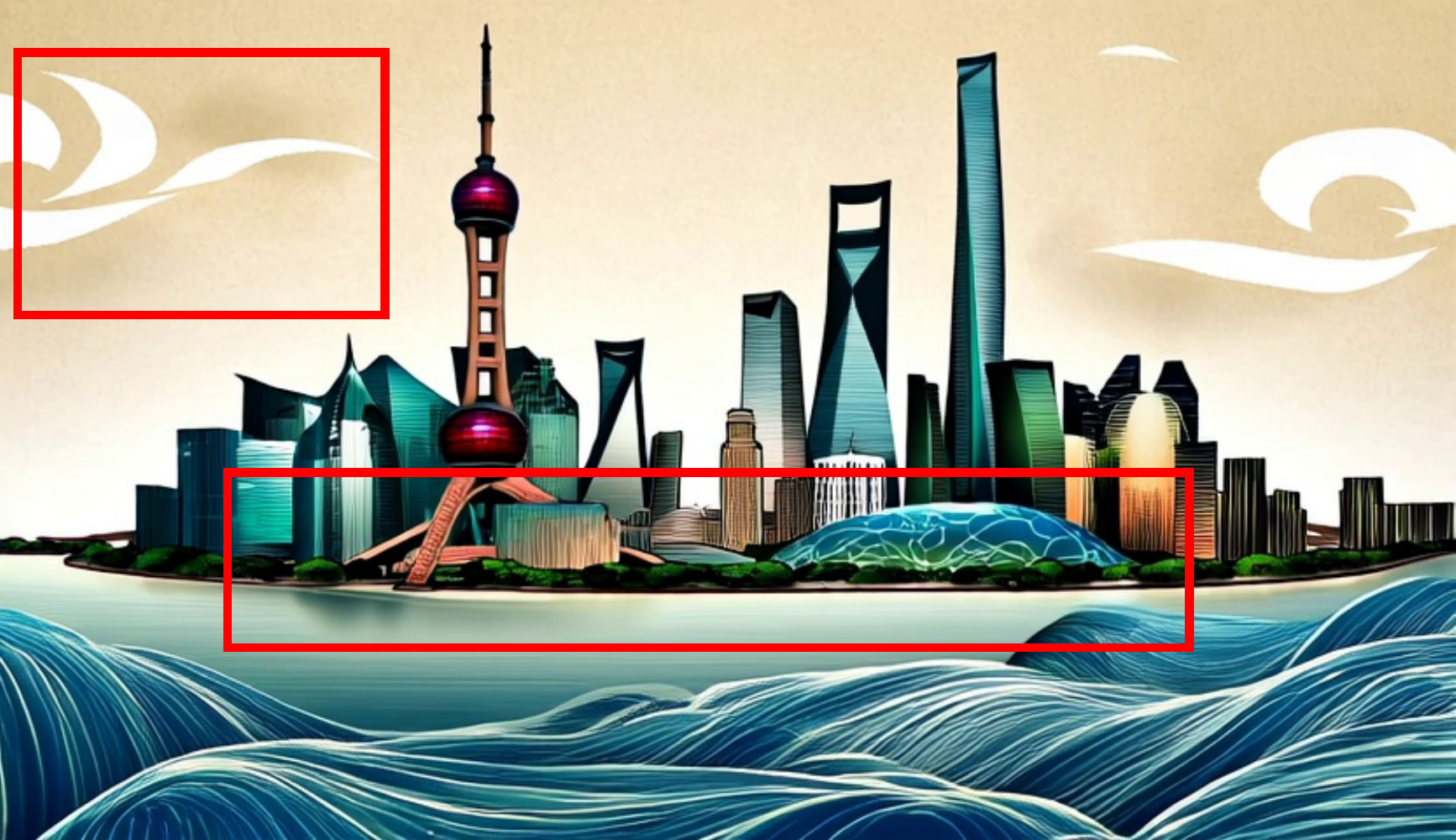}
            % \captionsetup{skip=2.5pt}
        \end{subfigure}
        \hfill
        \captionsetup{skip=2.5pt}
        \begin{subfigure}{0.240\textwidth}
            \centering
            \includegraphics[width=\textwidth]{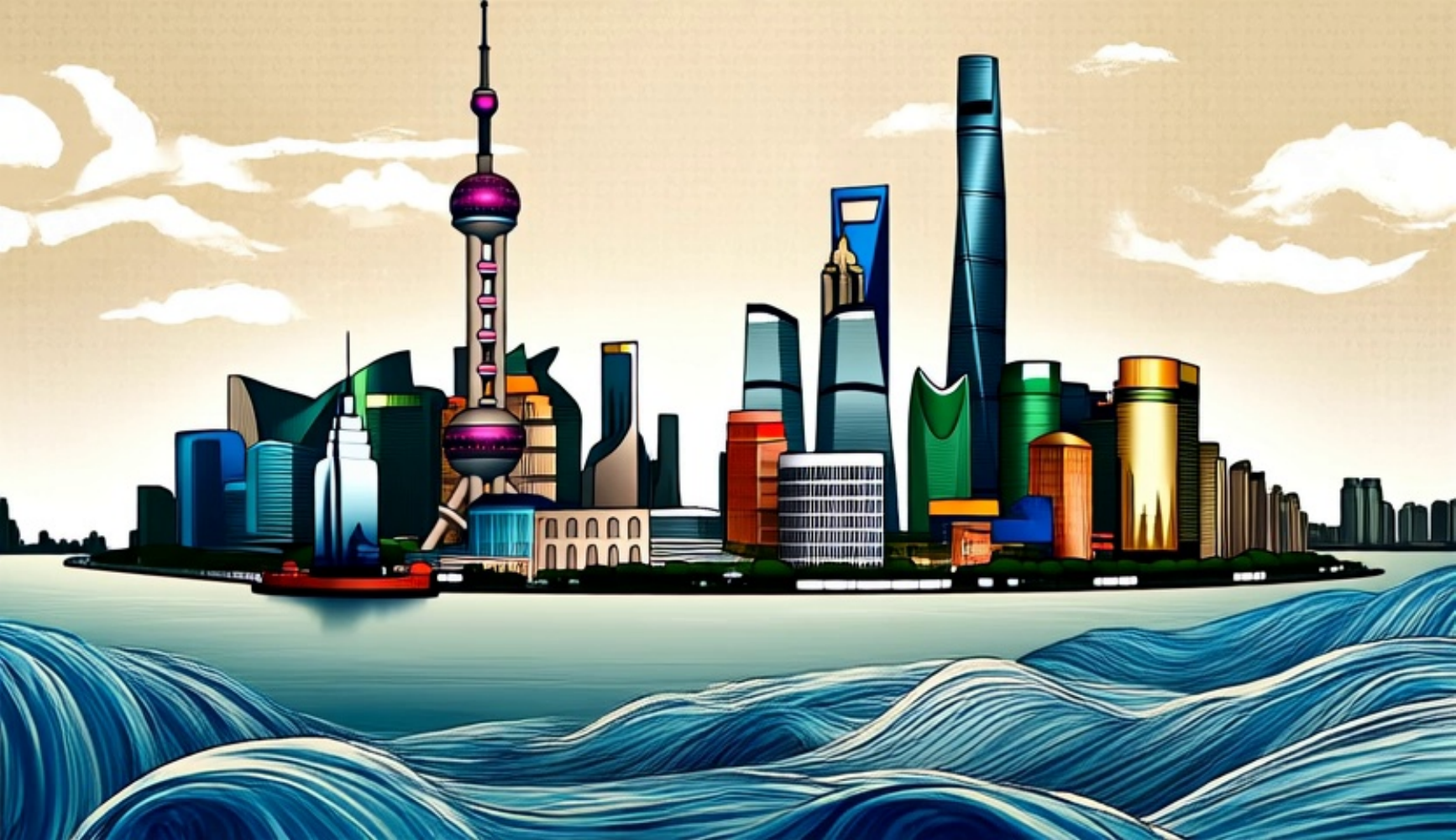}
            % \captionsetup{skip=2.5pt}
        \end{subfigure}
    \end{subfigure}%
    \\
    \begin{subfigure}{\textwidth}
        \centering
        \begin{subfigure}{0.240\textwidth}
            \centering
            \includegraphics[width=\textwidth]{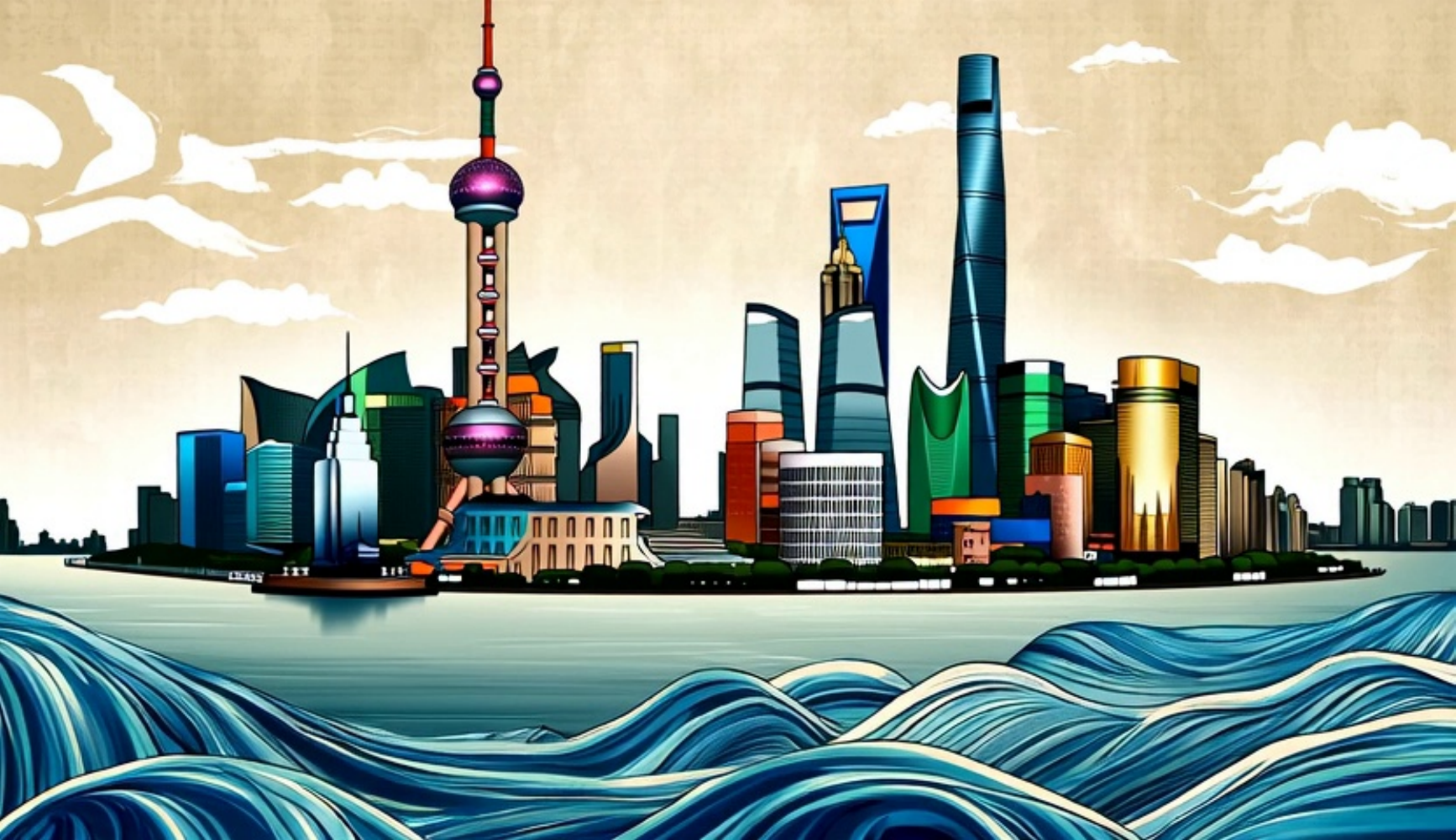}
        \end{subfigure}%
        \hfill
        \begin{subfigure}{0.240\textwidth}
            \centering
            \includegraphics[width=\textwidth]{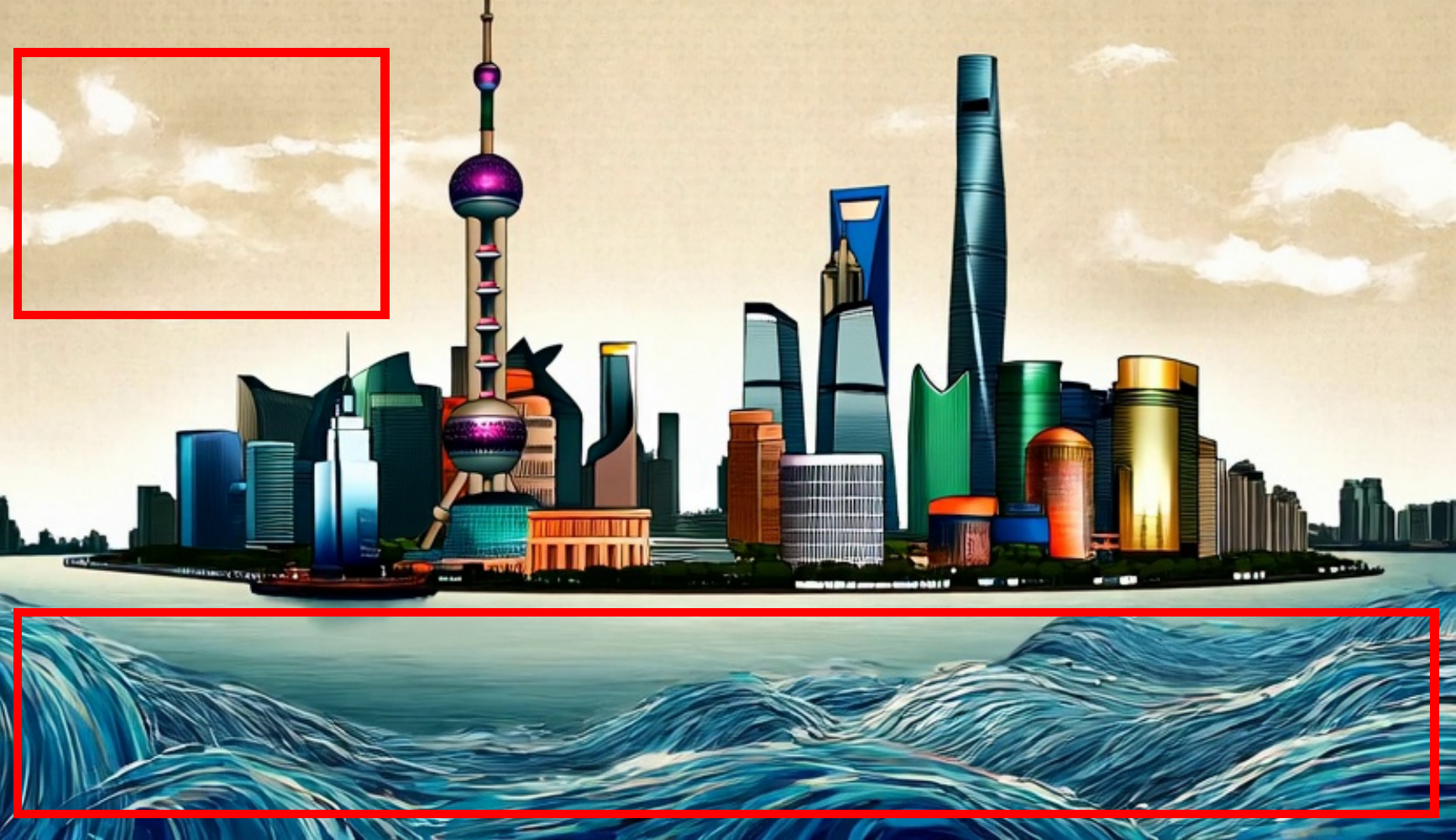}
        \end{subfigure}
        \hfill
        \begin{subfigure}{0.240\textwidth}
            \centering
            \includegraphics[width=\textwidth]{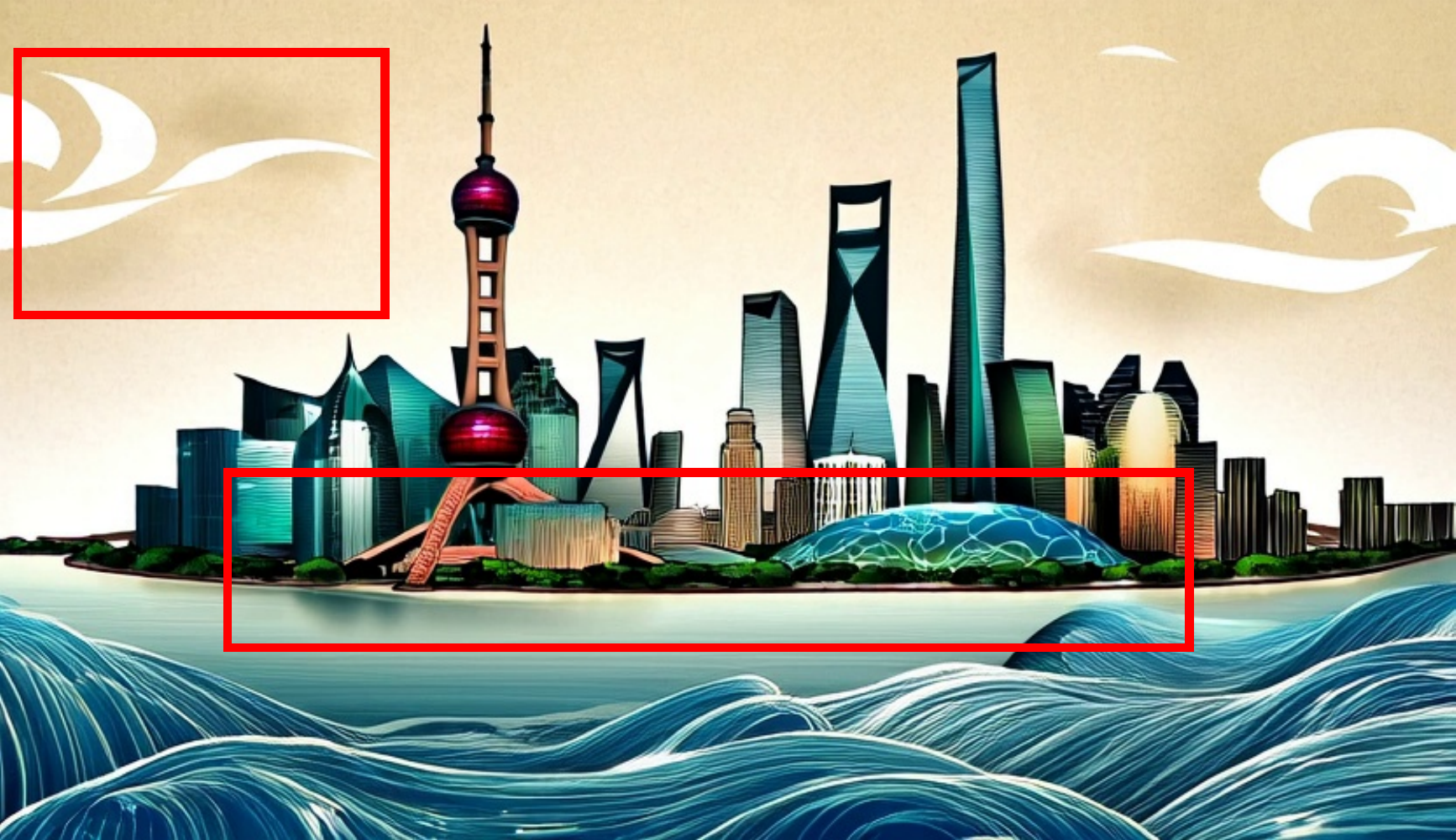}
        \end{subfigure}
        \hfill
        \begin{subfigure}{0.240\textwidth}
            \centering
            \includegraphics[width=\textwidth]{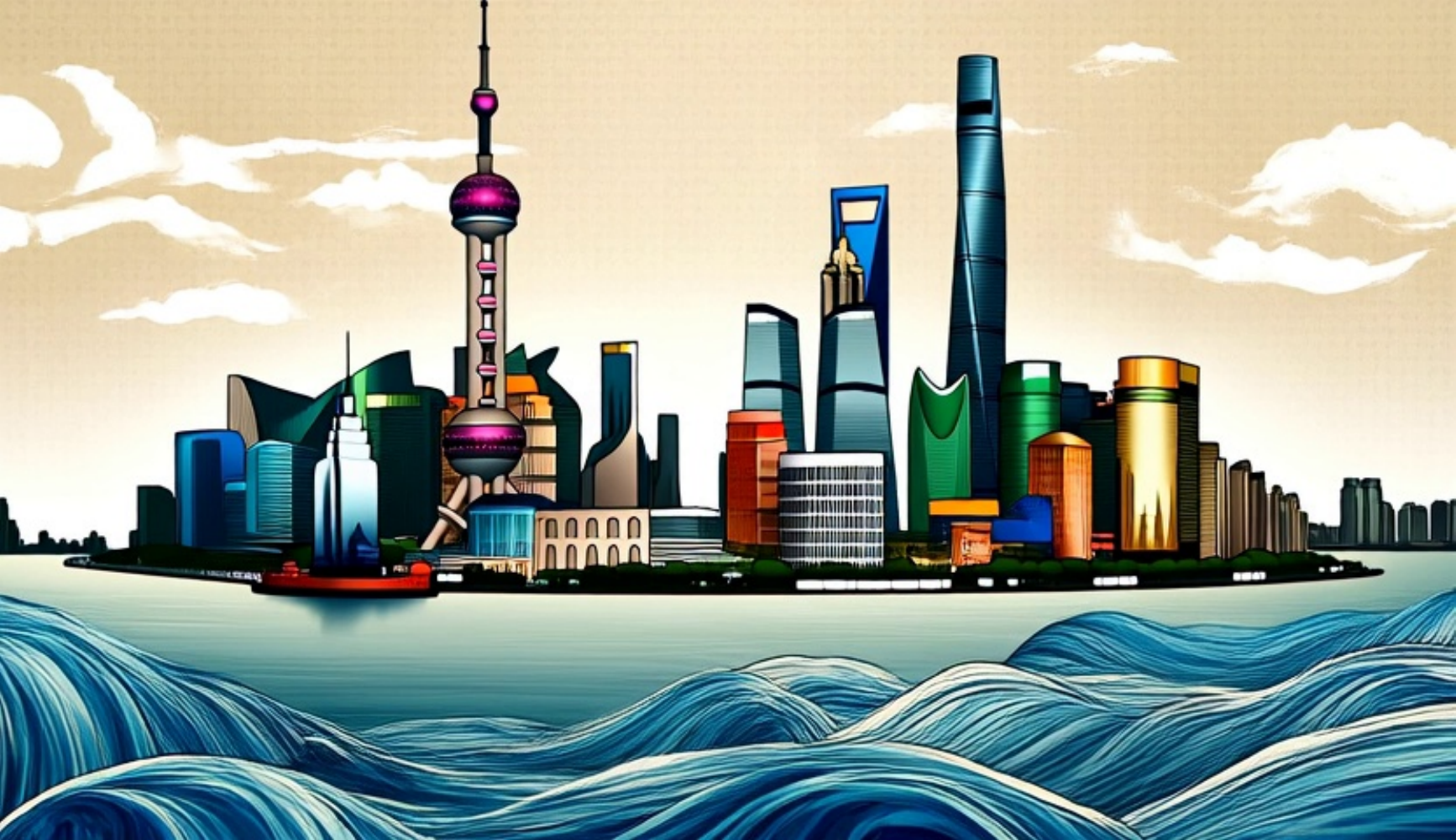}
        \end{subfigure}
    \end{subfigure}%
    \\
    \begin{subfigure}{\textwidth}
        \centering
        % \caption*{\small{Prompt: ``The bund Shanghai by Hokusai, in the style of Ukiyo.''}}
        \begin{subfigure}{0.240\textwidth}
            \centering
            \includegraphics[width=\textwidth]{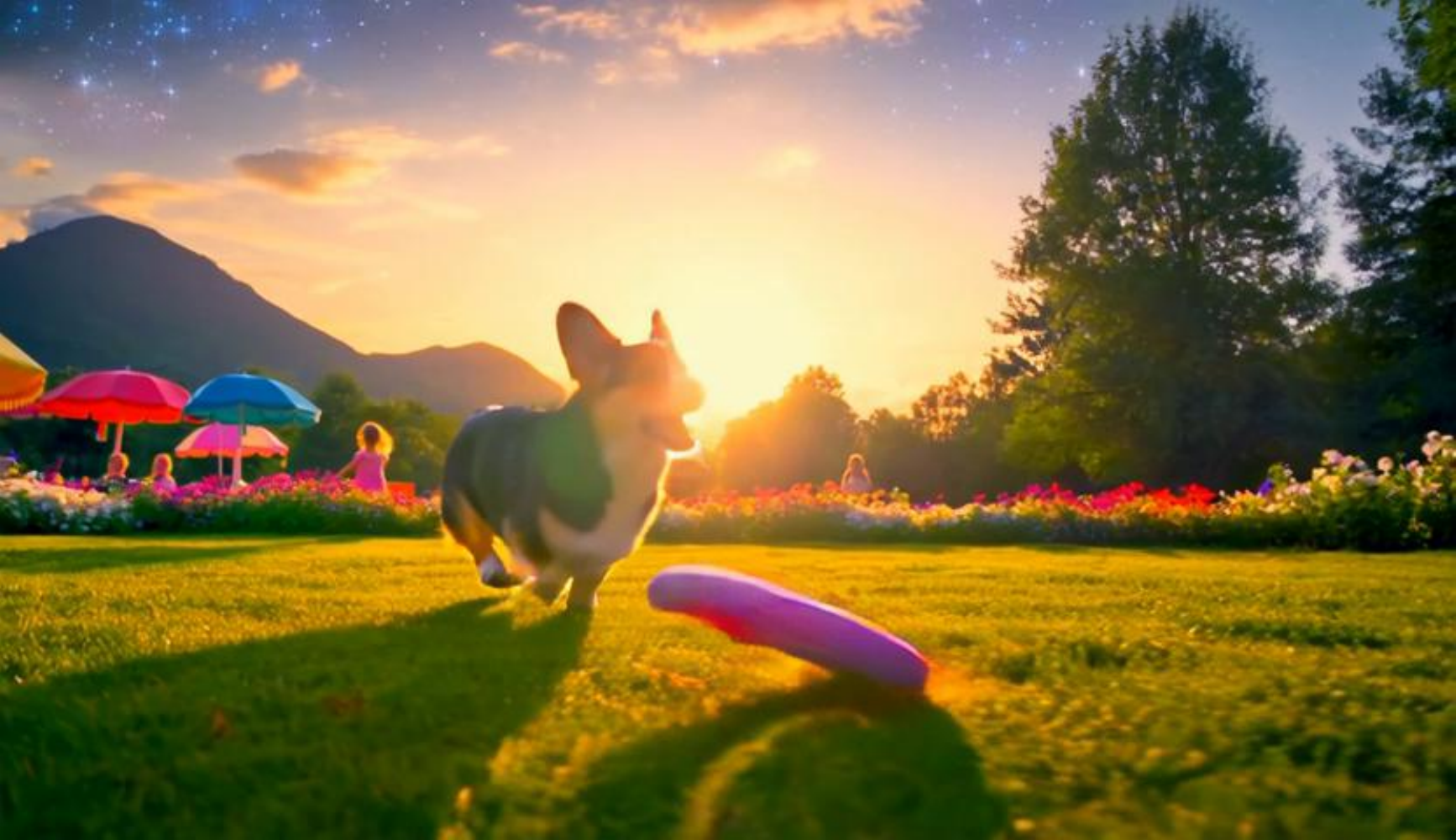}
            % \captionsetup{skip=2.5pt}
        \end{subfigure}%
        \hfill
        \captionsetup{skip=2.5pt}
        \begin{subfigure}{0.240\textwidth}
            \centering
            \includegraphics[width=\textwidth]{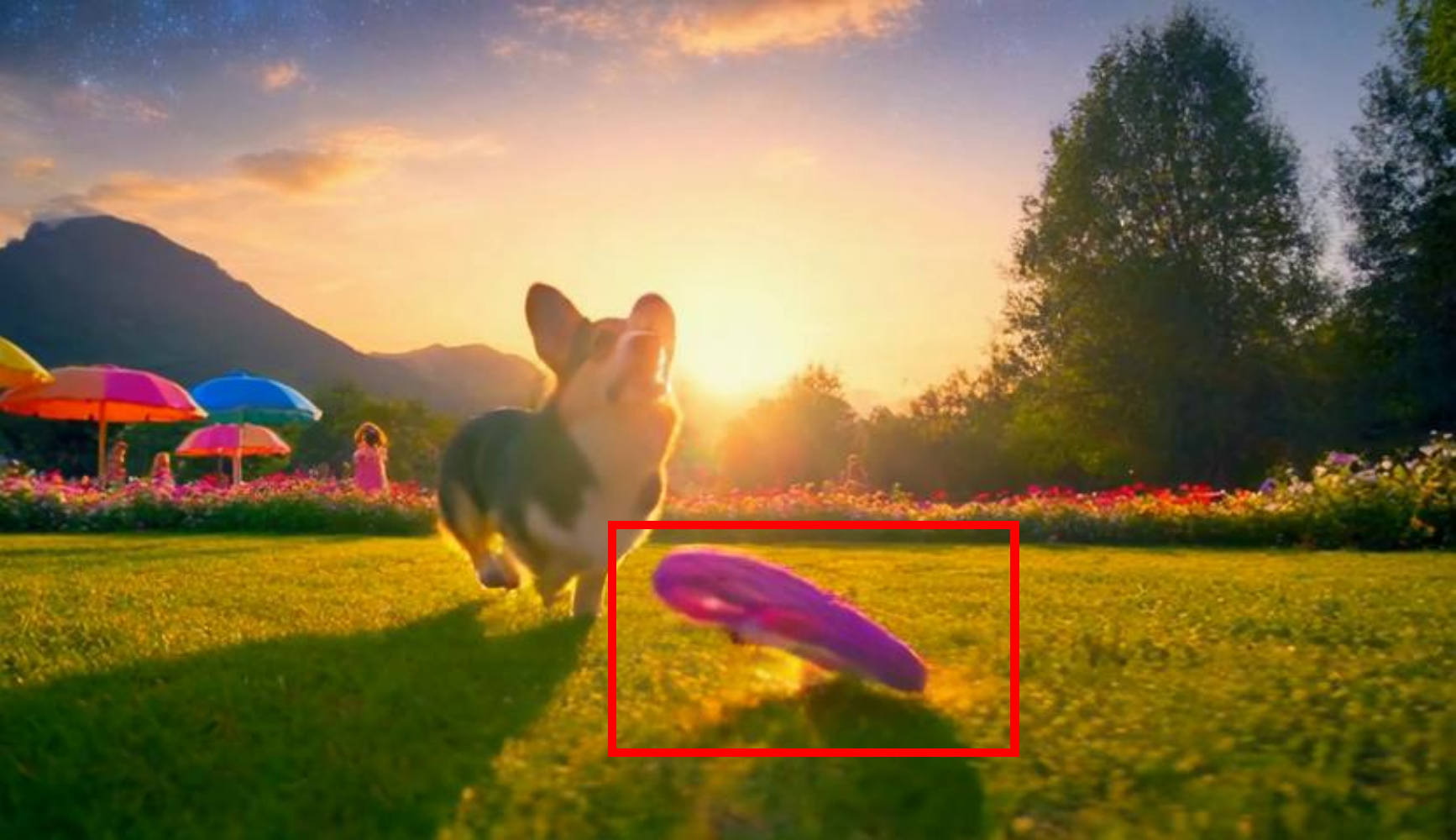}
            % \captionsetup{skip=2.5pt}
        \end{subfigure}
        \hfill
        \captionsetup{skip=2.5pt}
        \begin{subfigure}{0.240\textwidth}
            \centering
            \includegraphics[width=\textwidth]{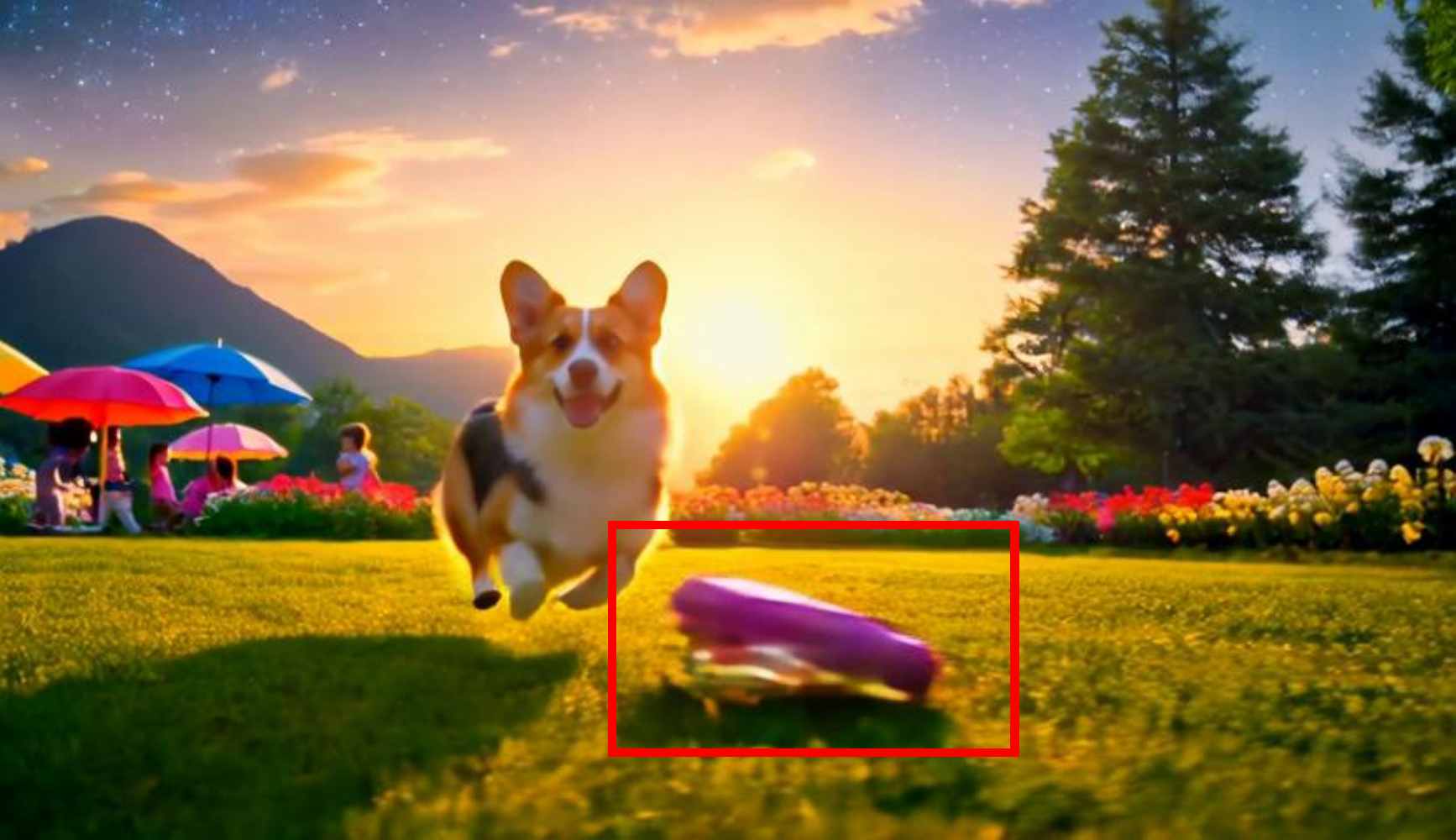}
            % \captionsetup{skip=2.5pt}
        \end{subfigure}
        \hfill
        \captionsetup{skip=2.5pt}
        \begin{subfigure}{0.240\textwidth}
            \centering
            \includegraphics[width=\textwidth]{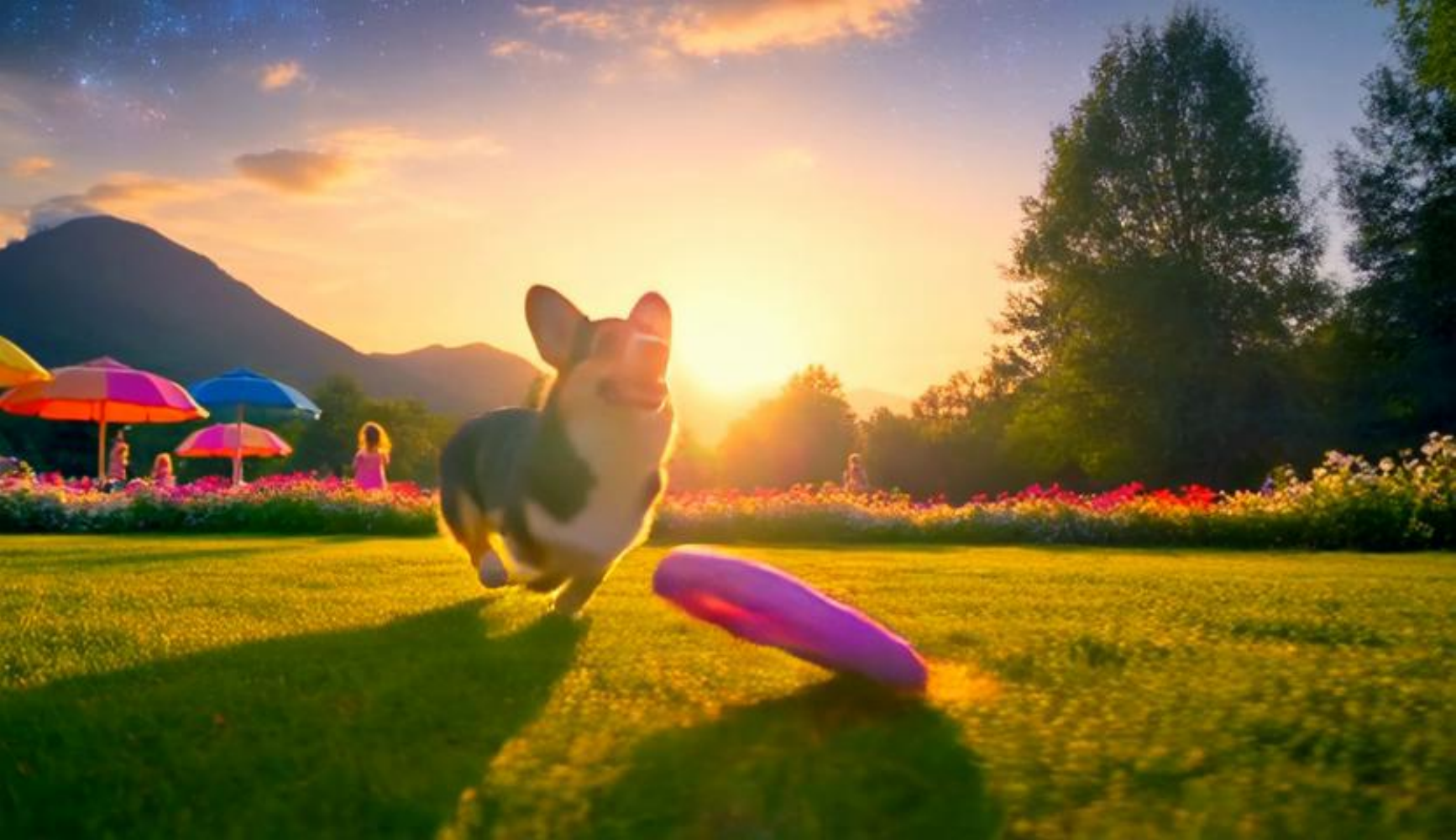}
            % \captionsetup{skip=2.5pt}
        \end{subfigure}
    \end{subfigure}%
    \\
    \begin{subfigure}{\textwidth}
        \centering
        \begin{subfigure}{0.240\textwidth}
            \centering
            \includegraphics[width=\textwidth]{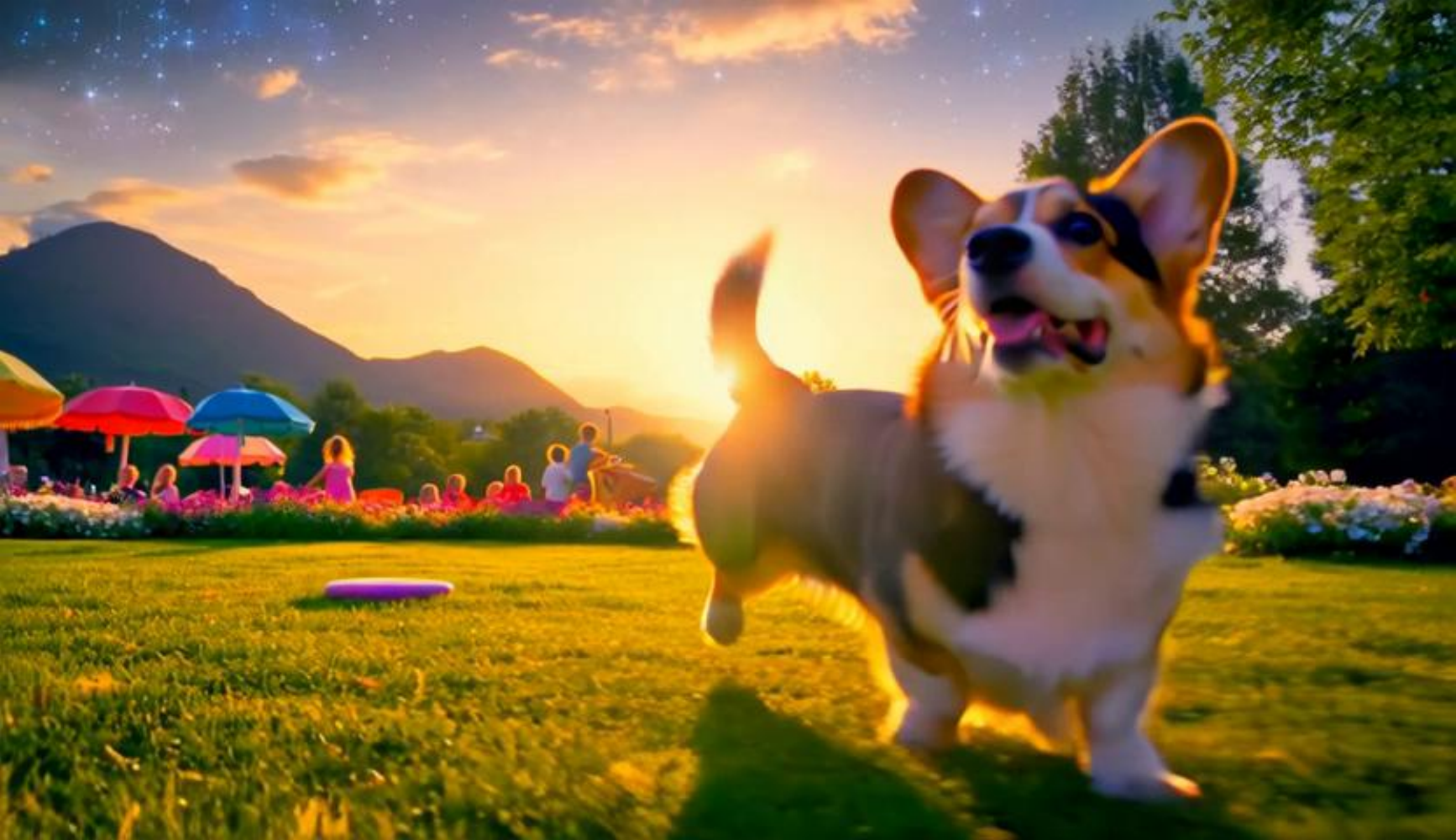}
        \end{subfigure}%
        \hfill
        \begin{subfigure}{0.240\textwidth}
            \centering
            \includegraphics[width=\textwidth]{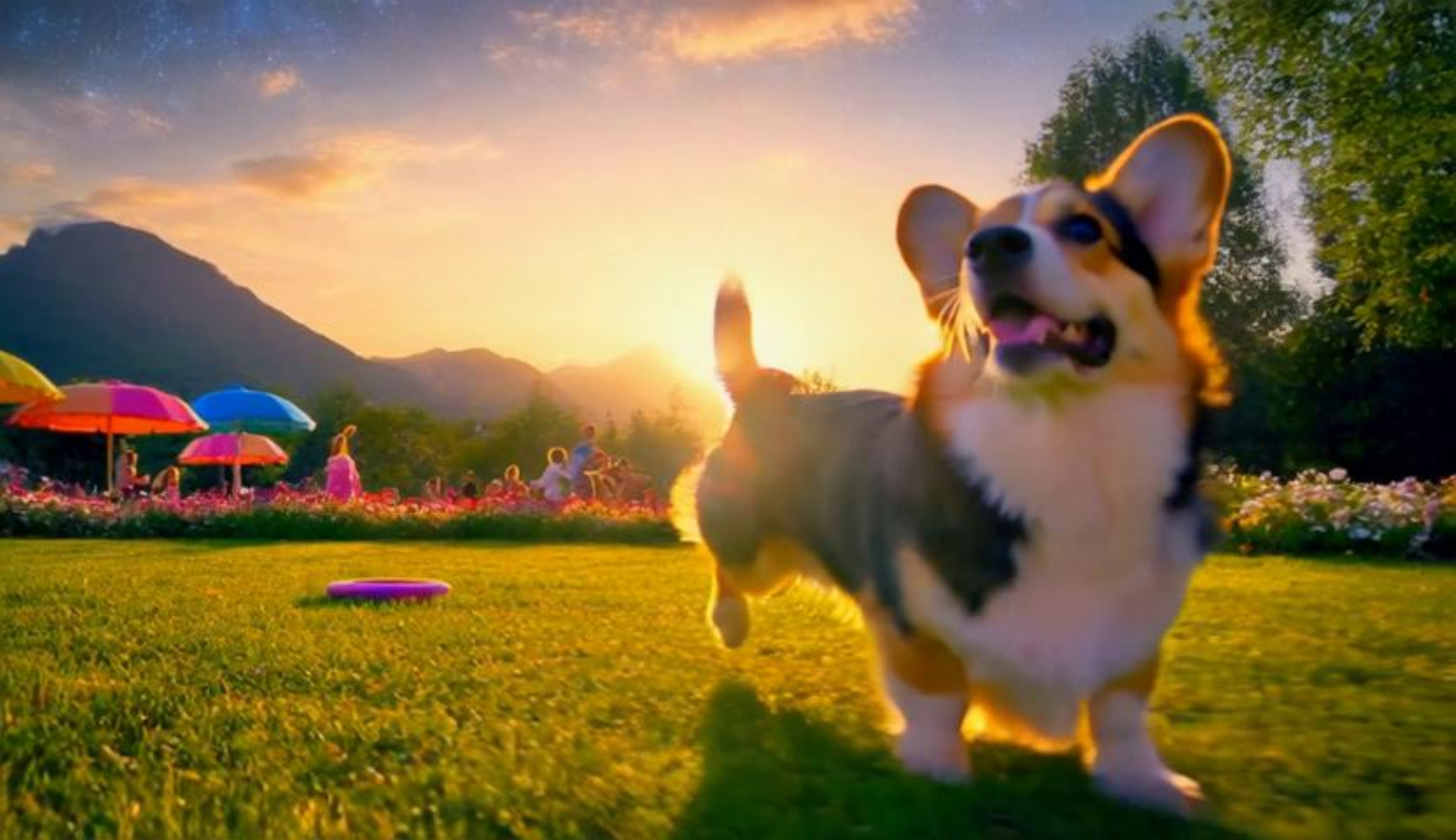}
        \end{subfigure}
        \hfill
        \begin{subfigure}{0.240\textwidth}
            \centering
            \includegraphics[width=\textwidth]{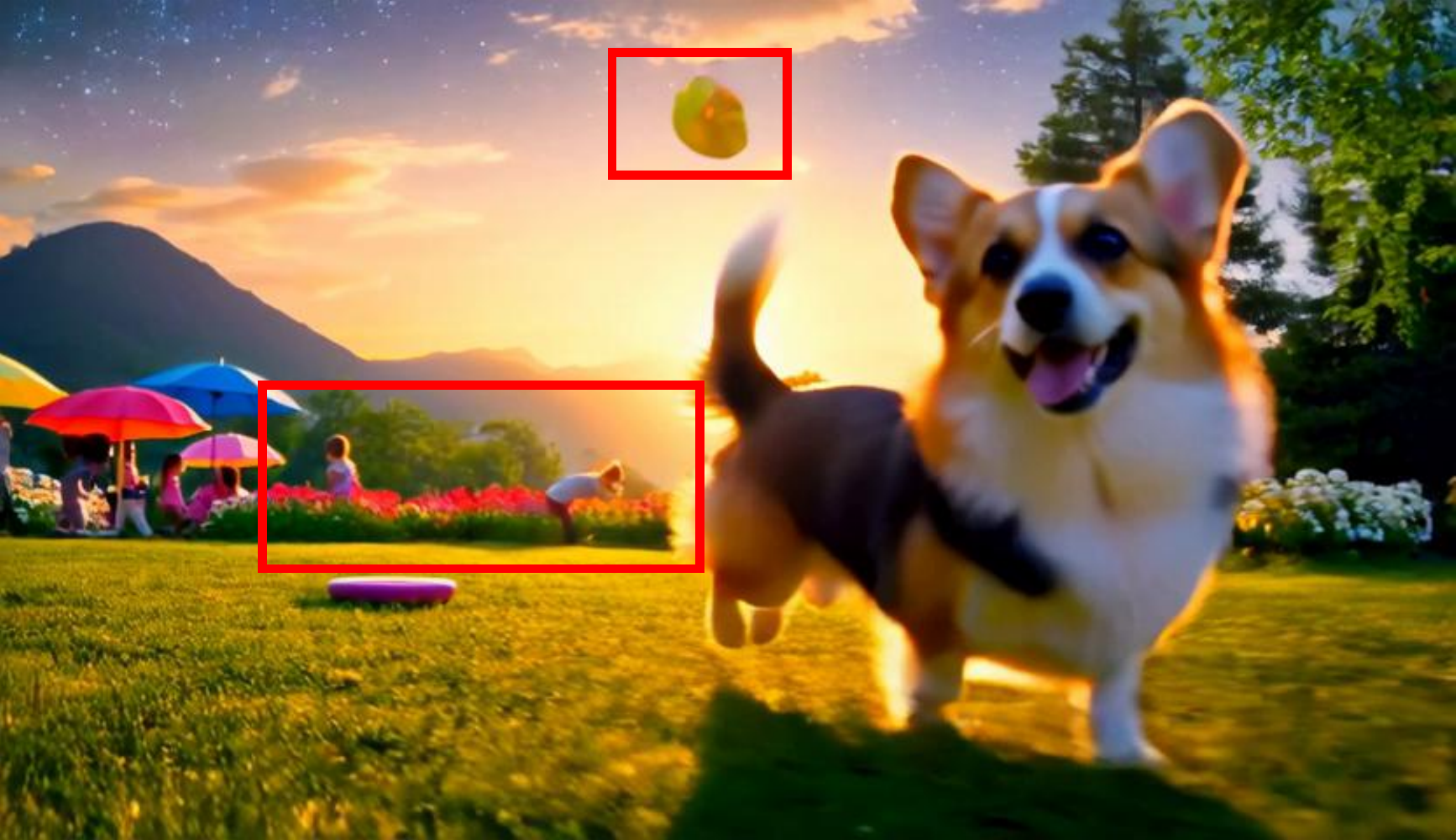}
        \end{subfigure}
        \hfill
        \begin{subfigure}{0.240\textwidth}
            \centering
            \includegraphics[width=\textwidth]{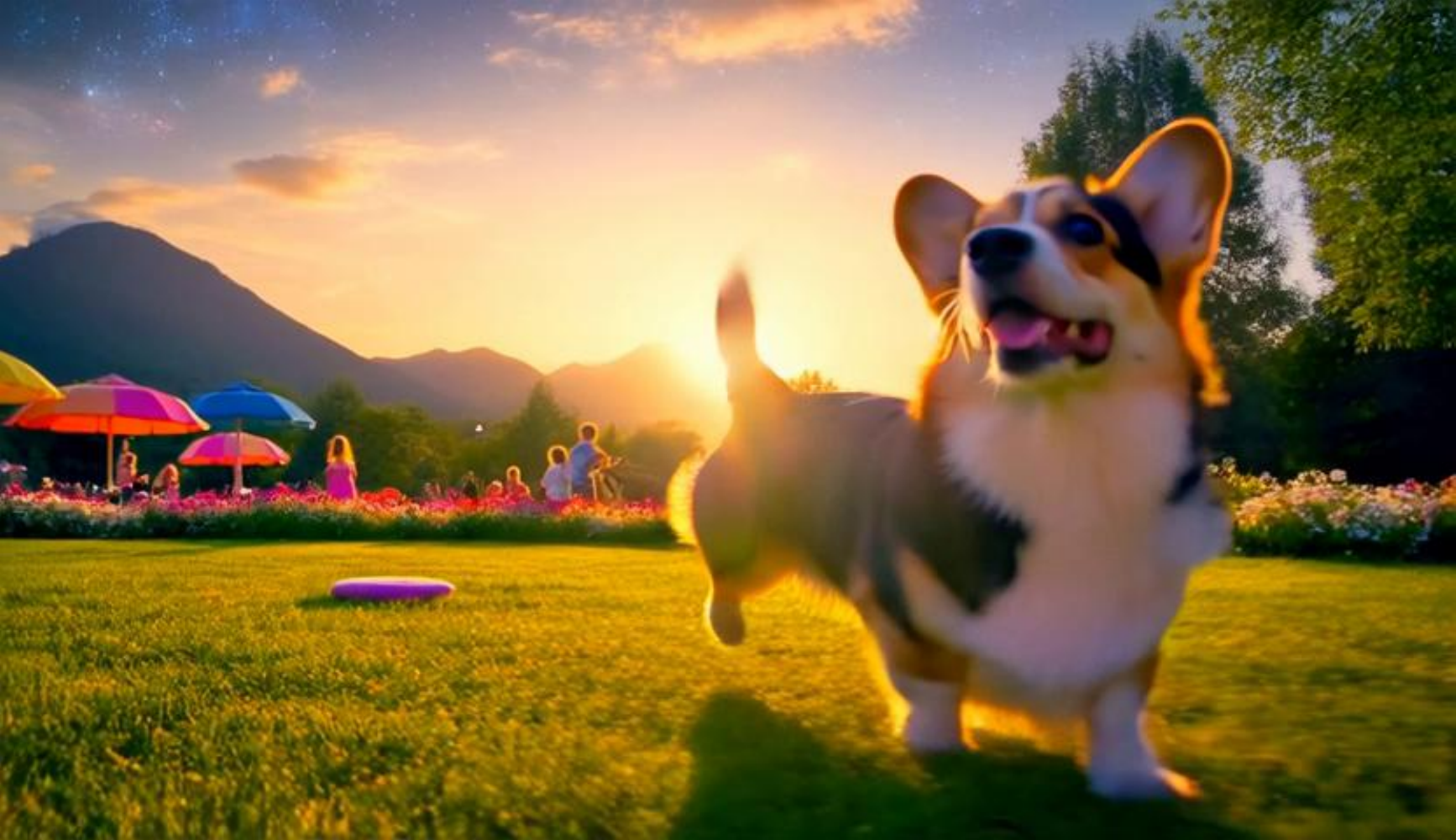}
        \end{subfigure}
    \end{subfigure}%
    \\
    \begin{subfigure}{\textwidth}
        \centering
        % \caption*{\small{Prompt: ``The bund Shanghai by Hokusai, in the style of Ukiyo.''}}
        \begin{subfigure}{0.240\textwidth}
            \centering
            \includegraphics[width=\textwidth]{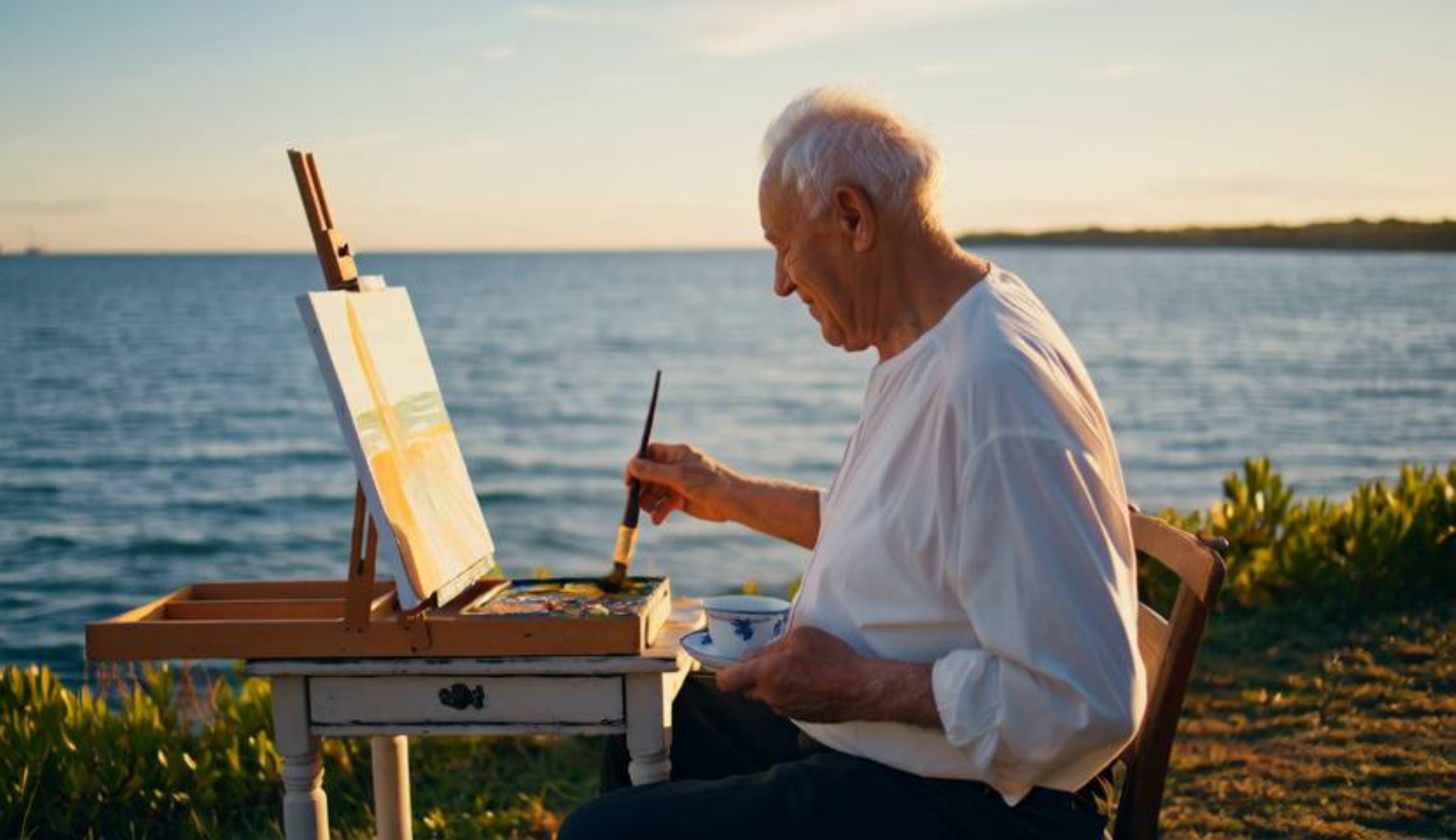}
            % \captionsetup{skip=2.5pt}
        \end{subfigure}%
        \hfill
        \captionsetup{skip=2.5pt}
        \begin{subfigure}{0.240\textwidth}
            \centering
            \includegraphics[width=\textwidth]{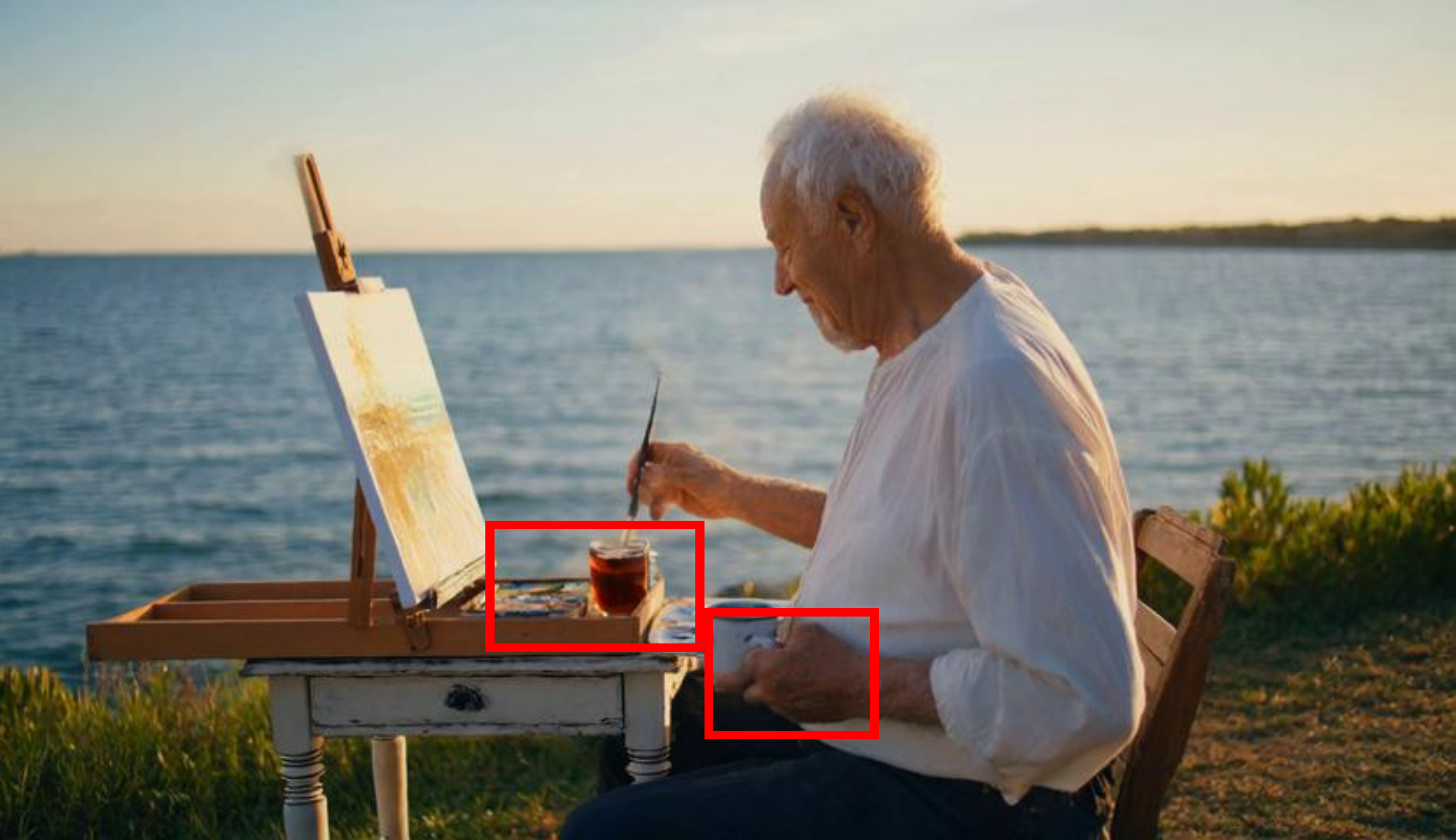}
            % \captionsetup{skip=2.5pt}
        \end{subfigure}
        \hfill
        \captionsetup{skip=2.5pt}
        \begin{subfigure}{0.240\textwidth}
            \centering
            \includegraphics[width=\textwidth]{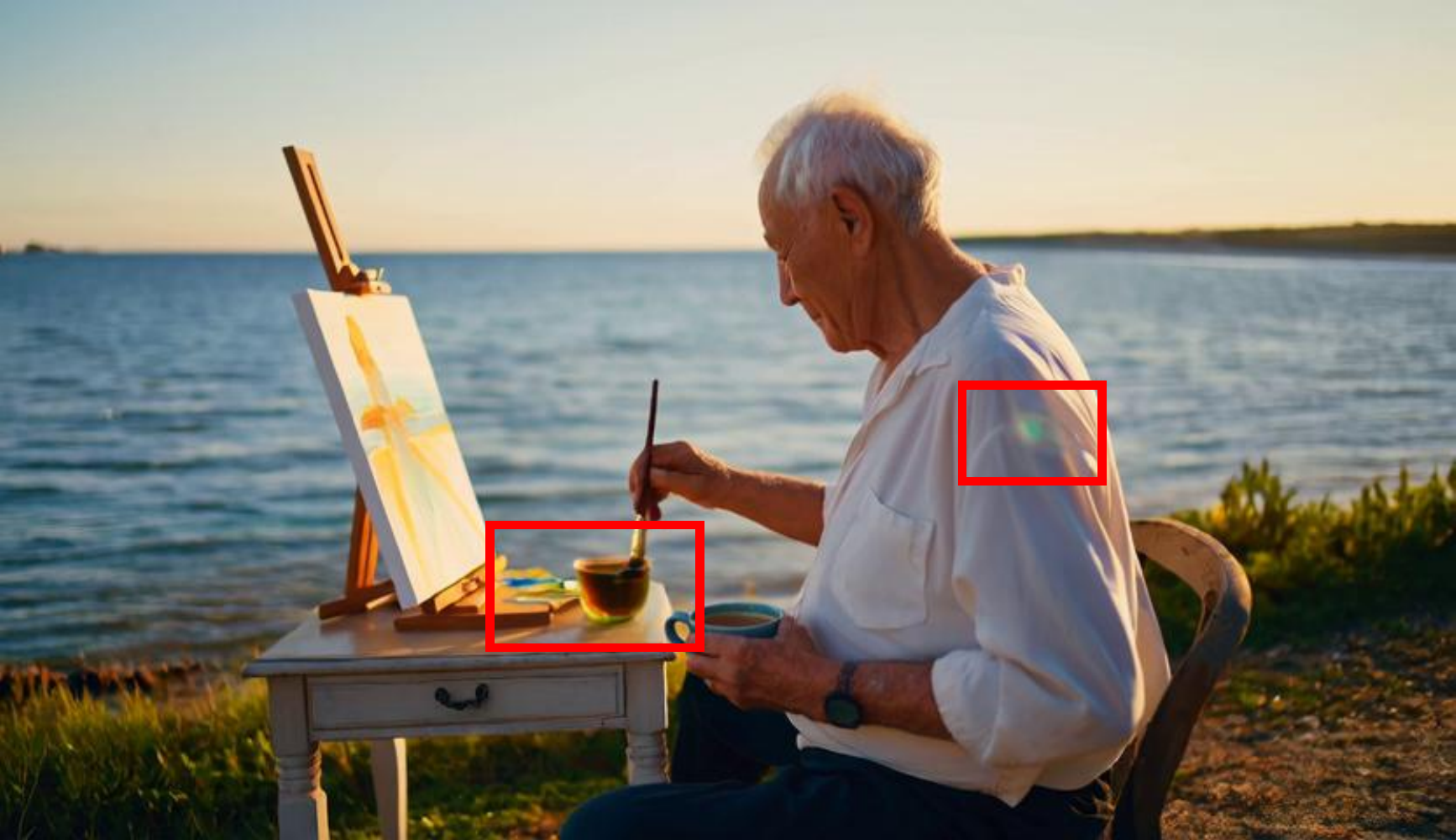}
            % \captionsetup{skip=2.5pt}
        \end{subfigure}
        \hfill
        \captionsetup{skip=2.5pt}
        \begin{subfigure}{0.240\textwidth}
            \centering
            \includegraphics[width=\textwidth]{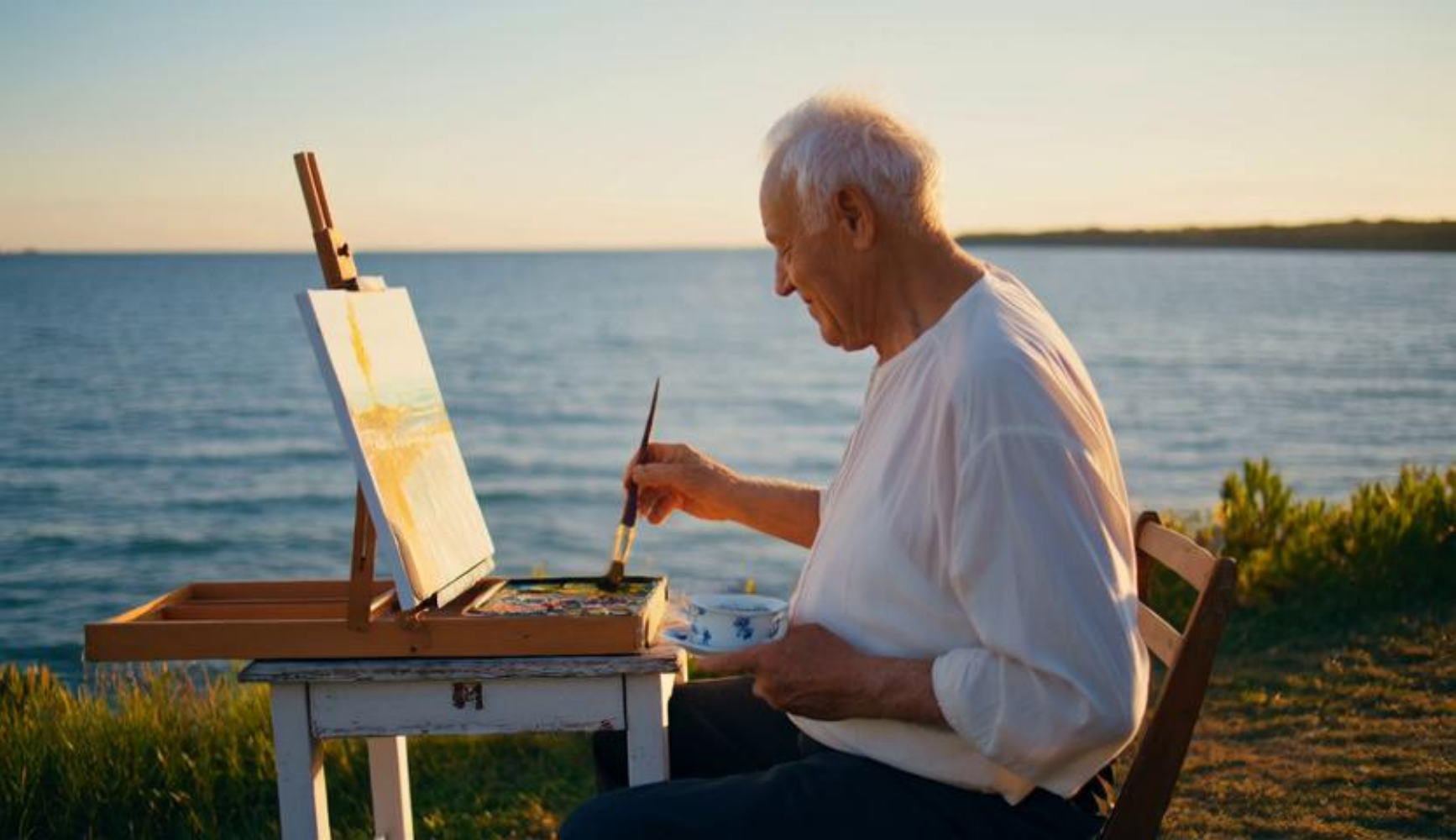}
            % \captionsetup{skip=2.5pt}
        \end{subfigure}
    \end{subfigure}%
    \\
    \begin{subfigure}{\textwidth}
        \centering
        \begin{subfigure}{0.240\textwidth}
            \centering
            \includegraphics[width=\textwidth]{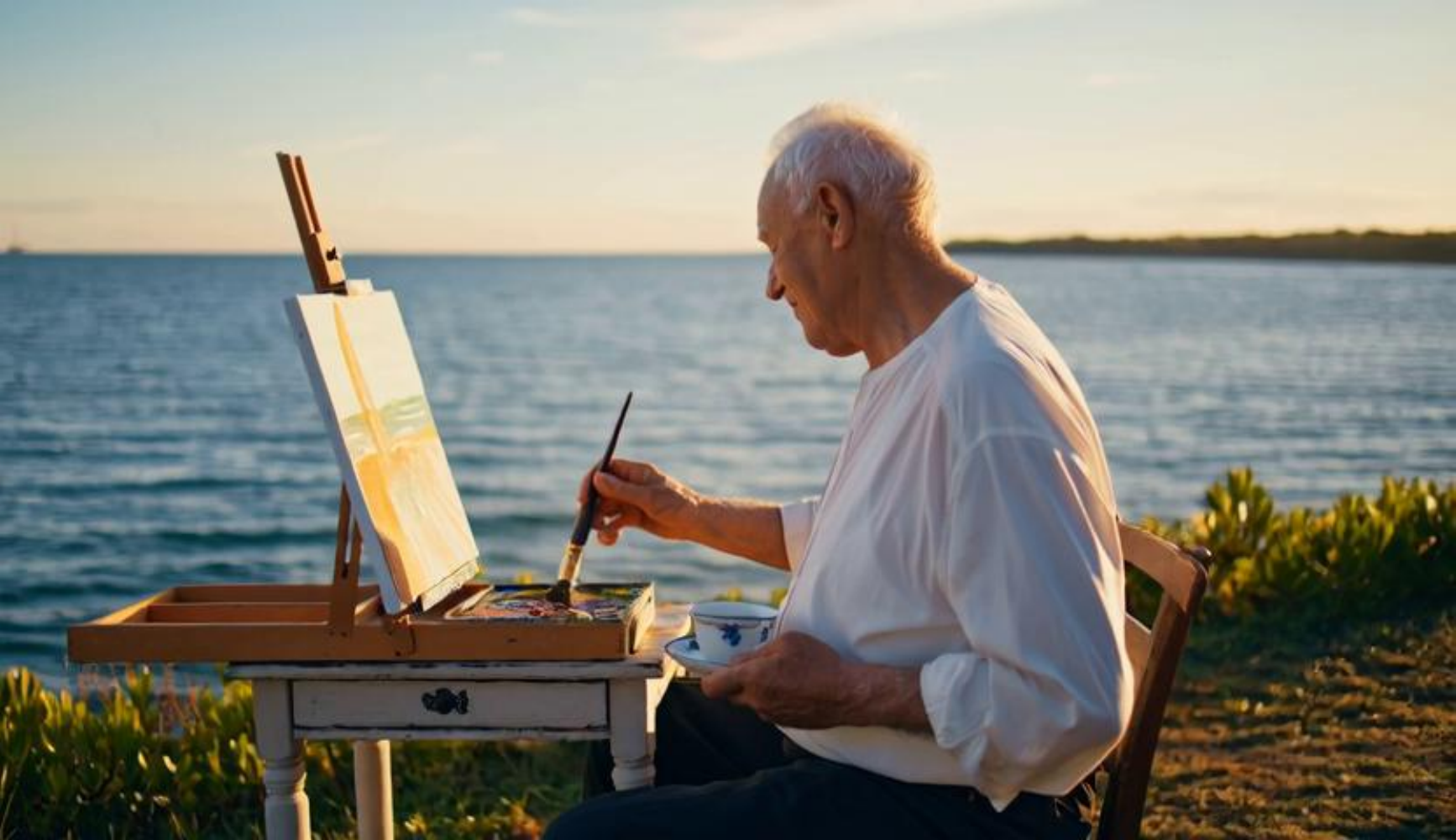}
            \caption*{\textbf{Original}}
            \caption*{{Latency: 948s}}
            \caption*{{}}
        \end{subfigure}%
        \hfill
        \begin{subfigure}{0.240\textwidth}
            \centering
            \includegraphics[width=\textwidth]{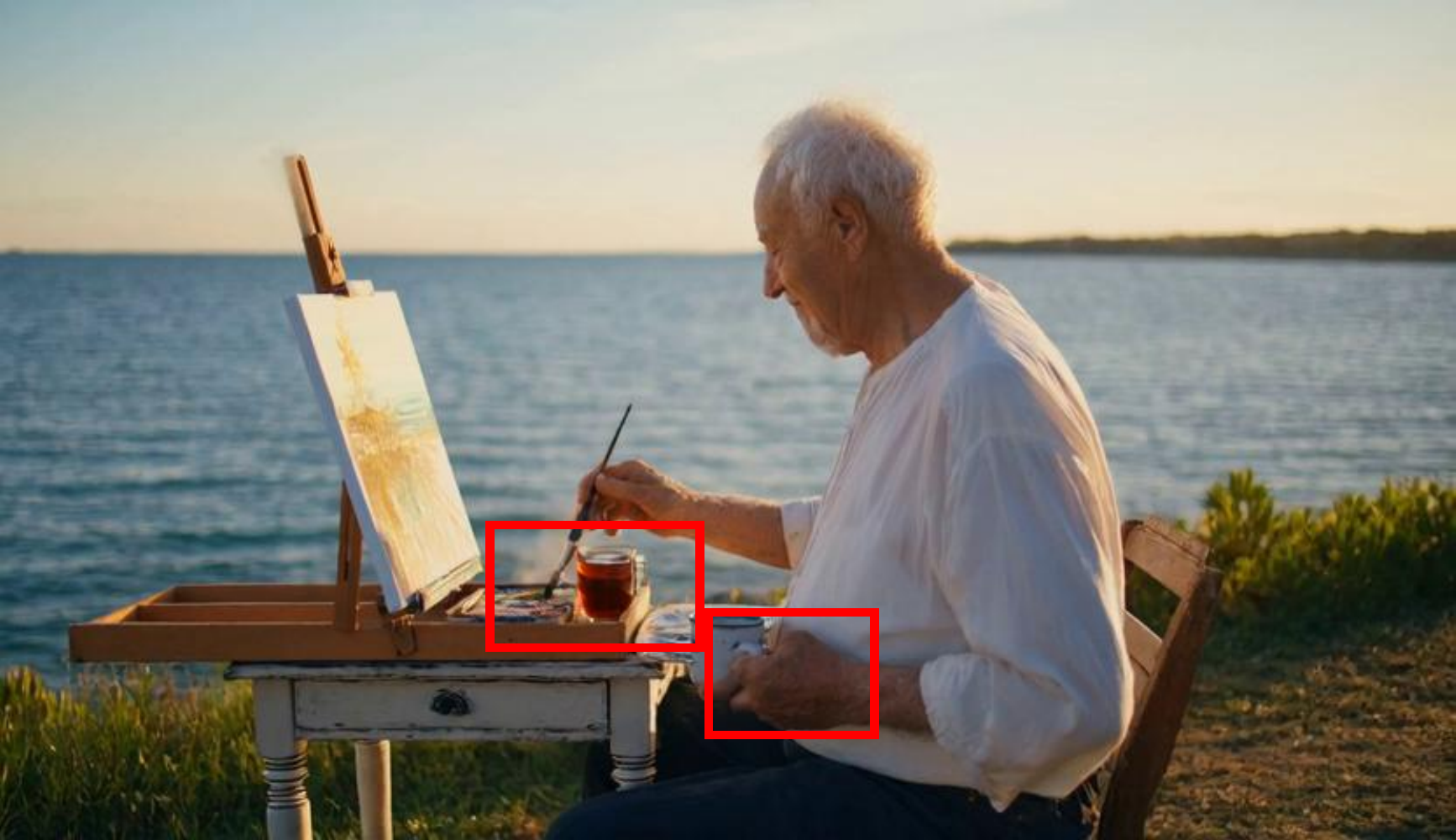}
            \caption*{\textbf{TeaCache}}
            \caption*{{Latency: 362s (2.62$\times$)}}
            \caption*{{LPIPS: 0.208}}
        \end{subfigure}
        \hfill
        \begin{subfigure}{0.240\textwidth}
            \centering
            \includegraphics[width=\textwidth]{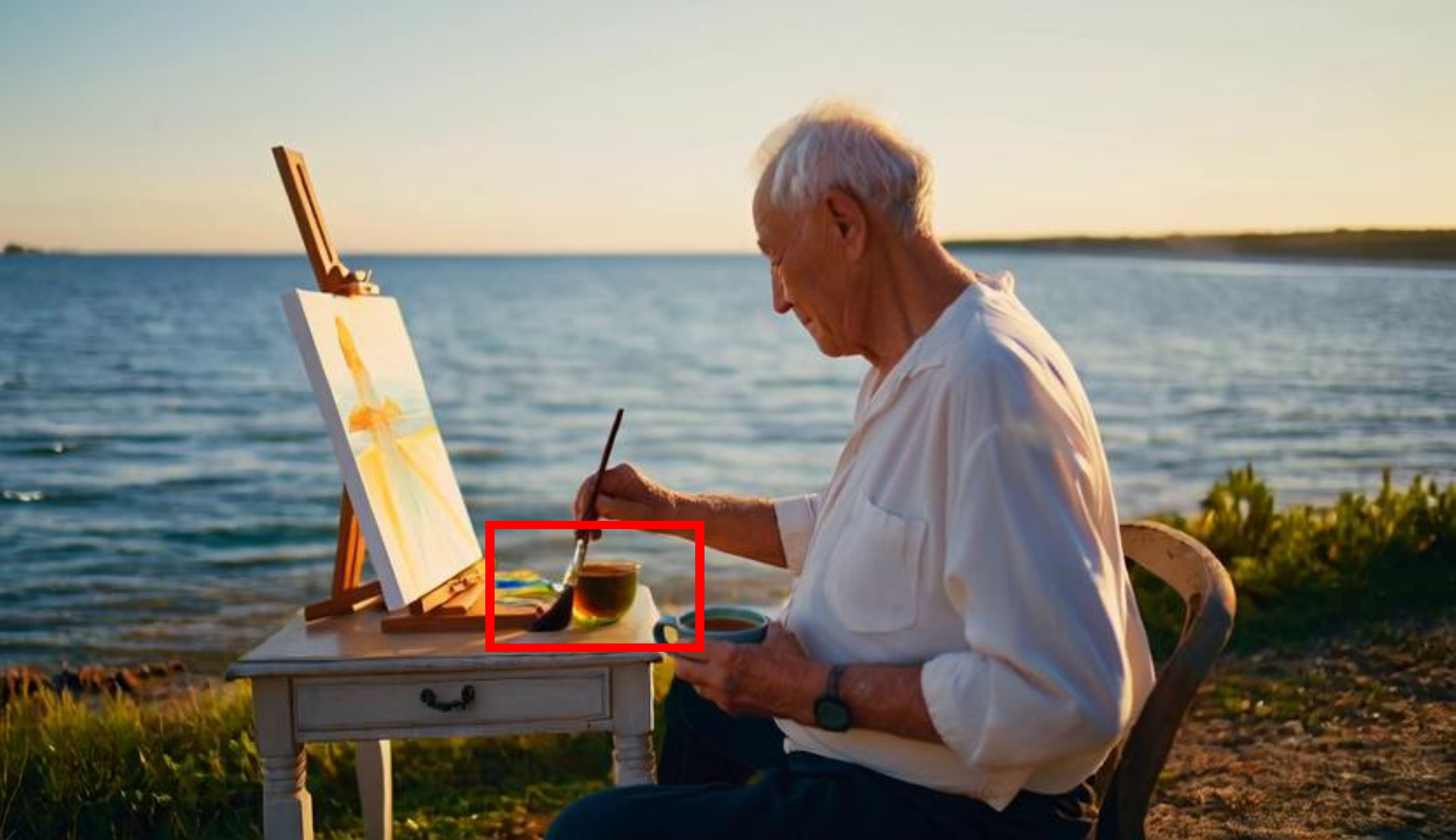}
            \caption*{\textbf{SRDiffusion}}
            \caption*{{Latency: 344s (2.76$\times$)}}
            \caption*{{LPIPS: 0.240}}
        \end{subfigure}
        \hfill
        \begin{subfigure}{0.240\textwidth}
            \centering
            \includegraphics[width=\textwidth]{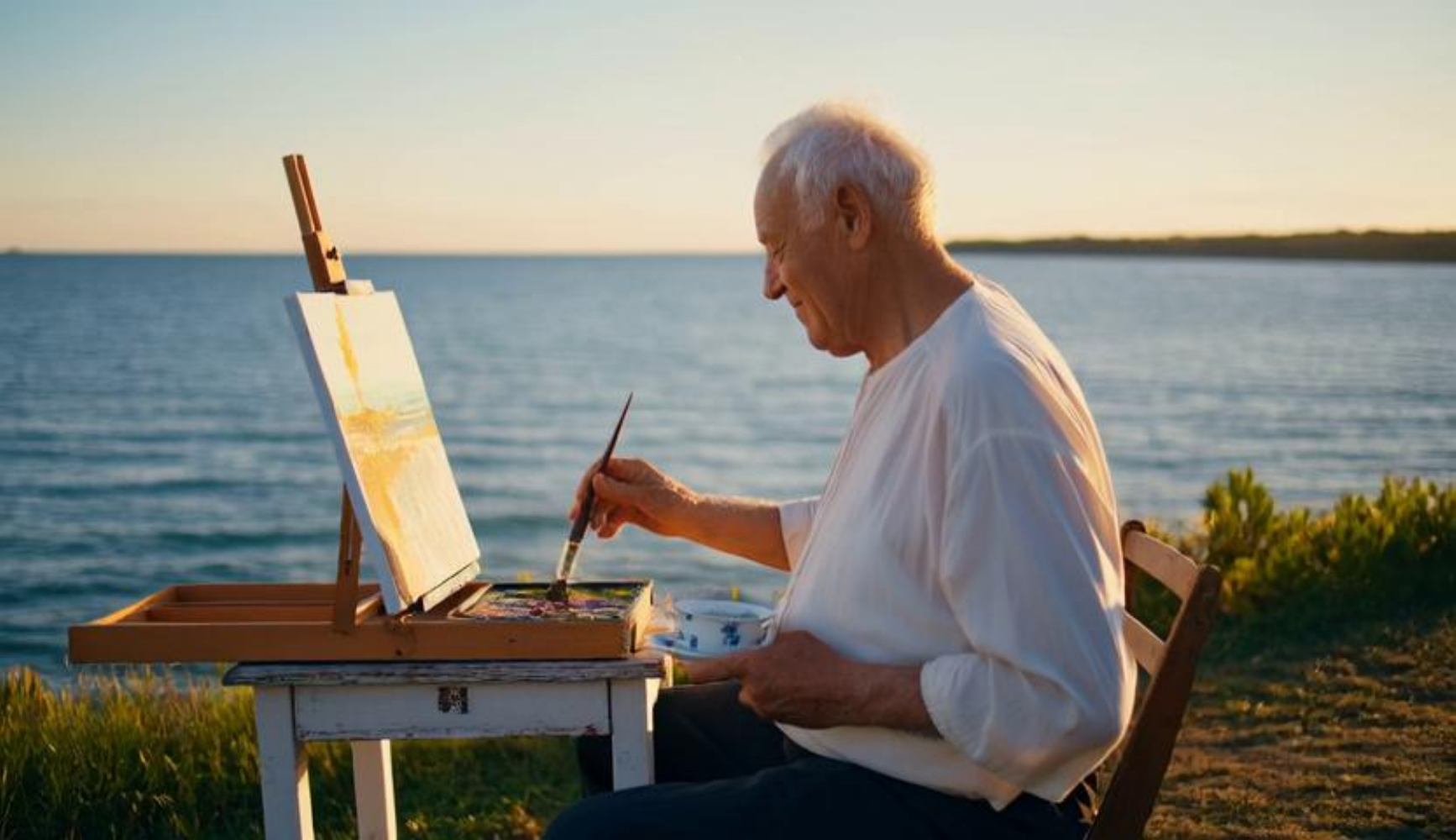}
            \caption*{\textbf{MDD (Ours)}}
            \caption*{Latency: \textbf{321s} \textbf{(2.95$\times$)}}
            \caption*{{LPIPS: \textbf{0.178}}}
        \end{subfigure}
    \end{subfigure}%
    \caption{Comparison of visual fidelity and sampling speed with competing methods. Latency is measured on a single A100 GPU. LPIPS denotes learned perceptual image patch similarity. Video synthesis configuration: 81 frames, 5s, 480P on Wan2.1-14B. Our method consistently generates outputs that better preserve the visual fidelity and richness of the original video content while achieving enhanced acceleration performance. The limitations of competing methods are highlighted in red boxes and are best examined under zoomed-in view.}
    \label{fig:compare}
\end{figure}

However, the prohibitive inference latency of diffusion models remains a critical bottleneck \cite{DBLP:conf/nips/LiWJHC0WTR23}. This core limitation arises from the inherently sequential denoising process, which becomes even more pronounced as models scale to higher resolutions and longer video durations \cite{DBLP:conf/eccv/ChenGXWYRWLLL24}. Existing work based on distillation \cite{DBLP:conf/iclr/SalimansH22,DBLP:conf/cvpr/MengRGKEHS23} or post-training quantization \cite{DBLP:conf/cvpr/ChenMTMJWW025} offers potential acceleration but requires costly model training and additional data resources. In contrast, training-free methods, including cache-based methods \cite{DBLP:conf/cvpr/MaFW24,DBLP:journals/corr/abs-2505-20353} and large-small model collaborative inference method \cite{DBLP:journals/corr/abs-2505-19151}, recognize that the original model is not required for all denoising steps and instead employ lightweight substitutes for certain steps to accelerate sampling. These methods are easy to use, cost-effective, and generally applicable, making them the main focus of this work.

Cache-based methods \cite{DBLP:conf/cvpr/MaFW24,DBLP:journals/corr/abs-2406-01125} observe that model outputs are similar across consecutive timesteps during denoising and propose to reduce redundancy using caching mechanisms, such as residual reuse \cite{DBLP:journals/corr/abs-2506-09045}. Although caching captures this redundancy, its static nature can cause rapid error accumulation and degrade performance under high cache reuse rates (example in Figure \ref{fig:compare}, Teacache \cite{DBLP:conf/cvpr/Liu0W0QZZY025}). The large-small model collaborative inference method SRDiffusion \cite{DBLP:journals/corr/abs-2505-19151} posits that the original large model plays a critical role during the initial stages of semantic construction. In the subsequent stages, a small model from the same family can be employed to refine visual details, serving as a lightweight alternative that accelerates the process. Despite the promising acceleration benefits, directly using the bias output of a limited-capacity small model may lead to denoising trajectory deviations, resulting in suboptimal content richness and visual retention (example in Figure \ref{fig:compare}, SRDiffusion \cite{DBLP:journals/corr/abs-2505-19151}). 

In this work, we conduct an error source analysis under the flow matching framework, comparing cache-based residual reuse outputs and lightweight model outputs with those of the original large model. By decoupling the output into magnitude and directional components, our empirical analysis shows that errors in residual-reuse outputs primarily arise from magnitude discrepancies, while the directional component is well approximated. In contrast, errors in lightweight model outputs are mainly due to directional misalignment, whereas the magnitude information is estimated reliably.

Based on these insights, we propose a novel method called MDD, which accelerates sampling by adaptively employing a reliable lightweight alternative to the original large model. Specifically, MDD improves the outputs of lightweight models by calibrating them along directions obtained from residual reuse, enabling the lightweight alternative output to better approximate the original denoising trajectory. To prevent the accumulation of directional errors from invariant residual reuse, MDD adaptively switches to the original large model for directional recalibration when it detects excessive error growth. Moreover, MDD further reduces inference costs by reusing magnitude information under CFG. As a result, MDD can preserve the content of the original video more faithfully while delivering improved acceleration (example in Figure \ref{fig:compare}, MDD). Qualitative experimental comparisons demonstrate that our MDD outperforms existing acceleration strategies in both sampling efficiency and visual fidelity, verifying the effectiveness of our proposal.

In summary, our contributions are as follows:
\begin{itemize}

\item We empirically find that, under the flow matching framework, the lightweight model can robustly estimate the output magnitude component of the original large model from the same family, while reusing cached residuals can reliably approximate the directional component.

\item We propose MDD, which adaptively employs a direction-calibrated small model as a lightweight alternative to the original large model during inference, while maintaining a faithful approximation of the original denoising trajectory. It also introduces CFG reuse to further accelerate inference.

\item We evaluate MDD on various flow-based video generation models and demonstrate that our approach consistently outperforms existing acceleration strategies, achieving greater speedups while simultaneously delivering superior visual fidelity and content richness.

\end{itemize}

\section{Related Work}
\noindent\textbf{Diffusion and Flow-based Models.}
In generative modeling, diffusion models \cite{DBLP:conf/nips/HoJA20, DBLP:conf/icml/Sohl-DicksteinW15} have become foundational for their ability to produce high-quality, diverse outputs \cite{DBLP:journals/corr/abs-2311-15127, DBLP:conf/cvpr/ChenZCXWWS24}. Early methods, such as DDPM \cite{DBLP:conf/nips/HoJA20}, DDIM \cite{DBLP:conf/iclr/SongME21}, and EDM \cite{DBLP:conf/nips/KarrasAAL22}, are score-based models that learn the stochastic differential equations (SDEs) governing the diffusion process. In contrast, flow matching \cite{DBLP:conf/iclr/LipmanCBNL23} provides an alternative by using ordinary differential equations (ODEs) to model sample trajectories, offering more stable and efficient generation through a deterministic mapping from the prior to the target distribution. Numerous studies \cite{DBLP:conf/iclr/LiuG023,DBLP:conf/icml/EsserKBEMSLLSBP24,DBLP:journals/corr/abs-2412-03603} have shown that flow matching models converge faster and offer better controllability in video generation, making them a strong alternative to stochastic diffusion models with better interpretability and stability. Thus, our analysis is based on flow matching models, aiming to provide a lightweight alternative that enables more efficient sampling.

% Originally developed using the U-Net architecture, these models have shown impressive performance in generation tasks \cite{DBLP:conf/cvpr/RombachBLEO22,DBLP:conf/nips/HoSGC0F22}. However, the scalability of U-Net-based diffusion models is limited, posing challenges for applications that require larger model capacities to enhance performance. By leveraging the inherently scalable architecture of transformers \cite{DBLP:conf/nips/VaswaniSPUJGKP17}, DiT \cite{DBLP:conf/iccv/PeeblesX23} offers an effective solution for scaling up model capacities and has made a considerable impact \cite{Open-Sora}. Despite the success of DiT-based large-scale diffusion models, their iterative nature results in significant inference latency, which remains a key barrier to adoption and necessitates further research into more efficient generation methods.

\noindent\textbf{Diffusion Model Acceleration.} Due to their high sampling costs, diffusion models have prompted extensive efforts for acceleration. One line of research focuses on methods based on training or fine-tuning, such as post-training quantization \cite{DBLP:conf/cvpr/ShangYXW023,DBLP:conf/nips/HeLLWZZ23,DBLP:conf/iccv/LiLLYDKZK23,DBLP:conf/cvpr/Wang0XT0L24,DBLP:conf/cvpr/ChenMTMJWW025} and step distillation \cite{DBLP:conf/iclr/SalimansH22,DBLP:conf/cvpr/MengRGKEHS23,DBLP:journals/corr/abs-2403-12706,DBLP:conf/eccv/SauerLBR24,DBLP:conf/icml/SongD0S23,DBLP:journals/corr/abs-2310-04378}. However, they necessitate costly retraining and additional data resources, which can constrain their feasibility for broad implementation.

Another line of research focuses on training-free methods. Foundational methods like DDIM \cite{DBLP:conf/iclr/SongME21} enable fewer sampling steps without sacrificing quality. Additional studies use efficient ODE or SDE solvers \cite{DBLP:conf/nips/SongE19, DBLP:conf/nips/KarrasAAL22, DBLP:conf/nips/0011ZB0L022}, employing pseudo-numerical methods for faster sampling. Furthermore, a series of studies \cite{DBLP:conf/cvpr/MaFW24,DBLP:journals/corr/abs-2406-01125,DBLP:journals/corr/abs-2504-10540} have shown that not every step in the iterative denoising process requires the full original model, motivating various lightweight alternatives to accelerate sampling. Cache-based methods \cite{DBLP:conf/cvpr/WimbauerWSDHHSZ24, DBLP:journals/corr/abs-2503-06923, DBLP:journals/corr/abs-2506-09045} exploit redundancies in model outputs across consecutive timesteps, caching outputs according to criteria such as content complexity (AdaCache \cite{DBLP:journals/corr/abs-2411-02397}) or timestep embeddings (Teacache \cite{DBLP:conf/cvpr/Liu0W0QZZY025}) to reduce computational overhead. Large-small model collaborative inference strategies \cite{DBLP:conf/iclr/YangCW0C24} leverage lightweight models from the same family as substitutes to accelerate sampling. For example, SRDiffusion \cite{DBLP:journals/corr/abs-2505-19151} uses the original large model during the early stages of denoising to generate coarse semantic structures, which are then refined by a lightweight model responsible for producing fine visual details. However, the limited capacity of the lightweight model can lead to deviations from the original large model’s denoising trajectory, compromising content richness and visual retention.
\section{Preliminaries}

\noindent \textbf{Flow Matching} \cite{DBLP:conf/iclr/LipmanCBNL23} is a family of generative models that transport samples from a data distribution \( p_0(x) \) into a simple prior distribution \( p_1(x) \) (e.g., Gaussian). The probability path $p_t(x)$ is constructed to interpolate between $p_0(x)$ and $p_1(x)$ over the continuous time variable $t \in [0, 1]$, and is commonly defined using linear interpolation \cite{DBLP:conf/icml/EsserKBEMSLLSBP24}:
\[
p_t(x) = (1 - t) \cdot p_0(x) + t \cdot p_1(x).
\]
The flow matching model is trained by learning a time-dependent velocity field \( \frac{d}{dt} x_t = v_\theta(x_t,t) \), where \( v_\theta(x_t,t) \) is parameterized by a neural network with parameters \( \theta \). Formally, the model is optimized by minimizing the following loss function:
\[
\mathcal{L}_{\text{FM}}(\theta) = \mathbb{E}_{t, x_0 \sim p_0(x), x_1 \sim p_1(x)} \left[ v_\theta(x_t,t) - (x_1 - x_0) \right].
\]
At generation time, new samples can be generated using any ODE solver \cite{suli2003introduction}.
% where \(c\) denotes the input condition such as a class label or a text prompt

\noindent\textbf{Residual Reuse.} 
A series of studies \cite{DBLP:journals/corr/abs-2411-02397,DBLP:conf/cvpr/Liu0W0QZZY025} have consistently observed that the denoising process often involves redundant computations, suggesting that reuse strategies could be employed to skip certain steps. Specifically, we define the residual $r$ at timestep $t$ as the difference between the model’s output and the corresponding input:
\begin{equation}
r = v_{\theta}(x_t,t) - x_t.
\end{equation}
In subsequent steps, this residual can be reused by adding it to the input, thus directly estimating the residual reuse output (e.g., $\hat{v}_\theta(x_{t-1},t-1)=x_{t-1}+r$). Although $r$ effectively captures the update signal, repeatedly reusing an invariant residual may result in cumulative errors, which can ultimately degrade the quality of the generated visual output.

\noindent \textbf{Classifier-Free Guidance (CFG)}~\cite{DBLP:journals/corr/abs-2207-12598} is a widely adopted strategy for improving the quality of conditional generation by steering samples toward the specified input condition. Let \(c\) denote the input condition, such as a class label or a text prompt. In CFG, a single flow model $v_\theta(x_t,t \mid c)$ is trained to output both conditional and unconditional velocity fields. During sampling, the model performs two forward evaluations: one conditioned on $c$ and the other unconditioned, with $c = \emptyset$. The guided velocity field is formed by:
\begin{equation}
v^{\text{cfg}}_\theta(x_t, t \mid c) = v_\theta(x_t,t \mid c=\emptyset) + w \cdot \Big(v_\theta(x_t,t \mid c) - v_\theta(x_t,t \mid c=\emptyset)\Big),
\end{equation}
where $w$ denotes the guidance scale. In particular, setting $w = 1$ corresponds to the non-guided case.

\section{Proposed Method}

\subsection{Empirical Analysis}
\label{sec:emp}
Recent flow-based model families \cite{DBLP:journals/corr/abs-2503-20314,DBLP:journals/corr/abs-2405-18991} offer both large and small model variants. By sharing a unified VAE within these model families, the models enable seamless switching between the high-capacity large model ($\theta$) and the lightweight small model ($\varphi$) during sampling, thus opening up opportunities for large-small model collaboration to accelerate inference.

Let $v_{\theta}(x_t,t)$ and $v_{\varphi}(x_t,t)$ denote the outputs of the large and small models, respectively, and $\hat{v}_{\theta}(x_t,t)$ represents the residual-reused outputs from the original large model. In this paper, we explore how $v_{\varphi}(x_t,t)$ and $\hat{v}_{\theta}(x_t,t)$ relate to $v_{\theta}(x_t,t)$, aiming to identify better lightweight alternatives for replacing the costly inference of the original large model. Specifically, we decouple the model outputs into direction and magnitude components for separate analysis. As shown in Figures \ref{fig:a}, \ref{fig:b}, \ref{fig:d}, and \ref{fig:e}, based on the empirical analysis of the intermediate diffusion  process (i.e., from 20\% to 95\%), we derive two key observations:

\begin{figure}[t]
    \centering
    \begin{subfigure}{\textwidth}
        \centering
        \caption*{Wan2.1 (large model: 14B, small model: 1.3B)}
        \begin{subfigure}{0.325\textwidth}
            \centering
            \includegraphics[width=\textwidth]{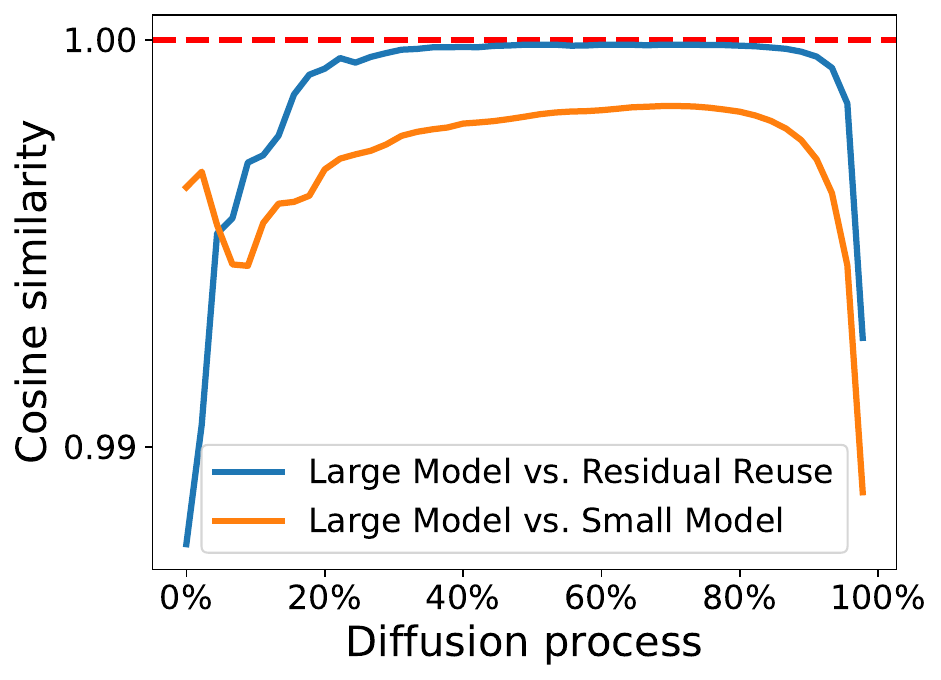}
            \captionsetup{skip=0.5pt}
            \caption{Directional Similarity}
            \label{fig:a}
        \end{subfigure}%
        \hfill
        \captionsetup{skip=2.5pt}
        \begin{subfigure}{0.325\textwidth}
            \centering
            \includegraphics[width=\textwidth]{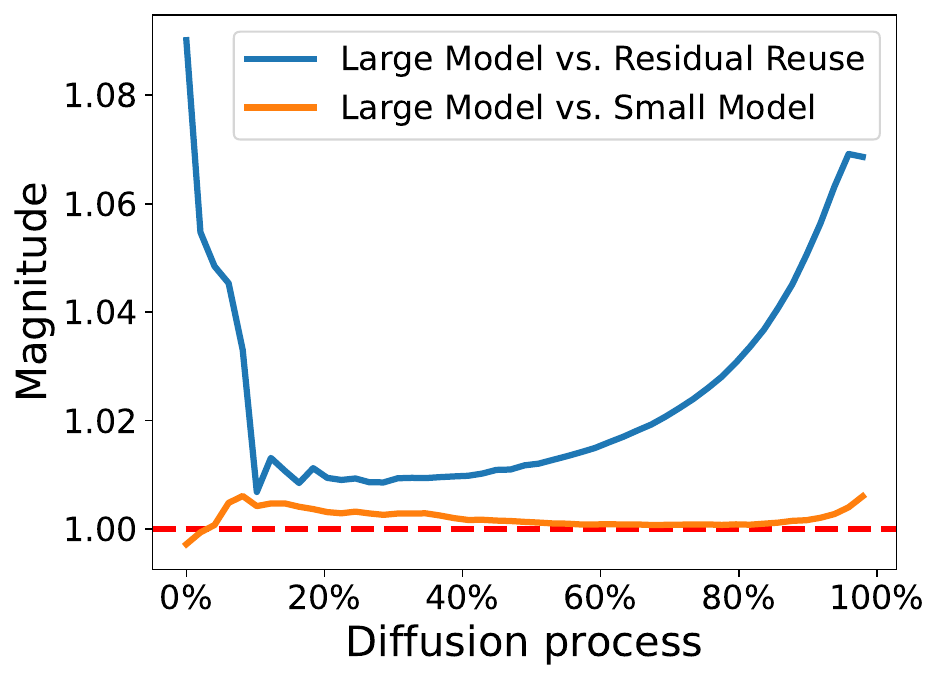}
            \captionsetup{skip=0.5pt}
            \caption{Magnitude Similarity}
            \label{fig:b}
        \end{subfigure}
        \hfill
        \captionsetup{skip=2.5pt}
        \begin{subfigure}{0.325\textwidth}
            \centering
            \includegraphics[width=\textwidth]{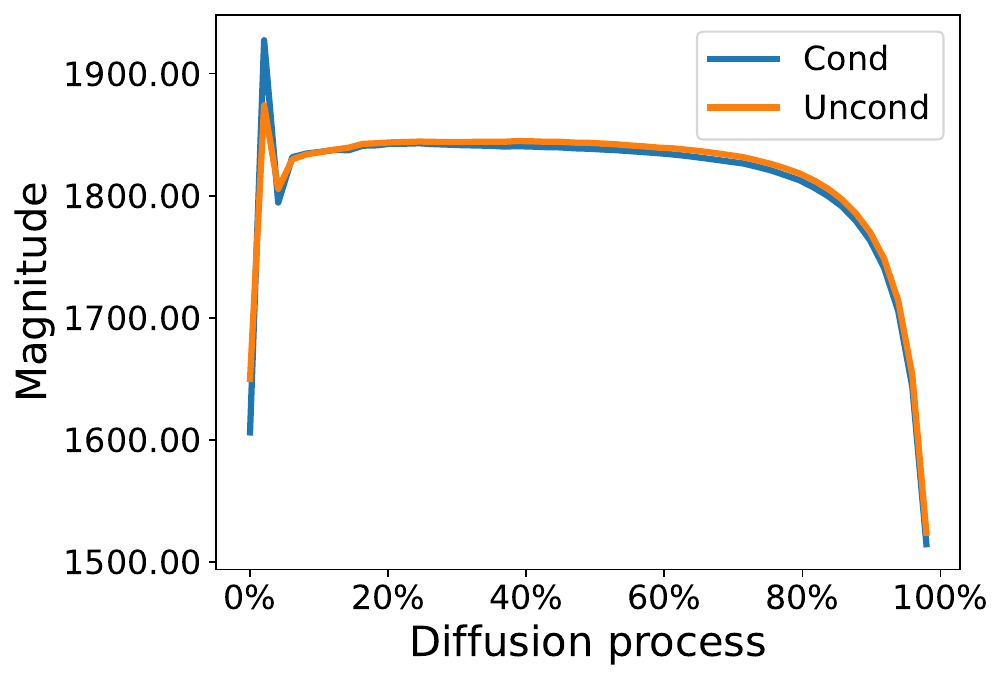}
            \captionsetup{skip=0.5pt}
            \caption{CFG Similarity}
            \label{fig:c}
        \end{subfigure}
    \end{subfigure}%
    \\
    \begin{subfigure}{\textwidth}
        \centering
        \caption*{EasyAnimateV5.1 (large model: 12B, small model: 7B)}
        \begin{subfigure}{0.325\textwidth}
            \centering
            \includegraphics[width=\textwidth]{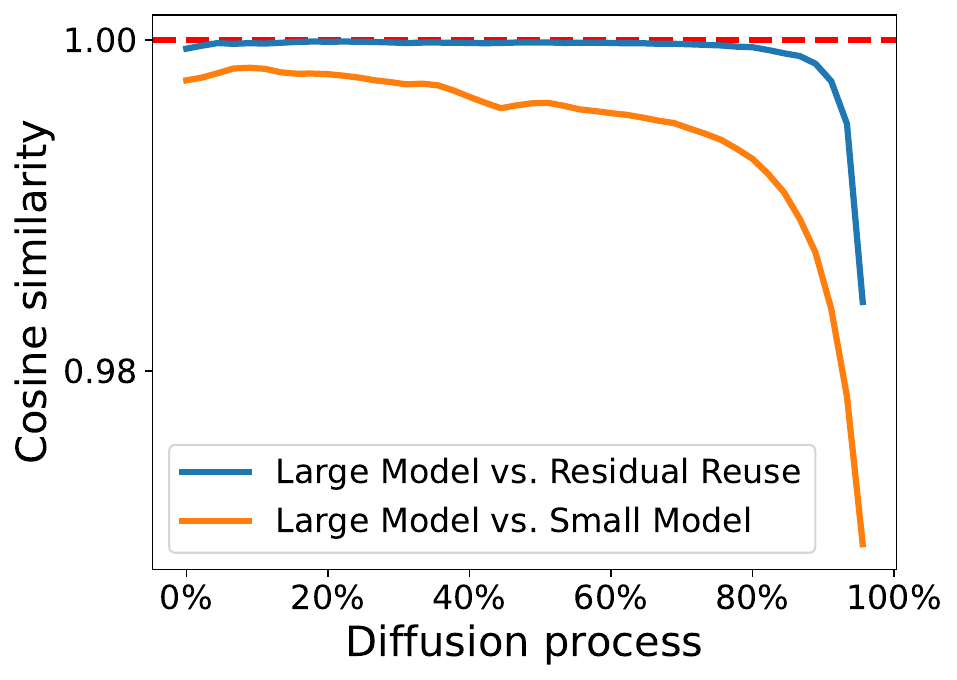}
            \captionsetup{skip=0.5pt}
            \caption{Directional Similarity}
            \label{fig:d}
        \end{subfigure}%
        \hfill
        \captionsetup{skip=2.5pt}
        \begin{subfigure}{0.325\textwidth}
            \centering
            \includegraphics[width=\textwidth]{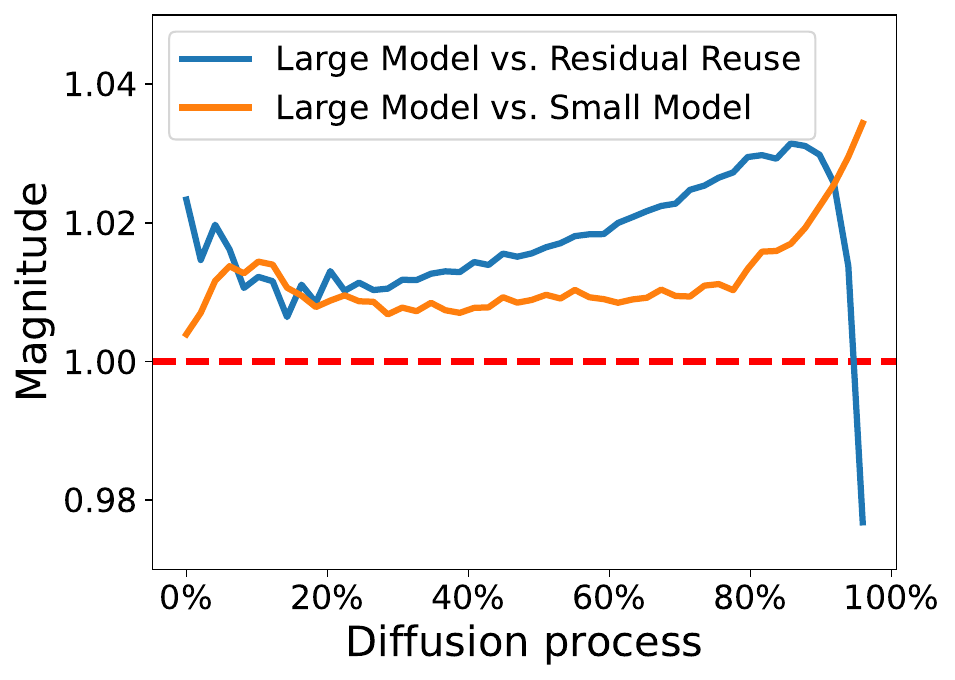}
            \captionsetup{skip=0.5pt}
            \caption{Magnitude Similarity}
            \label{fig:e}
        \end{subfigure}
        \hfill
        \captionsetup{skip=2.5pt}
        \begin{subfigure}{0.325\textwidth}
            \centering
            \includegraphics[width=\textwidth]{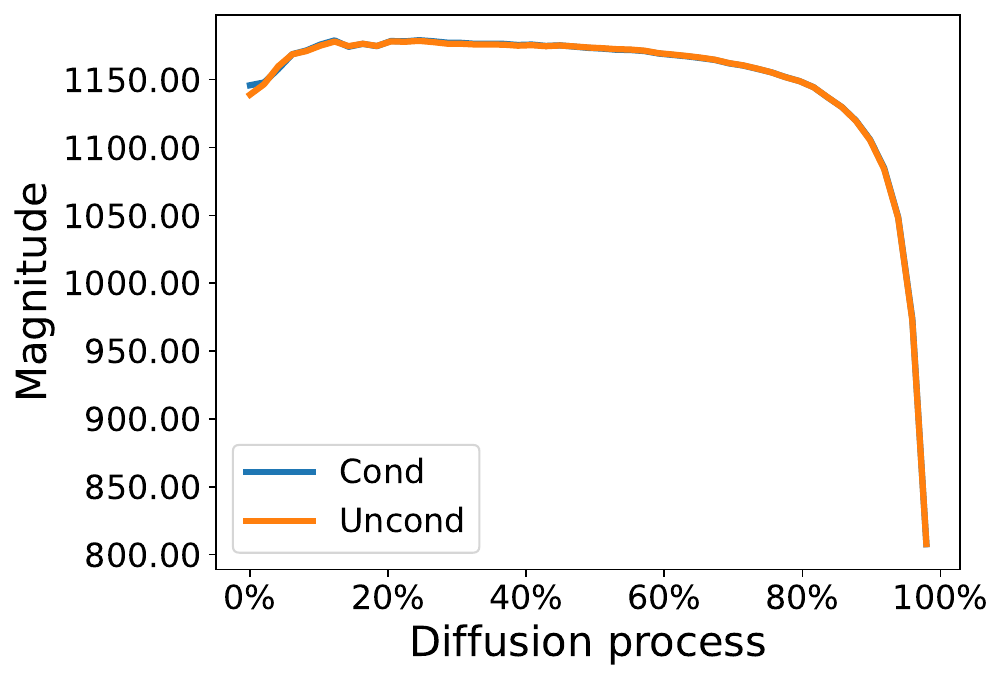}
            \captionsetup{skip=0.5pt}
            \caption{CFG Similarity}
            \label{fig:f}
        \end{subfigure}
    \end{subfigure}%
    % \captionsetup{skip=1.0pt}
    \caption{
        Visualization of decoupled model output relationships. (a,d) directional similarities of $v_{\theta}(x_t,t)$ with $v_{\varphi}(x_t,t)$ and $\hat{v}_{\theta}(x_t,t)$; (b,e) magnitude ratios of $v_{\theta}(x_t,t)$ to $v_{\varphi}(x_t,t)$ and to $\hat{v}_{\theta}(x_t,t)$; (c,f) magnitudes of $v_{\varphi}(x_t,t \mid c)$ (Cond) and $v_{\varphi}(x_t,t \mid c=\emptyset)$ (Uncond). Directional similarities are evaluated using cosine similarity, and magnitudes are measured using $\ell_2$ norms.
    }
    \label{fig:expamles}
\end{figure}

\textit{\textbf{Observation 1:} In terms of directional component, the residual-reuse outputs $\hat{v}_{\theta}(x_t,t)$ are closely aligned with those of the original large model $v_{\theta}(x_t,t)$, i.e., $\text{sim}(\hat{v}_{\theta}(x_t,t), v_{\theta}(x_t,t)) \approx 1$, where $\text{sim}(\cdot,\cdot)$ denotes the cosine similarity. In contrast, the small model's directional estimates typically diverge more, i.e., $\text{sim}(v_{\theta}(x_t,t), \hat{v}_{\theta}(x_t,t)) > \text{sim}(v_{\theta}(x_t,t), v_{\varphi}(x_t,t))$.}

\textit{\textbf{Observation 2:} In terms of magnitude component, the small-model outputs $v_{\varphi}(x_t,t)$ closely resemble those of the large model $v_{\theta}(x_t,t)$, i.e., $\frac{\|v_{\theta}(x_t,t)\|_2}{\|v_{\varphi}(x_t,t)\|_2} \approx 1$, where $\|\cdot\|_2$ denotes the $\ell_2$ norm. In contrast, the residual-reuse outputs typically show a larger magnitude discrepancy, i.e., $\left| \frac{ \| v_{\theta}(x_t,t) \|_2}{\| v_{\varphi}(x_t,t) \|_2}-1 \right| < \left| \frac{\| v_{\theta}(x_t,t) \|_2}{\| \hat{v}_{\theta}(x_t,t) \|_2} - 1 \right|$.}

With respect to the direction component, the residual-reuse outputs closely approximate those of the large model, whereas the small-model outputs display directional deviations. This implies that directly employing a capacity-limited lightweight model for collaborative inference may cause it to diverge from the large model’s denoising trajectory, ultimately leading to suboptimal visual retention. Conversely, the learning objective in flow matching models allows for efficient capture of directional information through residual reuse, a finding also supported by \cite{DBLP:journals/corr/abs-2506-09045}.

With respect to magnitude component, the outputs of the large and small models are comparatively consistent. This suggests that the small model can serve as an effective lightweight proxy for the large model in estimating output magnitude. However, consecutive reuse of residual outputs with inconsistent magnitudes can result in rapid error accumulation, ultimately degrading visual quality.

Collectively, in approximating the outputs of the large model, the residual-reuse outputs provide more accurate directional estimates, whereas the small-model outputs better capture the output magnitudes. This insight informs the design of our subsequent acceleration strategy.

\subsection{Acceleration Strategy}
\begin{figure*}[t]
 \centering
 \includegraphics[scale=0.3445]{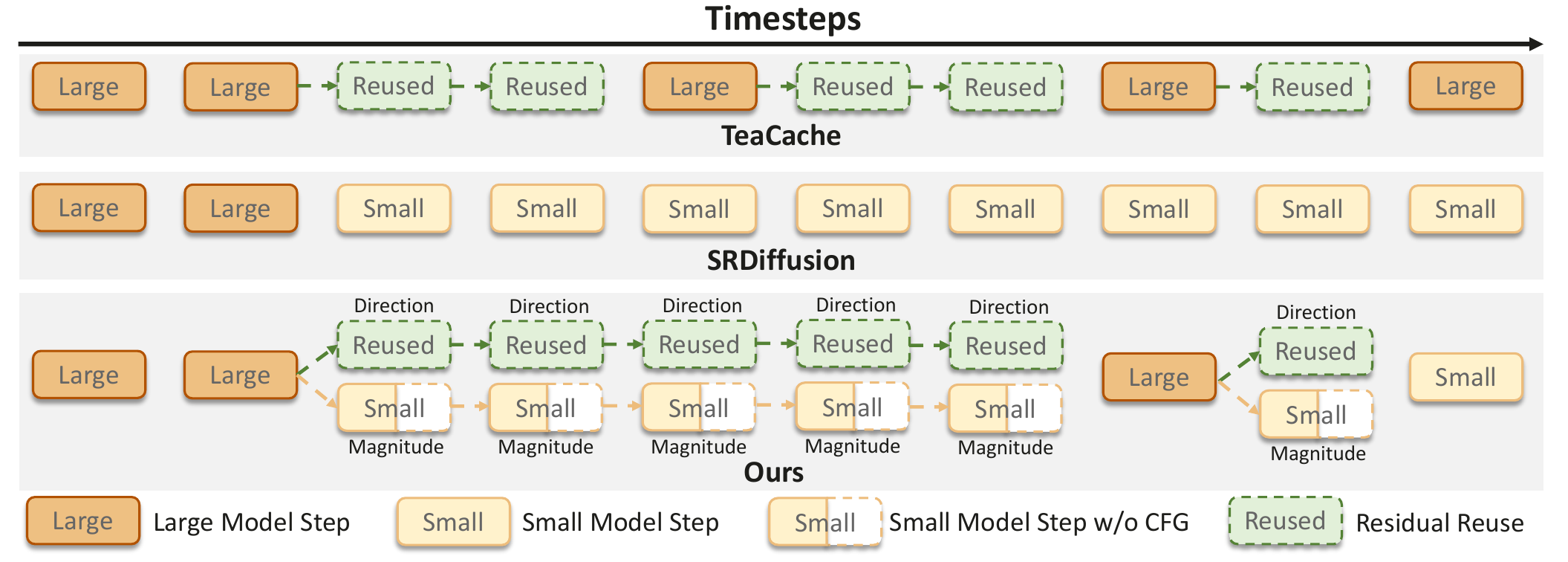}
 \caption{Comparison of our method with the adaptive caching method, TeaCache \cite{DBLP:conf/cvpr/Liu0W0QZZY025}, and the large-small model collaborative inference strategy, SRDiffusion \cite{DBLP:journals/corr/abs-2505-19151}. Our method improves the approximation of the original large output by combining direction estimation from residual reuse with magnitude estimation from the small model, offering a more reliable lightweight alternative for accelerating inference.}
 \label{fig:framework}
\end{figure*}

In this subsection, building upon the previous empirical analysis, we propose a novel sampling acceleration strategy called MDD. The objective is to provide a reliable and lightweight alternative to the original large model. The framework of our proposal and its comparison with competing methods are illustrated in Figure \ref{fig:framework}, with a detailed description provided below.

\noindent\textbf{Lightweight Alternative Strategy.}
As discussed in Section \ref{sec:emp}, $\hat{v}_{\theta}(x_t,t)$ and $v_{\varphi}(x_t,t)$ are, respectively, more accurate in capturing the direction and magnitude components of the original large model output $v_{\theta}(x_t,t)$. By leveraging this complementary advantage, MDD retains the magnitude from the lightweight model $v_{\varphi}(x_t,t)$ while utilizing the residual-reused output $\hat{v}_{\theta}(x_t,t)$ to guide the direction. Formally, at timestep $t$ with input $x_t$ and condition $c$, the output estimated by our MDD strategy can be expressed as:
\begin{equation} 
\label{eq:MDD}
v_{\text{MDD}}(x_{t},t \mid c) = \underbrace{\| v_{\varphi}(x_t,t \mid c) \|_2}_{\text{Magnitude}} \cdot \underbrace{\frac{\hat{v}_{\theta}(x_t,t \mid c)}{\| \hat{v}_{\theta}(x_t,t \mid c) \|_2}}_{\text{Direction}}. \end{equation}

Compared to SRDiffusion \cite{DBLP:journals/corr/abs-2505-19151}, which performs sampling directly using a small model with directional prediction biases, MDD improves by calibrating the direction through residual reuse. As a result, MDD more closely approximates the denoising trajectory of the original large model, thus enhancing visual retention. Compared with Teacache \cite{DBLP:conf/cvpr/Liu0W0QZZY025}, a cache-based method that reuses invariant residuals, MDD combined with the lightweight assistance of the small model for magnitude prediction. From the perspective of caching, MDD equips residual reuse with magnitude-aware capabilities, thereby improving visual fidelity.

% The MDD strategy provides an approximation that closely mirrors the outputs of the original large model. In contrast to SRDiffusion \cite{DBLP:journals/corr/abs-2505-19151}, MDD effectively mitigates the issue of denoising trajectory deviation caused by directly utilizing the outputs of the small model, thereby preserving the generative capabilities of the original large model.

% We empirically validate the approach on a toy example consisting of a Gaussian mixture, with results shown in Fig. 3(a), where we observe that samples generated with CFG-Zero⋆ more closely match the target distribution than those with CFG.

\noindent\textbf{CFG Reuse.} As illustrated in Figures \ref{fig:c} and \ref{fig:e}, the magnitude of the model outputs in CFG are nearly identical for both the conditional output $v_{\varphi}(x_t,t \mid c)$ and the unconditional output $v_{\varphi}(x_t,t \mid c=\emptyset)$, i.e., $\|v_{\varphi}(x_t,t \mid c)\|_2 \approx \|v_{\varphi}(x_t,t \mid c=\emptyset)\|_2$. Consequently, a single forward evaluation to compute the magnitude of the conditional output in CFG is sufficient. This magnitude information can then be combined with both the residual-guided conditional and unconditional directions, further halving the inference cost of the small model. Formally, the unconditional output estimated by our MDD is expressed as:

\begin{equation} 
v_{\text{MDD}}(x_{t},t \mid c=\emptyset) = \underbrace{\| v_{\varphi}(x_t,t \mid c) \|_2}_{\text{Reuse}} \cdot \frac{\hat{v}_{\theta}(x_t,t \mid c=\emptyset)}{\| \hat{v}_{\theta}(x_t,t \mid c=\emptyset) \|_2}. \end{equation}

\noindent\textbf{Adaptive Strategy.} 
The reuse of invariant residuals for direction estimation lacks the intrinsic self-correction capability present in small-model magnitude estimation. As a result, even minor directional errors can accumulate progressively over successive reuse steps, leading to significant error amplification. Once the directional error exceeds a tolerable threshold, it becomes necessary to invoke the original large model to promptly correct the direction. In practice, we employ the cumulative cosine error of the output directional components as the criterion for triggering MDD. To reduce the computational cost associated with repeated forward passes of the large model, we instead utilize the small model as a surrogate to estimate the directional error. Formally, we maintain a running cumulative error $\mathcal{E}$ defined as:
\begin{equation}
\label{eq:error}
\mathcal{E} = \sum_{i=t}^{t'-1} \Bigl( 1 - \text{sim}(v_{\varphi}(x_i,i \mid c), v_{\varphi}(x_{t'-1},t'-1 \mid c) )\Bigr),
\end{equation}
where $t'$ denotes the most recent timestep at which a residual is cached, and $\text{sim}(\cdot,\cdot)$ represents the cosine similarity. $\mathcal{E}$ quantifies the accumulated deviation of output directions over successive timesteps: smaller values indicate that the directions remain well-aligned and can be safely reused, whereas larger values trigger the use of the original large model to update the directions. Accordingly, the decision function at timestep $t$ is defined as:
\begin{align}
\label{eq:th}
D_\tau(t) &= 
\begin{cases}
v_{\text{MDD}}(x_t,t \mid c),  & \text{if } \mathcal{E} \leq \tau; \\
v_{\theta}(x_t,t \mid c), & \text{if } \mathcal{E} > \tau,
\end{cases}
\end{align}
where $\tau$ denotes the threshold for the maximum tolerable directional error, serving to balance computational efficiency and visual quality. The pseudo code of MDD is available in Algorithm \ref{alg:main}.

\begin{algorithm}[t]
    \begin{minipage}{\linewidth}
    \begin{footnotesize}
        \caption{Magnitude-Direction Decoupling (MDD)}
        \label{alg:main}
        \begin{algorithmic}[1]
            \STATE \textbf{Initialize} latent variable $x_T$, MDD interval $[T_1, T_2]$, directional error $\mathcal{E}$, and threshold $\tau$
            \FOR{each timestep $t$ in $\{{T_2}, \dots, {T_1}\}$}
                \IF{$\mathcal{E} > \tau$}
                    \STATE Predict outputs: $v_t \gets  \text{LargeModel}(x_t, t)$~~~~~~
                    \texttt{ \# With CFG}
                    \STATE Cache residuals: $r \gets v_t - x_t$
                    \STATE Reset the error: $\mathcal {E} \gets 0$
                \ELSE
                    \STATE Predict output: $v'_t \gets \text{SmallModel}(x_t, t)$~~~~~~~
                    \texttt{ \# Without CFG}
                    \STATE Reuse residuals: $\hat{v}_t \gets x_t+r$
                    \STATE MDD: $v_t \gets \text{norm}(v'_t) \cdot (\hat{v}_t~/~\text{norm}(\hat{v}_t))$
                    \STATE Update the error: $\mathcal{E} \gets \mathcal{E} + $ directional error of $v'_t$ 
                    % ~~
                    % \texttt{ \# According to Eq.(\ref{eq:error})}
                \ENDIF
                    \STATE Update latents: $x_{t-1} \gets \text{ODEStep}(v_t, t, x_t)$
            \ENDFOR
            \RETURN $x_{T_1}$
        \end{algorithmic} 
        \end{footnotesize}
    \end{minipage}
\end{algorithm}

\noindent\textbf{Discussion of threshold $\tau$.} Following prior work \cite{DBLP:journals/corr/abs-2505-19151,DBLP:conf/cvpr/Liu0W0QZZY025}, we adopt a fixed threshold to trigger switching in the original large model for corrective purposes. In practice, the error accumulation is driven by the small neural network, and under a fixed threshold, this naturally enables dynamic, self-adaptive adjustment, allowing the strategy to be modulated according to the generated error for different inputs. For parameter selection, a practical approach is to perform a parameter sweep over $\tau$ on a small subset of samples from the new model, relating the threshold to perceptual quality metrics. For each candidate value, quality-preservation metrics such as SSIM \cite{DBLP:journals/tip/WangBSS04} or LPIPS \cite{DBLP:conf/cvpr/ZhangIESW18} can be evaluated against the baseline (non-accelerated) results. The largest $\tau$ that maintains an acceptable quality level (e.g., SSIM > 0.7) is then selected as the final threshold.

\section{Experiments}
In this section, we first describe our experimental setup, then present the comparison of quantitative results, followed by the ablation study and further analysis.

\subsection{Experimental setup}

\noindent\textbf{Base Models and Baselines.} 
To verify the effectiveness of our method, we compare it against the most competitive methods across different flow matching models. Specifically, we apply our acceleration method to Wan2.1 \cite{DBLP:journals/corr/abs-2503-20314} and EasyAnimateV5.1 \cite{DBLP:journals/corr/abs-2405-18991}, each offered in two model scales with a shared VAE. Wan2.1 offers 14B and 1.3B variants, while EasyAnimateV5.1 provides 12B and 7B variants. We compare our method with two competitive baselines: the large-small model collaborative inference method, SRDiffusion \cite{DBLP:journals/corr/abs-2505-19151}, and the cache-based method, TeaCache \cite{DBLP:conf/cvpr/Liu0W0QZZY025}.

\noindent\textbf{Evaluation Metrics.}
For the quantitative evaluation of video generation acceleration methods, we focus on two key aspects: inference efficiency and visual quality. Following \cite{DBLP:journals/corr/abs-2505-19151,DBLP:conf/cvpr/Liu0W0QZZY025}, inference efficiency is assessed via per-sample inference latency, whereas visual quality is evaluated using VBench \cite{DBLP:conf/cvpr/HuangHYZS0Z0JCW24} in conjunction with three widely adopted metrics capturing perceptual consistency, pixel-level fidelity, and structural similarity: Learned Perceptual Image Patch Similarity (LPIPS) \cite{DBLP:conf/cvpr/ZhangIESW18}, Peak Signal-to-Noise Ratio (PSNR), and Structural Similarity Index Measure (SSIM) \cite{DBLP:journals/tip/WangBSS04}.

\noindent\textbf{Implementation Detail.}
Following the evaluation setup in \cite{DBLP:conf/cvpr/Liu0W0QZZY025}, we use the standard prompt set provided by VBench \cite{DBLP:conf/cvpr/HuangHYZS0Z0JCW24} for text-conditioned video generation. All experiments are conducted on NVIDIA A100 GPUs, with FlashAttention \cite{DBLP:conf/nips/DaoFERR22} enabled by default. Latency is measured on a single A100 GPU. Following \cite{DBLP:journals/corr/abs-2502-21079,DBLP:journals/corr/abs-2506-09045}, we preserve the first 20\% of the diffusion steps with the large model for TeaCache and our MDD, as these initial steps are critical to the overall generation process. In line with \cite{DBLP:conf/cvpr/Liu0W0QZZY025}, the error accumulation threshold for the TeaCache baseline is set to 0.2. According to \cite{DBLP:journals/corr/abs-2505-19151}, the threshold for transitioning from the large model to the small model in the SRDiffusion baseline is set to 0.1. For our MDD, we set the cumulative directional error threshold $\tau$ in Eq. (\ref{eq:th}) to 0.005. Furthermore, we employ the small model for the final 5\% of the steps to ensure the reliability of visual outputs. 

\subsection{Main Results}

% \noindent\textbf{Quantitative Comparison.} 
Table \ref{tab: main} provides a comprehensive quantitative evaluation of our method compared to the most competitive baselines. The results indicate that: (1) our MDD consistently achieves highest acceleration (2.95$\times$ for Wan2.1, 1.90$\times$ for EasyAnimateV5.1) while preserving high visual retention (e.g., LPIPS 0.178, PSNR 22.72, SSIM 0.748 on Wan2.1), demonstrating the effectiveness of our lightweight alternative strategy; (2) compared with the large-small model collaborative inference baseline SRDiffusion, MDD leverages direction-calibrated outputs from a lightweight model to more faithfully approximate of the original denoising trajectory, rather than directly using the biased directional predictions of the small model for acceleration. This design, further enhanced by CFG reuse, yields higher acceleration while delivering superior visual retention; and
(3) compared with the cache-based baseline TeaCache, MDD can be considered as adopting a magnitude-aware residual reuse strategy assisted by the lightweight model, rather than relying on invariant residual reuse. Consequently, our method can achieve higher reuse rates through more reliable lightweight substitutions, enabling faster acceleration while maintaining better visual fidelity.

\begin{table*}[t]
    \centering
    \small
    \renewcommand\arraystretch{1.02}
    \caption{Quantitative evaluation of inference efficiency and visual quality in video generation models. Visual retention metrics, including LPIPS, SSIM, and PSNR, are calculated against the original large models (Wan2.1-14B and EasyAnimateV5.1-12B).}
    \label{tab: main}
    \begin{tabular}{l|cc|cccc}
        \toprule
        \multirow{2}{*}{\textbf{Method}} & \multicolumn{2}{c|}{\textbf{Efficiency}} & \multicolumn{4}{c}{\textbf{Visual Quality}} \\ 
        \cline{2-7}
        & \textbf{Speedup$\uparrow$} & \textbf{Latency$\downarrow$} & \textbf{LPIPS$\downarrow$} & \textbf{PSNR$\uparrow$} & \textbf{SSIM$\uparrow$} & \textbf{VBench$\uparrow$} \\
        \midrule
        \midrule
        \multicolumn{7}{c}{\textbf{Wan2.1} (832$\times$480, 81 frames, $T=50$)} \\
        \midrule
        \rowcolor[gray]{0.9} Wan2.1-14B & 1$\times$ & 948~s & - & - & - & 83.09\% \\
        \rowcolor[gray]{0.9} Wan2.1-1.3B & - & 192~s & 0.597 & 11.81 & 0.336 & 81.00\% \\
        TeaCache \cite{DBLP:conf/cvpr/Liu0W0QZZY025} & 2.62$\times$ & 362~s & 0.208 & 20.95 & 0.680 & 82.41\% \\
        SRDiffusion \cite{DBLP:journals/corr/abs-2505-19151} & 2.75$\times$ & 344~s & 0.240 & 19.10 & 0.636 & 83.01\% \\
        \midrule
        \textbf{MDD (Ours)} & \textbf{2.95$\times$} & \textbf{321~s} & \textbf{0.178} & \textbf{22.72} & \textbf{0.748} & 82.62\% \\

        \midrule
        \midrule
        \multicolumn{7}{c}{\textbf{EasyAnimateV5.1} (672$\times$384, 49 frames, $T=50$)} \\
        \midrule
        \rowcolor[gray]{0.9} EasyAnimateV5.1-12B & 1$\times$ & 246~s & - & - & -  & 78.89\% \\
        \rowcolor[gray]{0.9} EasyAnimateV5.1-7B & - & 133~s & 0.622 & 12.90 & 0.404 & 75.71\% \\
        TeaCache \cite{DBLP:conf/cvpr/Liu0W0QZZY025} & 1.84$\times$ & 134~s & 0.179 & 21.30 & 0.708 & 78.53\% \\
        SRDiffusion \cite{DBLP:journals/corr/abs-2505-19151} & 1.57$\times$ & 157~s & 0.380 & 17.75 & 0.540 & 78.24\% \\
        \midrule
        \textbf{MDD (Ours)} & \textbf{1.90$\times$} & \textbf{129~s} & \textbf{0.150} & \textbf{22.66} & \textbf{0.755} & 78.65\% \\

        \bottomrule
    \end{tabular}
\end{table*}

% \noindent\textbf{Visualization Results.}
% Figure~\ref{fig:compare} and Appendix~\ref{apd: vis} showcase a comparison of video generation results from the original Wan2.1-14B model, TeaCache \cite{DBLP:conf/cvpr/Liu0W0QZZY025}, SRDiffusion \cite{DBLP:journals/corr/abs-2505-19151}, and our MDD. The results show that: (1) in pursuit of higher acceleration, TeaCache accumulates excessive errors due to residual reuse across multiple steps, ultimately degrading the quality of the generated videos; (2) SRDiffusion relies on a less capable lightweight model for content rendering, which can deviate from the denoising trajectory of the original large model, leading to the loss of content details and suboptimal visual retention; and (3) our method more faithfully preserves the denoising trajectory of the original large model, achieving higher fidelity in the generated details while delivering greater acceleration.

\subsection{Ablation Study}

\noindent\textbf{The Effect of Threshold $\tau$.} 
Table \ref{tab:ablation} examines the impact of varying the threshold $\tau$ in Eq. (\ref{eq:th}) on inference efficiency and visual quality. The results indicate that: (1) A higher $\tau$ allows greater accumulation of directional error, enabling the lightweight model to perform more substitution steps and thus increasing acceleration. However, excessive error may adversely affect performance. For example, setting $\tau = 0.010$ yields a 3.12× speedup but reduces fidelity (LPIPS = 0.207, PSNR = 22.21, SSIM = 0.728). (2) A lower $\tau$ limits error accumulation, facilitating timely corrections by the original large model and cache updates, which reduces efficiency but improves quality. For example, $\tau = 0.001$ results in LPIPS = 0.131, PSNR = 23.76, SSIM = 0.786, with a reduced speedup of 2.21×. (3) A moderate value, $\tau = 0.005$, provides a favorable trade-off, achieving a 2.95× speedup while maintaining acceptable visual fidelity (LPIPS = 0.178, PSNR = 22.72, SSIM = 0.748).

\begin{table*}[t]
    \centering
    \small
    \renewcommand\arraystretch{1.05}
    \caption{Ablation study of threshold $\tau$ in Eq. (\ref{eq:th}) on Wan2.1-14B. A larger threshold enables a greater proportion of steps to be approximated by the lightweight small-model, yielding higher acceleration at the expense of visual quality.}
    \label{tab:ablation}
    \begin{tabular}{l|cc|cccc}
        \toprule
        \multirow{2}{*}{\textbf{Threshold $\tau$}} & \multicolumn{2}{c|}{\textbf{Efficiency}} & \multicolumn{4}{c}{\textbf{Visual Quality}} \\ 
        \cline{2-7}
        & \textbf{Speedup$\uparrow$} & \textbf{Latency$\downarrow$} & \textbf{LPIPS$\downarrow$} & \textbf{PSNR$\uparrow$} & \textbf{SSIM$\uparrow$} & \textbf{VBench$\uparrow$} \\
        \midrule
        \midrule
        \rowcolor[gray]{0.9} Wan2.1-14B & 1$\times$ & 948~s & - & - & - & 83.09\% \\
        $\tau = 0.001$ & 2.21$\times$ & 428~s & 0.131 & 23.76 & 0.786 & 82.74\% \\
        $\tau = 0.002$ & 2.76$\times$ & 344~s & 0.148 & 23.33 & 0.771 & 82.85\% \\
        $\tau = 0.005$ & 2.95$\times$ & 321~s & 0.178 & 22.72 & 0.748 & 82.62\% \\
        $\tau = 0.010$ & 3.12$\times$ & 303~s & 0.207 & 22.21 & 0.728 & 82.20\% \\
        \bottomrule
    \end{tabular}
\end{table*}

\noindent\textbf{The Effect of Rescaling.} 
Table~\ref{tab: rescaling} presents an ablation study evaluating the impact of applying a rescaling strategy using a lightweight model. Relative to MMD without rescaling, incorporating the rescaling strategy leads to consistent improvements across all metrics, resulting in better visual fidelity and overall output quality. These results indicate that rescaling effectively compensates for the limitations imposed by the invariant residual caching strategy, yielding performance gains and validating the effectiveness of the proposed rescaling approach.

\begin{table*}[t]
    \centering
    \small
    \renewcommand\arraystretch{1.05}
    \caption{Ablation study of rescaling with the lightweight model on Wan2.1-14B.}
    \label{tab: rescaling}
    \begin{tabular}{l|cccc}
        \toprule
        \textbf{Methods} & \textbf{LPIPS$\downarrow$} & \textbf{SSIM$\uparrow$} & \textbf{PSNR$\uparrow$} & \textbf{VBench$\uparrow$} \\
        \midrule
        \midrule
        MDD w/o rescaling & 0.197 & 21.40 & 0.689 & 81.16\% \\
        MDD & 0.178 & 22.72 & 0.748 & 82.62\% \\
        \bottomrule
    \end{tabular}
\end{table*}

\noindent\textbf{The Effect of CFG Reuse.} 
Table \ref{tab: cfg} compares the effect of reusing the magnitude of the conditional output for the unconditional output in CFG. The results demonstrate that: (1) in terms of inference latency, reusing the magnitude of the conditional output for the unconditional output can reduce the inference cost of small models by half, providing an additional 1.20× speedup in overall inference and further decreasing computational overhead; and (2) in terms of visual quality, since the magnitudes of the conditional and unconditional outputs in CFG are nearly identical, reusing the magnitude effectively preserves the original visual quality. Quantitative evaluations demonstrate that the performance exhibits a negligible decline after CFG reuse, further confirming the effectiveness of this acceleration strategy.

\begin{table*}[t]
    \centering
    \small
    \renewcommand\arraystretch{1.05}
    \caption{Ablation study of CFG Reuse on Wan2.1. Visual retention metrics (LPIPS, SSIM, and PSNR) are computed relative to the Wan2.1-14B model, with $\tau$ set to 0.005.}
    \label{tab: cfg}
    \begin{tabular}{c|cc|cccc}
        \toprule
        \multirow{2}{*}{\textbf{CFG Reuse}} & \multicolumn{2}{c|}{\textbf{Efficiency}} & \multicolumn{4}{c}{\textbf{Visual Quality}} \\ 
        \cline{2-7}
        & \textbf{Speedup$\uparrow$} & \textbf{Latency$\downarrow$} & \textbf{LPIPS$\downarrow$} & \textbf{SSIM$\uparrow$} & \textbf{PSNR$\uparrow$} & \textbf{VBench$\uparrow$} \\
        \midrule
        \midrule
        $\times$ & - & 385~s & 0.177 & 22.73 & 0.750 & 82.64\% \\
        $\checkmark$ & 1.20$\times$ & 321~s & 0.178 & 22.72 & 0.748 & 82.62\% \\
        \bottomrule
    \end{tabular}
\end{table*}

\subsection{Further Analysis}

\noindent\textbf{Scaling to 720p resolution.} Given the substantial computational cost of high-resolution video generation, we evaluate the scalability of our method on Wan2.1-14B using a randomly selected subset of 200 prompts from VBench. As shown in Table~\ref{tab: 720}, our approach demonstrates strong scalability to 720p resolution, maintaining robust performance at higher resolutions. Notably, MDD achieves up to a 2.77$\times$ speedup while consistently outperforming prior acceleration methods in terms of LPIPS, SSIM, and PSNR. These results indicate that the efficiency gains are achieved while preserving high visual fidelity and perceptual similarity.

\begin{table*}[t]
    \centering
    \small
    \renewcommand\arraystretch{1.05}
    \caption{Results for Wan2.1-14B 720p, with latency measured under an 8$\times$NVIDIA A100 GPU setup.}
    \label{tab: 720}
    \begin{tabular}{l|cc|ccc}
        \toprule
        \multirow{2}{*}{\textbf{Interval}} & \multicolumn{2}{c|}{\textbf{Efficiency}} & \multicolumn{3}{c}{\textbf{Visual Quality}} \\ 
        \cline{2-6}
        & \textbf{Speedup$\uparrow$} & \textbf{Latency$\downarrow$} & \textbf{LPIPS$\downarrow$} & \textbf{SSIM$\uparrow$} & \textbf{PSNR$\uparrow$} \\
        \midrule
        \midrule
        Wan2.1-14B & 1$\times$ & 476~s & - & - & - \\
        TeaCache \cite{DBLP:conf/cvpr/Liu0W0QZZY025} & 2.37$\times$ & 201~s & 0.367 & 17.54 & 0.607 \\
        SRDiffusion \cite{DBLP:journals/corr/abs-2505-19151} & 2.47$\times$ & 193~s & 0.381 & 16.75 & 0.585 \\
        Ours ($\tau$ = 0.002) & 2.50$\times$ & 190~s & 0.276 & 18.93 & 0.661 \\
        Ours ($\tau$ = 0.005) & 2.77$\times$ & 172~s & 0.292 & 18.48 & 0.655 \\
        \bottomrule
    \end{tabular}
\end{table*}

\noindent\textbf{Compatibility of MDD with Efficient ODE Solvers.} 
Efficient ODE solvers aim to achieve high-quality sampling with fewer steps, whereas our method reduces inference cost by replacing parts of the original large-model sampling process with lightweight substitutes. Table \ref{tab: ode} presents a supplementary analysis of step reduction using MDD with the UniPC \cite{DBLP:conf/nips/ZhaoBR0L23} and DPM++ \cite{DBLP:journals/ijautcomp/LuZBCLZ25} solvers, demonstrating its compatibility with efficient ODE solvers. 

\begin{table*}[t]
    \centering
    \small
    \renewcommand\arraystretch{1.05}
    \caption{Empirical analysis of lightweight substitution intervals.}
    \label{tab: ode}
    \resizebox{\textwidth}{!}{
    \begin{tabular}{l|l|cc|cccc}
        \toprule
        \multirow{2}{*}{\textbf{Methods}} & \multirow{2}{*}{\textbf{ODE solver}} & \multicolumn{2}{c|}{\textbf{Efficiency}} & \multicolumn{4}{c}{\textbf{Visual Quality}} \\ 
        \cline{3-8}
        & & \textbf{Speedup$\uparrow$} & \textbf{Latency$\downarrow$} & \textbf{LPIPS$\downarrow$} & \textbf{SSIM$\uparrow$} & \textbf{PSNR$\uparrow$} & \textbf{VBench$\uparrow$} \\
        \midrule
        \midrule
        Wan2.1-14B (T=50) & UniPC & 1$\times$ & 948~s & - & - & - & 83.09\% \\
        Ours (T=50) & UniPC & 2.95$\times$ & 321~s & 0.178 & 22.72 & 0.748 & 82.62\% \\
        \midrule
        Wan2.1-14B (T=40) & DPM++ & 1$\times$ & 766~s & - & - & - & 82.86\% \\
        Ours (T=40) & DPM++ & 2.68$\times$ & 286~s & 0.169 & 23.00 & 0.765 & 82.45\% \\
        \bottomrule
    \end{tabular}
    }
\end{table*}

\noindent\textbf{Analysis of MDD in Early Denoising.}
In the early denoising stage, the model needs to synthesize semantic content from pure noise \cite{DBLP:journals/corr/abs-2505-19151}. Unlike the intermediate stage (20\%–95\%), the initial phase is critical and exhibits low redundancy, as confirmed by prior post-hoc acceleration studies \cite{DBLP:journals/corr/abs-2505-19151,DBLP:conf/cvpr/Liu0W0QZZY025,DBLP:conf/iclr/LvSSY00W25}. Consequently, the directional component, which relies on redundancy-based residual reuse, is constrained at this point. In contrast, magnitude estimation remains relatively stable, as it is derived from lightweight neural networks. Quantitative results demonstrating the effects of applying lightweight substitutions in the early denoising stage are presented in Table \ref{tab: Interval}. The results show that lightweight substitutions in these critical early steps incur substantial performance degradation. Overall, it is generally advisable to rely on the original large model for reliable generation during the early denoising phase, a strategy that has also been widely adopted in post-hoc acceleration methods \cite{DBLP:journals/corr/abs-2505-19151,DBLP:conf/cvpr/Liu0W0QZZY025,DBLP:conf/iclr/LvSSY00W25}.

\begin{table*}[t]
    \centering
    \small
    \renewcommand\arraystretch{1.05}
    \caption{Compatibility analysis results between MDD and efficient ODE solvers.}
    \label{tab: Interval}
    \begin{tabular}{l|cc|cccc}
        \toprule
        \multirow{2}{*}{\textbf{Interval}} & \multicolumn{2}{c|}{\textbf{Efficiency}} & \multicolumn{4}{c}{\textbf{Visual Quality}} \\ 
        \cline{2-7}
        & \textbf{Speedup$\uparrow$} & \textbf{Latency$\downarrow$} & \textbf{LPIPS$\downarrow$} & \textbf{SSIM$\uparrow$} & \textbf{PSNR$\uparrow$} & \textbf{VBench$\uparrow$} \\
        \midrule
        \midrule
        0\%–95\% & 3.85$\times$ & 246~s & 0.567 & 13.43 & 0.431 & 79.02\% \\
        20\%–95\% (Ours) & 2.95$\times$ & 321~s & 0.178 & 22.72 & 0.748 & 82.62\% \\
        \bottomrule
    \end{tabular}
\end{table*}

\noindent
\begin{adjustbox}{valign=t} 
\begin{minipage}[t]{0.52\textwidth} 
\textbf{Acceleration in Multi-GPU Setups.} 
In line with prior work \cite{DBLP:journals/corr/abs-2503-20314}, we assess the performance of MDD under the same multi-GPU context-parallel configuration. Figure~\ref{fig:multigpu} presents results on Wan2.1-14B, comparing the latency of MDD with SRDiffusion \cite{DBLP:journals/corr/abs-2505-19151} and TeaCache \cite{DBLP:conf/cvpr/Liu0W0QZZY025} on A100 GPUs. Experimental results show that, when scaled across multiple GPUs, MDD consistently achieves higher inference efficiency than both SRDiffusion and TeaCache, highlighting the effectiveness of the MDD inference strategy in multi-GPU settings.
\end{minipage}
\end{adjustbox}%
\hfill
\begin{adjustbox}{valign=t} % 顶端对齐
\begin{minipage}[t]{0.46\textwidth} % 右侧图片
\centering
\includegraphics[width=\linewidth]{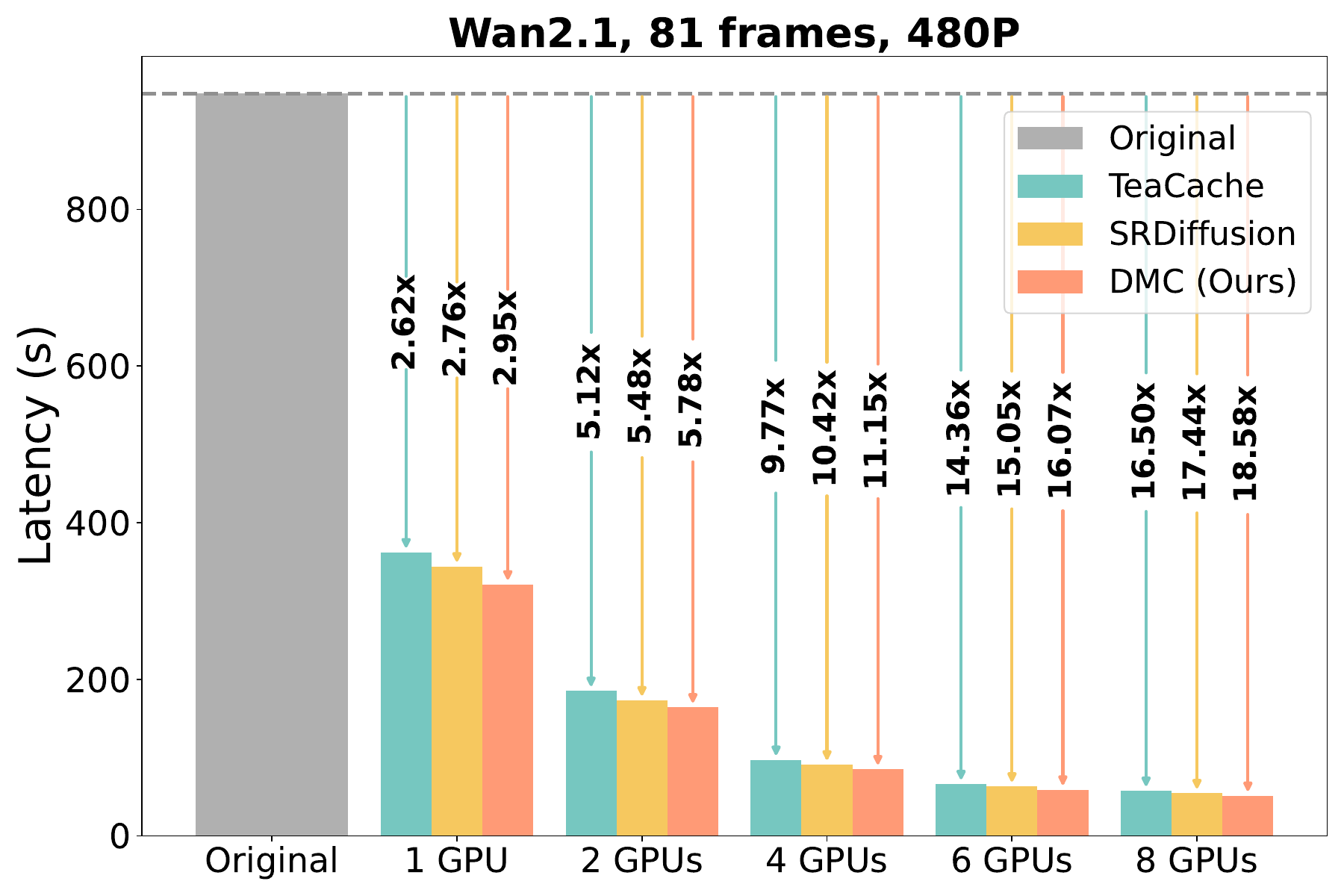} 
\captionof{figure}{Multi-GPU inference efficiency analysis.} 
\label{fig:multigpu}
\end{minipage}
\end{adjustbox}

\section{Conclusion}
This study explores lightweight alternatives to accelerate flow-based video generation. Our analysis shows that within the same model family, the small model effectively captures the large model’s magnitude information, while residual reuse provides reliable directional guidance. Based on this, we propose MDD, a training-free method that speeds up inference by replacing part of the denoising steps with lightweight alternatives. Specifically, MDD adaptively integrates magnitude estimates from the small model with residual-based directional guidance to approximate the large model’s denoising trajectory, and further reduces computational cost via CFG reuse. Extensive experiments show that MDD achieves substantial acceleration while maintaining high visual fidelity, offering a practical solution for faster sampling of flow matching models.

\bibliographystyle{splncs04}
\bibliography{main}

@inproceedings{DBLP:conf/iclr/LipmanCBNL23,
  author       = {Yaron Lipman and
                  Ricky T. Q. Chen and
                  Heli Ben{-}Hamu and
                  Maximilian Nickel and
                  Matthew Le},
  title        = {Flow Matching for Generative Modeling},
  booktitle    = {ICLR},
  year         = {2023},
}

@inproceedings{DBLP:conf/nips/DhariwalN21,
  author       = {Prafulla Dhariwal and
                  Alexander Quinn Nichol},
  title        = {Diffusion Models Beat GANs on Image Synthesis},
  booktitle    = {NeurIPS},
  pages        = {8780--8794},
  year         = {2021},
}

@inproceedings{DBLP:conf/nips/HoJA20,
  author       = {Jonathan Ho and
                  Ajay Jain and
                  Pieter Abbeel},
  title        = {Denoising Diffusion Probabilistic Models},
  booktitle    = {NeurIPS},
  year         = {2020},
}

@inproceedings{DBLP:conf/iclr/YangTZ00XYHZFYZ25,
  author       = {Zhuoyi Yang and
                  Jiayan Teng and
                  Wendi Zheng and
                  Ming Ding and
                  Shiyu Huang and
                  Jiazheng Xu and
                  Yuanming Yang and
                  Wenyi Hong and
                  Xiaohan Zhang and
                  Guanyu Feng and
                  Da Yin and
                  Yuxuan Zhang and
                  Weihan Wang and
                  Yean Cheng and
                  Bin Xu and
                  Xiaotao Gu and
                  Yuxiao Dong and
                  Jie Tang},
  title        = {CogVideoX: Text-to-Video Diffusion Models with An Expert Transformer},
  booktitle    = {ICLR},
  year         = {2025},
}

@article{DBLP:journals/corr/abs-2412-20404,
  author       = {Zangwei Zheng and
                  Xiangyu Peng and
                  Tianji Yang and
                  Chenhui Shen and
                  Shenggui Li and
                  Hongxin Liu and
                  Yukun Zhou and
                  Tianyi Li and
                  Yang You},
  title        = {Open-Sora: Democratizing Efficient Video Production for All},
  journal      = {CoRR},
  volume       = {abs/2412.20404},
  year         = {2024},
}

@article{DBLP:journals/corr/abs-2412-03603,
  author       = {Weijie Kong and
                  others},
  title        = {HunyuanVideo: {A} Systematic Framework For Large Video Generative
                  Models},
  journal      = {CoRR},
  volume       = {abs/2412.03603},
  year         = {2024},
}

@article{DBLP:journals/corr/abs-2503-20314,
  author       = {Ang Wang and
                  others},
  title        = {Wan: Open and Advanced Large-Scale Video Generative Models},
  journal      = {CoRR},
  volume       = {abs/2503.20314},
  year         = {2025},
}

@article{DBLP:journals/tmlr/Ma0CJ0LC025,
  author       = {Xin Ma and
                  Yaohui Wang and
                  Xinyuan Chen and
                  Gengyun Jia and
                  Ziwei Liu and
                  Yuan{-}Fang Li and
                  Cunjian Chen and
                  Yu Qiao},
  title        = {Latte: Latent Diffusion Transformer for Video Generation},
  journal      = {Trans. Mach. Learn. Res.},
  volume       = {2025},
  year         = {2025},
}

@inproceedings{DBLP:conf/cvpr/MengRGKEHS23,
  author       = {Chenlin Meng and
                  Robin Rombach and
                  Ruiqi Gao and
                  Diederik P. Kingma and
                  Stefano Ermon and
                  Jonathan Ho and
                  Tim Salimans},
  title        = {On Distillation of Guided Diffusion Models},
  booktitle    = {CVPR},
  pages        = {14297--14306},
  year         = {2023},
}

@inproceedings{DBLP:conf/cvpr/ChenMTMJWW025,
  author       = {Lei Chen and
                  Yuan Meng and
                  Chen Tang and
                  Xinzhu Ma and
                  Jingyan Jiang and
                  Xin Wang and
                  Zhi Wang and
                  Wenwu Zhu},
  title        = {Q-DiT: Accurate Post-Training Quantization for Diffusion Transformers},
  booktitle    = {CVPR},
  pages        = {28306--28315},
  year         = {2025},
}

@article{DBLP:journals/corr/abs-2505-19151,
  author       = {Shenggan Cheng and
                  Yuanxin Wei and
                  Lansong Diao and
                  Yong Liu and
                  Bujiao Chen and
                  Lianghua Huang and
                  Yu Liu and
                  Wenyuan Yu and
                  Jiangsu Du and
                  Wei Lin and
                  Yang You},
  title        = {SRDiffusion: Accelerate Video Diffusion Inference via Sketching-Rendering
                  Cooperation},
  journal      = {CoRR},
  volume       = {abs/2505.19151},
  year         = {2025},
}

@inproceedings{DBLP:conf/cvpr/Liu0W0QZZY025,
  author       = {Feng Liu and
                  Shiwei Zhang and
                  Xiaofeng Wang and
                  Yujie Wei and
                  Haonan Qiu and
                  Yuzhong Zhao and
                  Yingya Zhang and
                  Qixiang Ye and
                  Fang Wan},
  title        = {Timestep Embedding Tells: It's Time to Cache for Video Diffusion Model},
  booktitle    = {CVPR},
  pages        = {7353--7363},
  year         = {2025},
}

@inproceedings{DBLP:conf/cvpr/MaFW24,
  author       = {Xinyin Ma and
                  Gongfan Fang and
                  Xinchao Wang},
  title        = {DeepCache: Accelerating Diffusion Models for Free},
  booktitle    = {CVPR},
  pages        = {15762--15772},
  year         = {2024},
}

@inproceedings{DBLP:conf/cvpr/WimbauerWSDHHSZ24,
  author       = {Felix Wimbauer and
                  Bichen Wu and
                  Edgar Sch{\"{o}}nfeld and
                  Xiaoliang Dai and
                  Ji Hou and
                  Zijian He and
                  Artsiom Sanakoyeu and
                  Peizhao Zhang and
                  Sam S. Tsai and
                  Jonas Kohler and
                  Christian Rupprecht and
                  Daniel Cremers and
                  Peter Vajda and
                  Jialiang Wang},
  title        = {Cache Me if You Can: Accelerating Diffusion Models through Block Caching},
  booktitle    = {CVPR},
  pages        = {6211--6220},
  year         = {2024},
}

@inproceedings{DBLP:conf/nips/LiWJHC0WTR23,
  author       = {Yanyu Li and
                  Huan Wang and
                  Qing Jin and
                  Ju Hu and
                  Pavlo Chemerys and
                  Yun Fu and
                  Yanzhi Wang and
                  Sergey Tulyakov and
                  Jian Ren},
  title        = {SnapFusion: Text-to-Image Diffusion Model on Mobile Devices within
                  Two Seconds},
  booktitle    = {NeurIPS},
  year         = {2023},
}

@inproceedings{DBLP:conf/eccv/ChenGXWYRWLLL24,
  author       = {Junsong Chen and
                  Chongjian Ge and
                  Enze Xie and
                  Yue Wu and
                  Lewei Yao and
                  Xiaozhe Ren and
                  Zhongdao Wang and
                  Ping Luo and
                  Huchuan Lu and
                  Zhenguo Li},
  title        = {PIXART-{\(\Sigma\)}: Weak-to-Strong Training of Diffusion Transformer
                  for 4K Text-to-Image Generation},
  booktitle    = {ECCV},
  volume       = {15090},
  pages        = {74--91},
  year         = {2024},
}

@inproceedings{DBLP:conf/icml/Sohl-DicksteinW15,
  author       = {Jascha Sohl{-}Dickstein and
                  Eric A. Weiss and
                  Niru Maheswaranathan and
                  Surya Ganguli},
  title        = {Deep Unsupervised Learning using Nonequilibrium Thermodynamics},
  booktitle    = {ICML},
  volume       = {37},
  pages        = {2256--2265},
  year         = {2015},
}

@inproceedings{DBLP:conf/cvpr/ChenZCXWWS24,
  author       = {Haoxin Chen and
                  Yong Zhang and
                  Xiaodong Cun and
                  Menghan Xia and
                  Xintao Wang and
                  Chao Weng and
                  Ying Shan},
  title        = {VideoCrafter2: Overcoming Data Limitations for High-Quality Video
                  Diffusion Models},
  booktitle    = {CVPR},
  pages        = {7310--7320},
  year         = {2024},
}

@article{DBLP:journals/corr/abs-2311-15127,
  author       = {Andreas Blattmann and
                  Tim Dockhorn and
                  Sumith Kulal and
                  Daniel Mendelevitch and
                  Maciej Kilian and
                  Dominik Lorenz and
                  Yam Levi and
                  Zion English and
                  Vikram Voleti and
                  Adam Letts and
                  Varun Jampani and
                  Robin Rombach},
  title        = {Stable Video Diffusion: Scaling Latent Video Diffusion Models to Large
                  Datasets},
  journal      = {CoRR},
  volume       = {abs/2311.15127},
  year         = {2023},
}

@inproceedings{DBLP:conf/cvpr/ShangYXW023,
  author       = {Yuzhang Shang and
                  Zhihang Yuan and
                  Bin Xie and
                  Bingzhe Wu and
                  Yan Yan},
  title        = {Post-Training Quantization on Diffusion Models},
  booktitle    = {CVPR},
  pages        = {1972--1981},
  year         = {2023},
}

@inproceedings{DBLP:conf/nips/HeLLWZZ23,
  author       = {Yefei He and
                  Luping Liu and
                  Jing Liu and
                  Weijia Wu and
                  Hong Zhou and
                  Bohan Zhuang},
  title        = {{PTQD:} Accurate Post-Training Quantization for Diffusion Models},
  booktitle    = {NeurIPS},
  year         = {2023},
}

@inproceedings{DBLP:conf/cvpr/Wang0XT0L24,
  author       = {Changyuan Wang and
                  Ziwei Wang and
                  Xiuwei Xu and
                  Yansong Tang and
                  Jie Zhou and
                  Jiwen Lu},
  title        = {Towards Accurate Post-Training Quantization for Diffusion Models},
  booktitle    = {CVPR},
  pages        = {16026--16035},
  year         = {2024},
}

@inproceedings{DBLP:conf/iclr/SalimansH22,
  author       = {Tim Salimans and
                  Jonathan Ho},
  title        = {Progressive Distillation for Fast Sampling of Diffusion Models},
  booktitle    = {ICLR},
  year         = {2022},
}

@article{DBLP:journals/corr/abs-2403-12706,
  author       = {Shanchuan Lin and
                  Xiao Yang},
  title        = {AnimateDiff-Lightning: Cross-Model Diffusion Distillation},
  journal      = {CoRR},
  volume       = {abs/2403.12706},
  year         = {2024},
}

@inproceedings{DBLP:conf/eccv/SauerLBR24,
  author       = {Axel Sauer and
                  Dominik Lorenz and
                  Andreas Blattmann and
                  Robin Rombach},
  title        = {Adversarial Diffusion Distillation},
  booktitle    = {ECCV},
  volume       = {15144},
  pages        = {87--103},
  year         = {2024},
}

@inproceedings{DBLP:conf/icml/SongD0S23,
  author       = {Yang Song and
                  Prafulla Dhariwal and
                  Mark Chen and
                  Ilya Sutskever},
  title        = {Consistency Models},
  booktitle    = {ICML},
  volume       = {202},
  pages        = {32211--32252},
  year         = {2023},
}

@article{DBLP:journals/corr/abs-2310-04378,
  author       = {Simian Luo and
                  Yiqin Tan and
                  Longbo Huang and
                  Jian Li and
                  Hang Zhao},
  title        = {Latent Consistency Models: Synthesizing High-Resolution Images with
                  Few-Step Inference},
  journal      = {CoRR},
  volume       = {abs/2310.04378},
  year         = {2023},
}

@article{DBLP:journals/corr/abs-2503-06923,
  author       = {Jiacheng Liu and
                  Chang Zou and
                  Yuanhuiyi Lyu and
                  Junjie Chen and
                  Linfeng Zhang},
  title        = {From Reusing to Forecasting: Accelerating Diffusion Models with TaylorSeers},
  journal      = {CoRR},
  volume       = {abs/2503.06923},
  year         = {2025},
}

@article{DBLP:journals/corr/abs-2504-10540,
  author       = {Zichao Yu and
                  Zhen Zou and
                  Guojiang Shao and
                  Chengwei Zhang and
                  Shengze Xu and
                  Jie Huang and
                  Feng Zhao and
                  Xiaodong Cun and
                  Wenyi Zhang},
  title        = {AB-Cache: Training-Free Acceleration of Diffusion Models via Adams-Bashforth
                  Cached Feature Reuse},
  journal      = {CoRR},
  volume       = {abs/2504.10540},
  year         = {2025},
}

@article{DBLP:journals/corr/abs-2506-09045,
  author       = {Zehong Ma and
                  Longhui Wei and
                  Feng Wang and
                  Shiliang Zhang and
                  Qi Tian},
  title        = {MagCache: Fast Video Generation with Magnitude-Aware Cache},
  journal      = {CoRR},
  volume       = {abs/2506.09045},
  year         = {2025},
}

@inproceedings{DBLP:conf/iclr/SongME21,
  author       = {Jiaming Song and
                  Chenlin Meng and
                  Stefano Ermon},
  title        = {Denoising Diffusion Implicit Models},
  booktitle    = {ICLR},
  year         = {2021},
}

@inproceedings{DBLP:conf/nips/SongE19,
  author       = {Yang Song and
                  Stefano Ermon},
  title        = {Generative Modeling by Estimating Gradients of the Data Distribution},
  booktitle    = {NeurIPS},
  pages        = {11895--11907},
  year         = {2019},
}

@inproceedings{DBLP:conf/nips/KarrasAAL22,
  author       = {Tero Karras and
                  Miika Aittala and
                  Timo Aila and
                  Samuli Laine},
  title        = {Elucidating the Design Space of Diffusion-Based Generative Models},
  booktitle    = {NeurIPS},
  year         = {2022},
}

@inproceedings{DBLP:conf/nips/0011ZB0L022,
  author       = {Cheng Lu and
                  Yuhao Zhou and
                  Fan Bao and
                  Jianfei Chen and
                  Chongxuan Li and
                  Jun Zhu},
  title        = {DPM-Solver: {A} Fast {ODE} Solver for Diffusion Probabilistic Model
                  Sampling in Around 10 Steps},
  booktitle    = {NeurIPS},
  year         = {2022},
}

@article{DBLP:journals/corr/abs-2505-20353,
  author       = {Dong Liu and
                  Jiayi Zhang and
                  Yifan Li and
                  Yanxuan Yu and
                  Ben Lengerich and
                  Ying Nian Wu},
  title        = {FastCache: Fast Caching for Diffusion Transformer Through Learnable
                  Linear Approximation},
  journal      = {CoRR},
  volume       = {abs/2505.20353},
  year         = {2025},
}

@article{DBLP:journals/corr/abs-2406-01125,
  author       = {Pengtao Chen and
                  Mingzhu Shen and
                  Peng Ye and
                  Jianjian Cao and
                  Chongjun Tu and
                  Christos{-}Savvas Bouganis and
                  Yiren Zhao and
                  Tao Chen},
  title        = {{\(\Delta\)}-DiT: {A} Training-Free Acceleration Method Tailored for
                  Diffusion Transformers},
  journal      = {CoRR},
  volume       = {abs/2406.01125},
  year         = {2024},
}

@article{DBLP:journals/corr/abs-2411-02397,
  author       = {Kumara Kahatapitiya and
                  Haozhe Liu and
                  Sen He and
                  Ding Liu and
                  Menglin Jia and
                  Chenyang Zhang and
                  Michael S. Ryoo and
                  Tian Xie},
  title        = {Adaptive Caching for Faster Video Generation with Diffusion Transformers},
  journal      = {CoRR},
  volume       = {abs/2411.02397},
  year         = {2024},
}

@article{DBLP:journals/corr/abs-2405-18991,
  author       = {Jiaqi Xu and
                  Xinyi Zou and
                  Kunzhe Huang and
                  Yunkuo Chen and
                  Bo Liu and
                  Mengli Cheng and
                  Xing Shi and
                  Jun Huang},
  title        = {EasyAnimate: {A} High-Performance Long Video Generation Method based
                  on Transformer Architecture},
  journal      = {CoRR},
  volume       = {abs/2405.18991},
  year         = {2024},
}

@inproceedings{DBLP:conf/iccv/LiLLYDKZK23,
  author       = {Xiuyu Li and
                  Yijiang Liu and
                  Long Lian and
                  Huanrui Yang and
                  Zhen Dong and
                  Daniel Kang and
                  Shanghang Zhang and
                  Kurt Keutzer},
  title        = {Q-Diffusion: Quantizing Diffusion Models},
  booktitle    = {ICCV},
  pages        = {17489--17499},
  year         = {2023},
}

@inproceedings{DBLP:conf/iclr/LiuG023,
  author       = {Xingchao Liu and
                  Chengyue Gong and
                  Qiang Liu},
  title        = {Flow Straight and Fast: Learning to Generate and Transfer Data with
                  Rectified Flow},
  booktitle    = {ICLR},
  year         = {2023},
}

@book{suli2003introduction,
  title={An introduction to numerical analysis},
  author={S{\"u}li, Endre and Mayers, David F},
  year={2003},
  publisher={Cambridge University Press}
}

@inproceedings{DBLP:conf/cvpr/HuangHYZS0Z0JCW24,
  author       = {Ziqi Huang and
                  Yinan He and
                  Jiashuo Yu and
                  Fan Zhang and
                  Chenyang Si and
                  Yuming Jiang and
                  Yuanhan Zhang and
                  Tianxing Wu and
                  Qingyang Jin and
                  Nattapol Chanpaisit and
                  Yaohui Wang and
                  Xinyuan Chen and
                  Limin Wang and
                  Dahua Lin and
                  Yu Qiao and
                  Ziwei Liu},
  title        = {VBench: Comprehensive Benchmark Suite for Video Generative Models},
  booktitle    = {CVPR},
  pages        = {21807--21818},
  year         = {2024},
}

@inproceedings{DBLP:conf/cvpr/ZhangIESW18,
  author       = {Richard Zhang and
                  Phillip Isola and
                  Alexei A. Efros and
                  Eli Shechtman and
                  Oliver Wang},
  title        = {The Unreasonable Effectiveness of Deep Features as a Perceptual Metric},
  booktitle    = {CVPR},
  pages        = {586--595},
  year         = {2018},
}

@inproceedings{DBLP:conf/nips/DaoFERR22,
  author       = {Tri Dao and
                  Daniel Y. Fu and
                  Stefano Ermon and
                  Atri Rudra and
                  Christopher R{\'{e}}},
  title        = {FlashAttention: Fast and Memory-Efficient Exact Attention with IO-Awareness},
  booktitle    = {NeurIPS},
  year         = {2022},
}

@article{DBLP:journals/corr/abs-2502-21079,
  author       = {Yifei Xia and
                  Suhan Ling and
                  Fangcheng Fu and
                  Yujie Wang and
                  Huixia Li and
                  Xuefeng Xiao and
                  Bin Cui},
  title        = {Training-free and Adaptive Sparse Attention for Efficient Long Video
                  Generation},
  journal      = {CoRR},
  volume       = {abs/2502.21079},
  year         = {2025},
}

@article{DBLP:journals/corr/abs-2207-12598,
  author       = {Jonathan Ho and
                  Tim Salimans},
  title        = {Classifier-Free Diffusion Guidance},
  journal      = {CoRR},
  volume       = {abs/2207.12598},
  year         = {2022},
}

@article{DBLP:journals/tip/WangBSS04,
  author       = {Zhou Wang and
                  Alan C. Bovik and
                  Hamid R. Sheikh and
                  Eero P. Simoncelli},
  title        = {Image quality assessment: from error visibility to structural similarity},
  journal      = {{IEEE} Trans. Image Process.},
  volume       = {13},
  number       = {4},
  pages        = {600--612},
  year         = {2004},
}

@inproceedings{DBLP:conf/iclr/YangCW0C24,
  author       = {Shuai Yang and
                  Yukang Chen and
                  Luozhou Wang and
                  Shu Liu and
                  Ying{-}Cong Chen},
  title        = {Denoising Diffusion Step-aware Models},
  booktitle    = {ICLR},
  year         = {2024},
}

@inproceedings{DBLP:conf/icml/EsserKBEMSLLSBP24,
  author       = {Patrick Esser and
                  Sumith Kulal and
                  Andreas Blattmann and
                  Rahim Entezari and
                  Jonas M{\"{u}}ller and
                  Harry Saini and
                  Yam Levi and
                  Dominik Lorenz and
                  Axel Sauer and
                  Frederic Boesel and
                  Dustin Podell and
                  Tim Dockhorn and
                  Zion English and
                  Robin Rombach},
  title        = {Scaling Rectified Flow Transformers for High-Resolution Image Synthesis},
  booktitle    = {ICML},
  year         = {2024},
}

@inproceedings{DBLP:conf/iclr/LvSSY00W25,
  author       = {Zhengyao Lv and
                  Chenyang Si and
                  Junhao Song and
                  Zhenyu Yang and
                  Yu Qiao and
                  Ziwei Liu and
                  Kwan{-}Yee K. Wong},
  title        = {FasterCache: Training-Free Video Diffusion Model Acceleration with
                  High Quality},
  booktitle    = {ICLR},
  year         = {2025},
}

@article{DBLP:journals/ijautcomp/LuZBCLZ25,
  author       = {Cheng Lu and
                  Yuhao Zhou and
                  Fan Bao and
                  Jianfei Chen and
                  Chongxuan Li and
                  Jun Zhu},
  title        = {DPM-Solver++: Fast Solver for Guided Sampling of Diffusion Probabilistic
                  Models},
  journal      = {Mach. Intell. Res.},
  volume       = {22},
  number       = {4},
  pages        = {730--751},
  year         = {2025},
}

@inproceedings{DBLP:conf/nips/ZhaoBR0L23,
  author       = {Wenliang Zhao and
                  Lujia Bai and
                  Yongming Rao and
                  Jie Zhou and
                  Jiwen Lu},
  title        = {UniPC: {A} Unified Predictor-Corrector Framework for Fast Sampling
                  of Diffusion Models},
  booktitle    = {NeurIPS},
  year         = {2023},
}
\end{document}